\documentclass[12pt,3p,times,twoside]{elsarticle}

\usepackage{lineno}
\usepackage{hyperref}
\hypersetup{
    colorlinks=true,
    linkcolor=blue,
    filecolor=magenta,      
    urlcolor=blue,
}
\modulolinenumbers[0]
\let\linenumbers\nolinenumbers\nolinenumbers

\usepackage{framed,multirow}
\usepackage{amsmath,amssymb,amsfonts}
\usepackage{algorithmic}
\usepackage{graphicx}
\usepackage{textcomp}
\usepackage{comment}
\usepackage{longtable}
\usepackage{afterpage}
\usepackage{caption}
\usepackage{adjustbox}
\usepackage{xltabular}
\usepackage{scalerel}
\usepackage{tikz}
\usetikzlibrary{svg.path}
\usepackage{tabularx} 
\usepackage{amssymb} 
\usepackage{enumitem}
\usepackage{natbib}
\usepackage{floatpag} 
\usepackage{rotating} 
\usepackage{lipsum}
\usepackage{docmute}
\usepackage{relsize}
\usepackage [english]{babel}
\usepackage [autostyle, english = british]{csquotes}
\MakeOuterQuote{"}

\journal{arXiv}

\definecolor{mycorrect}{rgb}{0, 0, 0} 
\definecolor{done}{rgb}{0, 0.502, 0} 
\definecolor{tbd}{rgb}{0, 0, 1}
\definecolor{newcolor}{rgb}{.8,.349,.1}

\begin{document}

\begin{frontmatter}
\begin{titlepage}

\title{FUTURE-AI: Guiding Principles and Consensus Recommendations for Trustworthy Artificial Intelligence in Medical Imaging}

\author[1,2]{Karim Lekadir\corref{mycorrespondingauthor}}

\cortext[mycorrespondingauthor]{Corresponding author}
\ead{karim.lekadir@ub.edu}

\author[1]{Richard Osuala}
\author[1]{Catherine Gallin}
\author[1]{Noussair Lazrak}
\author[1]{Kaisar Kushibar}

\author[3]{Gianna Tsakou}

\author[4]{Susanna Aussó}

\author[5]{Leonor Cerdá Alberich}

\author[6]{Kostas Marias}
\author[6]{Manolis Tsiknakis}

\author[7]{Sara Colantonio}

\author[8]{Nickolas Papanikolaou}

\author[9]{Zohaib Salahuddin}
\author[9]{Henry C Woodruff}

\author[9]{Philippe Lambin}

\author[5]{Luis Martí-Bonmatí}

\address[1]{Barcelona Artificial Intelligence in Medicine Lab (BCN-AIM), \textcolor{mycorrect}{Departament de Matemàtiques i Informàtica}, Universitat de Barcelona, \textcolor{mycorrect}{Barcelona}, Spain}

\address[2]{\textcolor{mycorrect}{Institució Catalana de Recerca i Estudis Avançats (ICREA), Barcelona, Spain}}

\address[3]{Maggioli SPA, Research and Development Lab, Athens, Greece}

\address[4]{Department of Health, Fundació TIC Salut i Social, Generalitat de Catalunya, Barcelona, Spain}

\address[5]{Medical Imaging Department, Biomedical Imaging Research Group, Hospital Universitario y Politécnico La Fe and Health Research Institute, Valencia, Spain}

\address[6]{Foundation for Research and Technology, Institute of Computer Science, Greece}

\address[7]{National Research Council, Institute of Information Science and Technologies, Italy}

\address[8]{Champalimaud Foundation, Computational Clinical Imaging Group, Lisboa, Portugal}

\address[9]{Maastricht University, Department of Precision Medicine, the Netherlands}

\end{titlepage}

\begin{abstract}

The recent advancements in artificial intelligence (AI) combined with the extensive amount of data generated by today’s clinical systems, has led to the development of imaging AI solutions across the whole value chain of medical imaging, including image reconstruction, medical image segmentation, image-based diagnosis and treatment planning. Notwithstanding the successes and future potential of AI in medical imaging, many stakeholders are concerned of the potential risks and ethical implications of imaging AI solutions, which are perceived as complex, opaque, and difficult to comprehend, utilise, and trust in critical clinical applications. \textcolor{mycorrect}{Addressing these concerns and risks, the FUTURE-AI framework has been proposed \cite{lekadir2023future}, which, sourced from a global multi-domain expert consensus, comprises guiding principles for increased trust, safety, and adoption for AI in healthcare. In this paper, we transform the general FUTURE-AI healthcare principles to a concise and specific AI implementation guide tailored to the needs of the medical imaging community. To this end, we carefully assess each building block of the FUTURE-AI framework consisting of (i) Fairness, (ii) Universality, (iii) Traceability, (iv) Usability, (v) Robustness and (vi) Explainability, and respectively define concrete best practices based on accumulated AI implementation experiences from five large European projects on AI in Health Imaging.
We accompany our concrete step-by-step medical imaging development guide with a practical AI solution maturity checklist, thus enabling AI development teams to design, evaluate, maintain, and deploy technically, clinically and ethically trustworthy imaging AI solutions into clinical practice.}
\end{abstract}


\end{frontmatter}

\linenumbers

\section{Introduction}

Amid hope and hype, artificial intelligence (AI) is widely regarded as one of the most promising and disruptive technologies for future healthcare. As machine learning techniques are suited for facilitating the analysis of large and complex datasets, medical imaging is the medical speciality that has seen the most developments in AI in the last years \cite{alexander2020intelligent}. With the advent of big data and machine learning, imaging AI solutions have been developed for the whole value chain of medical imaging and radiology, including image reconstruction \cite{rivenson2018phase, schlemper2017deep}, medical image segmentation \cite{tajbakhsh2020embracing, campello2021multi}, image-based diagnosis \cite{martin2020image, liu2019comparison}, and treatment planning \cite{horvat2018mr, xu2019deep}. The recent developments in the field are also well illustrated by the comprehensive list of FDA-approved AI algorithms that is maintained online by the American College of Radiology \cite{fda_cleared}. 

If suitably implemented, AI is expected to play an important role in future medical imaging, by enhancing the acquisition, processing and interpretation of medical images, by helping to extract and combine new information and imaging biomarkers for enhanced patient assessment, prediction and decision making, and thus, assisting the clinician in diagnosing and managing patients more efficiently and more accurately. However, despite the advances and developments in the field over the last years, the adoption and deployment of imaging AI technologies remains limited in clinical practice. A recent survey of clinicians performed in Australia and New Zealand showed that, while the vast majority of radiologists agree that the introduction of AI would improve their field, over 80\% of respondents have not yet used AI in their day-to-day practice \cite{scheetz2021survey}. 

At the same time, many stakeholders have expressed concerns on the potential risks, ethical implications and lack of trust in AI in healthcare in general, including medical imaging in particular. AI tools continue to be viewed as complex, opaque technologies that are difficult to comprehend, utilise and fully trust by clinicians and patients alike \cite{reyes2020interpretability}. There is a concern that the AI tools can generate undetected errors, with harmful consequences to the patient, when they are applied on imaging conditions that may differ or unexpectedly deviate, even slightly, from those used for training. Because existing imaging databases are often imbalanced according to sex, ethnicity, geography and socioeconomics, there is a risk that trained AI algorithms become biased towards under-represented groups and hence exacerbate existing health disparities \cite{seyyed2020chexclusion, kinyanjui2020fairness}. There are also concerns about the effect of AI tools on the decision-making and interpretation skills of experienced and less experienced radiologists \cite{povyakalo2013discriminate}.

Importantly, current AI solutions for medical imaging are rarely developed and validated with mechanisms to enable their monitoring throughout their deployment lifetime, to periodically assess changes in performance, especially when changes in the imaging hardware or protocol take place, or to enable continuous learning and assess its effect on the AI tools as new, additional imaging studies and richer datasets become available over time. Despite these concerns and risks, there are currently no concrete guidelines and best practices for guiding future AI developments in medical imaging towards increased trust, safety and adoption. A joint statement of European and North-American radiological associations on ethical challenges of AI in radiology recently stated that "the radiology community should start now to develop codes of ethics and practice for AI" \cite{geis2019ethics}.

\begin{figure*}
    \centering
        \includegraphics[width=0.65\textwidth]{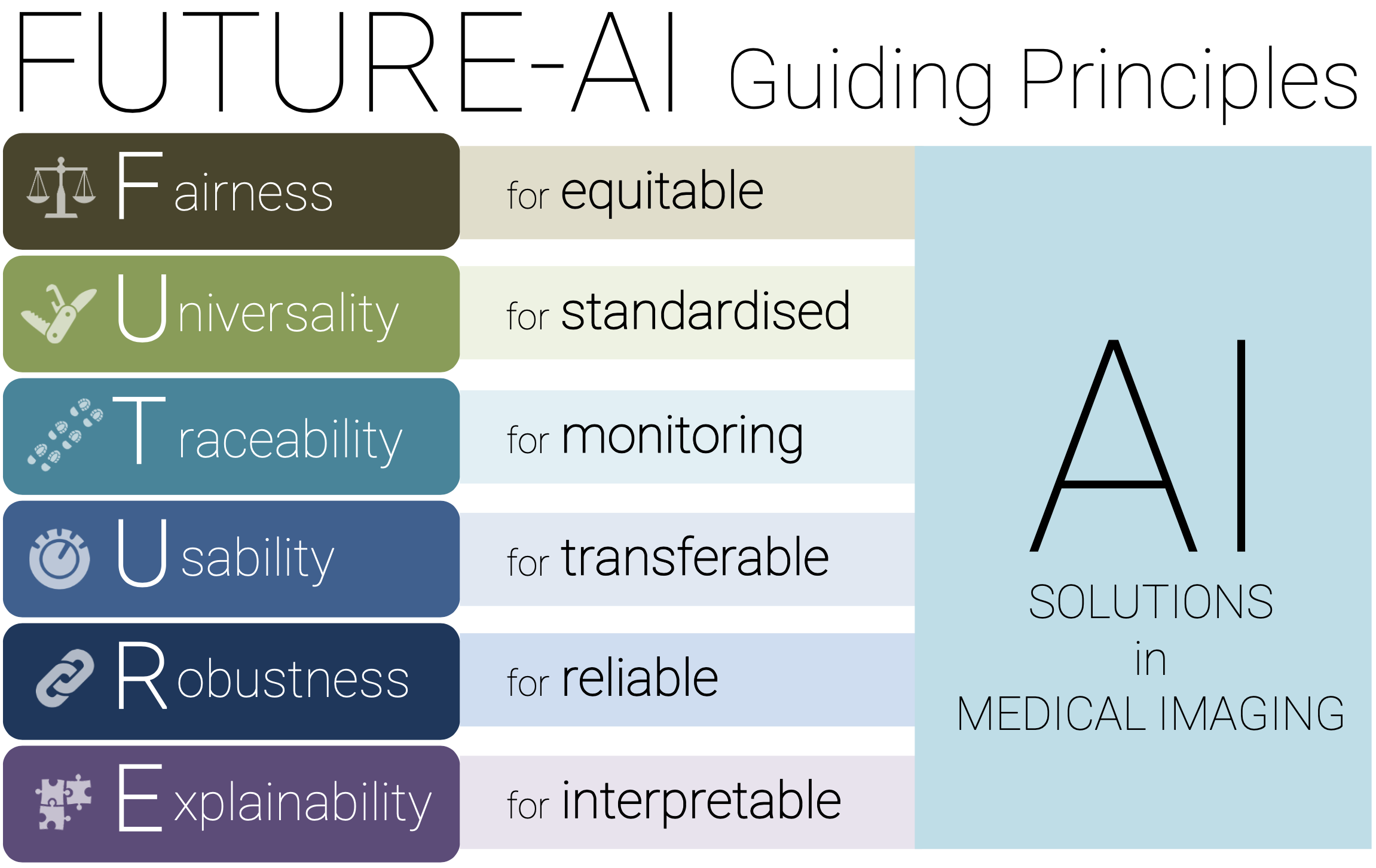}
	\caption[]{The \textit{FUTURE-AI Guiding Principles} proposed by \textcolor{mycorrect}{\textit{Lekadir et al.} \cite{lekadir2023future} based upon which we define a concise guide assessing tools and methods} for developing, evaluating and deploying in clinical practice trusted, safe and ethical AI solutions for medical imaging. \textcolor{mycorrect}{Our FUTURE-AI medical imaging AI implementation guide has been} defined based on \textcolor{mycorrect}{experiences} and best practices from five large European medical imaging AI implementation projects.}
	\label{fig:future-ai}
\end{figure*}

This paper \textcolor{mycorrect}{builds upon the} new \textcolor{mycorrect}{healthcare} guiding principles named FUTURE-AI \textcolor{mycorrect}{\cite{lekadir2023future}} and translates these into concrete recommendations and best practices for developing future AI solutions \textcolor{mycorrect}{specifically for the important domain of} medical imaging. \textcolor{mycorrect}{The latter is not only the primary area of healthcare applications from the fast-paced computer vision AI innovation cycles, but also a domain with its own numerous challenges and specific intricacies calling for correspondingly adapted methods and guidelines}. \textcolor{mycorrect}{Assessing and elucidating each of the building blocks of the FUTURE-AI principles, namely, (i) Fairness, (ii) Universality, (iii) Traceability, (iv) Usability, (v) Robustness and (vi) Explainability (depicted in Figure \ref{fig:future-ai}), we define and provide concrete} best \textcolor{mycorrect}{AI development} practices building on accumulated experiences and results from five large European projects on AI in Health Imaging (the AI4HI Network, comprising the EuCanImage, PRIMAGE, CHAIMELEON, INCISIVE and ProCancer-I projects).
The current recommendations, as detailed in this paper, facilitate the application of the FUTURE-AI guiding principles and include a \textcolor{mycorrect}{practical imaging AI maturity checklist consisting of a} set of 55 checklist \textcolor{mycorrect}{items} aimed at guiding designers, developers, evaluators, end-users and regulators of AI in medical imaging. In a step-by-step approach, these guidelines will enhance the specification, implementation, evaluation and deployment of imaging AI algorithms that can be trusted, technically, clinically and ethically, within future \textcolor{mycorrect}{medical imaging clinical} practices. 

\section{Fairness – For Equitable AI in Medical Imaging} \label{sec:fairness}
Despite continuous advances in medical research and practice, there remain important health inequalities between individuals and groups of individuals, in particular due to differences in sex/gender, age, ethnicity, income, education and geography \cite{mccartney2019defining}. There are concerns that the existing disparities may be embedded and even amplified by emerging algorithms in healthcare if these are not properly implemented \cite{paulus2020predictably}. A 2019 study received a lot of media attention when it showed that an algorithm widely used in the United States for patient referral discriminated against Black patients \cite{obermeyer2019dissecting}. The authors of the study explained that remedying this bias would increase the percentage of Black patients receiving additional help from 17.7\% to 46.5\%. Among the few works that investigated AI bias in medical imaging, a recent study evaluated the extent to which deep learning based algorithms for the detection of abnormalities in chest X-ray images (e.g., bone fractures, lung lesions, nodules, pneumonia) are biased with respect to attributes such as sex, age, ethnicity and socioeconomic status \cite{seyyed2020chexclusion}. The authors found that the highest rate of AI-based under-diagnosis was in young females (age $<$ 20 years), in Black patients, and in patients on basic health insurance, and concluded that "models trained on large datasets do not provide equality of opportunity naturally, leading instead to potential disparities in care if deployed without modification". 

Fairness can be negatively affected by both quantitative and qualitative biases. Indicative examples of quantitative biases in medical imaging follow. A review of five years’ worth of peer-reviewed articles analysed the data used in training deep learning algorithms for image-based diagnosis in various medical specialities in the United States \cite{kaushal2020geographic}. They found that among the studies in which geographic location was known, the vast majority (71\%) used training data from just three states: California, Massachusetts, and New York, and that no data at all was used from 24 of the 50 states in the country. 

In a study on fairness in cardiac image segmentation based on a training dataset that is gender balanced (52\% male, 48\% female) but ethnicity unbalanced (80\% White, 20\% other ethnicities including Asian, Black, Chinese and Mixed), the baseline AI algorithm performed consistently across men and women, but there were significant biases against the under-represented ethnic groups \cite{puyol2021fairness} (e.g., 93\% average segmentation accuracy for White subjects vs. 83\% for the Chinese subjects). Similarly, a study on missed imaging appointments in the United States, found that racial minorities and people with low socioeconomic status missed more healthcare appointments than other groups, which further contributes to their prevailing under-representation in the imaging datasets of their healthcare provider \cite{glover2017socioeconomic}.

Qualitative biases affecting fairness may consist in cognitive biases of clinicians generating, interpreting or annotating the imaging data: A study on mammography screening, for example, has shown that the detection of malignant breast lesions in patients with a minority ethnic background or low income was less likely \cite{rauscher2013potentially}\footnote{Further studies are needed to investigate as to why radiologists were less likely to detect malignant breast lesion in minority / low-income patient groups. Young age might be a confounder for lower income and high breast density i.e. it is more difficult to detect lesions in breasts with high density. On the other hand, it is likely that low income and minority ethnic background are associated with the absence of private health insurance translating to a lower standard of care for these patients, which could also have caused the lower detection rate.}.Such biases can find their ways into training datasets (e.g., datasets can contain less accurate or more missing image annotations for particularly these subject groups) and, therefore, potentially get exacerbated through AI decision support solutions trained on these datasets. 

These studies illustrate the main cause of algorithmic bias in AI: the training datasets often lack the required quantitative and qualitative diversity and balance to obtain AI solutions that can maintain the same performance across human groups and sub-populations. In particular, if an AI algorithm is trained with imaging data that are imbalanced with respect to sex \cite{Larrazabal12592}, socioeconomics or ethnicity \cite{puyol2021fairness}, and given the health differences within and across these groups, it is likely that the resulting model will lead to biased predictions. Some data imbalance is application-specific and can be more difficult to identify based on standard attributes. For example, in breast cancer, women with high breast density are generally under-represented in  fthe existing clinical registries. The problem of bias in AI is common to all medical applications, but it is even more problematic in medical imaging as personal attributes such as sex, age, ethnicity and socioeconomics are not always retained during the data preparation and anonymisation process, which minimises the possibility of patient identification \cite{diaz2021data}. To make things more complex, a recent early-stage research study \cite{banerjee2021reading} reports evidence that AI algorithms can identify the race from a person’s medical scans, such as chest and hand X-rays and mammograms, when human medical experts cannot. This raises concerns relevant not only to the \textit{Explainability} principle \ref{sec:explainability} but also to fairness and the extent to which these unexplained AI predictions could imply a different, unequal AI-based decision support outcome for the treatment of patients depending on their race, as a kind of shortcut decision boundary, even in cases where race is not a relevant treatment criterion.

Equally importantly, even if the perfect balance across diverse groups could be achieved in AI training and testing datasets, the integration of AI in real-life clinical practice can still raise fairness issues, especially with regard to its usage by experienced and less experienced clinicians and the effect that it has on their decision-making capabilities. For instance, in relation to the interpretation of mammograms, a study \cite{povyakalo2013discriminate} has found that while automated support positively influenced the decision-making of radiologists with less advanced interpretation skills, it had a negative effect on the decision-making of radiologists with advanced image interpretation skills. Evidently, this is related to over-trusting the outcome of the automated support and devaluing the clinicians’ professional experience accumulated over years of medical practice, or to misconceptions about the limitations and strengths of AI. 

The first principle of the FUTURE-AI guidelines is the one of \textit{Fairness}, which states that "imaging AI algorithms should be impartial and maintain the same performance when applied to similarly situated individuals (individual fairness) or to different groups of individuals, including under-represented groups (group fairness)". Equal access to the "highest attainable standard of health" is considered a fundamental right of every human without distinction of race, beliefs, or economic conditions \cite{10665-339554}. Medical imaging, which is an expensive but critical service in medical care, should, hence, be provided equally for all patients independently of their gender, ethnicity, geographical location and socioeconomic level. AI algorithms should not exacerbate existing health disparities, such as those indicatively presented above, but instead should facilitate and enhance access to high-quality radiology services for all individuals and groups. To assess and achieve fairness when developing imaging AI algorithms, we recommend the following quality checks.
\\
\\
\textit{Recommendations for Fairness:}
\begin{itemize}
    \item \textit{Multi-disciplinarity}: 
    The design, implementation and testing of the AI algorithms should take into account diverse perspectives brought by multidisciplinary teams of stakeholders, including AI developers, patients, radiologists, specialists (e.g. oncologists) and social scientists (e.g. ethicists). Multi-disciplinarity will help to eliminate subjectivity and identify as many possible sources of application-specific bias and inequity as possible. The views that social scientists and ethicists offer will enable AI to reach a wider, more diverse public, while engaging the communities most impacted by health disparities. 
    \item \textit{Definition of fairness}: 
    Fairness is subjective by nature \cite{pot2021not} therefore, it should be defined in both the general and the application/context specific settings. Apart from general fairness requirements that should be satisfied in all cases (e.g. with respect to sex/gender), there are often application- and medical context-specific sources of bias that should be identified taking into account the specific clinical goals and end users of the AI solution. Therefore, the very first step is to define and prioritise AI fairness requirements in a specific medical imaging setting from a combined quantitative and qualitative perspective. This entails the definition of possible sources of inequity in the given context of use during all AI development phases, including requirements, implementation and evaluation phases, as well as the specification of actions to counteract possible biases and measurements of success in doing so. Such actions should include the identification of critical steps or AI subsystems requiring "human-in-the-loop" decision-making and feedback by radiologists or physicians to avoid automation bias. Furthermore, in collaboration with domain experts such as radiologists and specialists, potentially \textit{hidden} qualitative (e.g. annotators’ cognitive biases), or quantitative biases (e.g. under-representation of high-density breasts in breast cancer imaging datasets, non-white skin colour in skin cancer imaging datasets) in the data collection and labelling should be investigated and identified for the particular clinical application, beyond the standard categories such as sex, ethnicity and socio-economics. The inclusion of a diversity of patients during the requirement analysis and prototype testing may help anticipate on some of the more complex, hidden biases. 
    \item \textit{Metadata labelling}:
    When collecting and preparing imaging databases for developing and testing new AI algorithms, standardised metadata and key variables (e.g. sex/gender, ethnicity, geography) which allow for the identification of groups and the verification of AI fairness should be included, in addition to the anonymised imaging data. However, this should be achieved while ensuring data privacy and informed consent according to existing data regulations, e.g. the European General Data Protection Regulation (GDPR). To ensure individual fairness, metadata which allow measuring similarity of medical situations should be included to verify equal treatment of all similar cases (e.g. clinical information). 
    \item \textit{Estimation of data (im)balance}: 
    After defining what fairness is in the specific AI application context, the diversity and distribution within the training and testing datasets should be carefully planned and inspected. In particular, the data should be balanced as much as possible across sex, age, ethnic and socio-economic groups. Mechanisms such as random sampling, stratified sampling and adaptive sampling can be used to increase data balance and representativeness. Biases due to the nature of the inclusion and exclusion criteria should be analyses and reported. A recent study of possible biases in the selection of individuals (Supp. Tables 5 \& 6) demonstrated that excluded individuals due to missing values (in staging, primary site, tumour size and differentiation) tended to live further away from the institution, had differences in some of the calculated clinical indices (e.g., the Framingham risk score) and had a higher percentage of deaths than the selected cohorts \cite{morin2021artificial}.
    \item \textit{Multi-centre data collection}: 
    When possible, training and testing datasets should be multi-centric, spanning across several radiology centres and/or localities and countries. In this case, the AI developers and collaborating clinicians should examine the extent to which the variations in imaging quality, imaging protocols and sample size may impact the fairness of the AI models across radiology centres and/or localities. In particular, the fair application of the AI algorithms when the datasets are applied with imaging equipment of reduced quality (such as in low-to-middle income countries) should be investigated.
    \item \textit{Transparency of fairness}: 
    The process of collecting and preparing the datasets used for training and testing AI solutions should be transparent and documented, including information on data diversity and imbalance. Where possible, it is recommended to link this quality check with the \textit{Explainability} principle \ref{sec:explainability}, which can help in the identification of the reasons why a specific AI-based outcome is unfair and help to define counter-measures. 
    \item \textit{Fairness evaluation and metrics}: 
    For bias estimation in imaging AI, dedicated metrics and statistical tests should be considered, such as True Positive Rates (TPR), Statistical Parity, Group Fairness, Equalised Odds and Predictive Equality \cite{barocas2017fairness}. The use of software toolkits, such as IBM’s AI Fairness 360, may also be deployed to regularly check for unwanted biases in AI algorithms \cite{bellamy2019ai}.
    \item \textit{Application of counter-measures}: 
    If biases are identified during any stage of development or testing of AI, mitigation measures should be investigated and evaluated, including (1) pre-processing approaches to improve the training dataset through re-sampling (under- or over-sampling), data augmentation (image synthesis using adversarial learning) or sample weighting to neutralise discriminatory effects; (2) in-processing approaches that modify the learning algorithm in order to remove discrimination during the model training process, such as by adding explicit constraints in the loss functions to minimise the performance difference between subgroups of individuals (e.g., learning bias-free representations via adversarial loss \cite{li2021estimating}); and (3) post-processing approaches to correct the outputs of the AI algorithm depending on the individual’s group, such as by using the equalised odds post-processing technique \cite{pleiss2017fairness}. All these techniques should be thoroughly evaluated to ensure their positive impact on fairness.
    \item \textit{Continuous monitoring of fairness}: 
    Once implemented, the AI algorithm should be thoroughly and continuously evaluated and retrained for fairness, once again by using imaging datasets with adequate population diversity. 
    \item \textit{Training material and deployment effects}: 
    Appropriate training material for targeted end users of AI solutions (e.g. radiologists, oncologists) should be an inseparable part of any AI deployment package. Such training material should seek to inform clinicians about the strengths and limitations of the given AI solution, raise their awareness about possible misconceptions or misuses in the specific clinical setting, e.g. due to overgeneralisation of the diagnostic task that the AI tool is meant to support, or over-trust in the AI outcome. Once the AI solution has been introduced in the everyday practice of the targeted end users, the training material should be continuously updated to take into account new findings about potential AI effects on the clinical practice that may come up over time. 
\end{itemize}

\section{Universality – For Standardised AI in Medical Imaging} \label{sec:universality}
While a certain degree of diversity in the design and implementation of AI solutions in medical imaging is both expected and desirable to promote innovation and differentiation, the Universality principle recommends the definition and application of standards during algorithm development, evaluation and deployment. These standards, including technical, clinical, ethical and regulatory standards, will achieve at least three key objectives: (1) They will enable the development of AI technologies with increased interoperability across clinical centres, radiology units and geographical locations; (2) they will promote a culture of quality, safety and trust in imaging AI based on well-proven, widely accepted frameworks; (3) they will facilitate co-creation and cooperation in imaging AI between AI developers, manufacturers, radiologists, physicians, data managers and healthcare bodies based on unified language and common approaches. In contrast, the development of imaging AI algorithms without relying on community standards will inevitably result in technologies that are restricted, incomparable and intransparent, and, therefore, likely will lack public and clinical acceptance.

A number of initiatives have already been established to define standards for artificial intelligence, though they are focused on AI in general. In 2018, the International Organisation for Standardisation (ISO) and the International Electrotechnical Commission (IEC) started a project -still ongoing- on AI standardisation (Subcommittee ISO/IEC JTC 1/SC 42 Artificial intelligence \cite{ISO_6794475}). At the national level, in the US for example, the National Institute for Standards and Technology (NIST) issued a plan for long-term definition and maintenance of technical standards and related tools for AI \cite{NIST2019}. In the medical imaging community, there is a need for concerted efforts by leading research centres, medical associations and radiological societies, open-source communities, standardisation bodies and private companies in the field, to define clinical and technical standards for imaging AI. At the same time, while standardisation holds evident benefits for interoperability, adoption and trust, there is a risk of hindering innovation by too many or inflexible standards. Hence, in FUTURE-AI, we recommend a minimal set of key standards for imaging AI focused on the following aspects:
\\
\\
\textit{Recommendations for Universality:}
    \paragraph{Clinical definition of the image analysis tasks}
    Imaging AI algorithms -such as for estimating the patient’s diagnosis, treatment or prognosis- can be built based on a wide range of clinical definitions and categorisation schemes. COVID-19 diagnosis based on chest imaging scans, for example, has been proposed by various research and clinical centres using multiple classification approaches, i.e. (1) two categories (Presence or absence of the disease); (2) four categories \cite{simpson2020radiological} (Typical appearance, indeterminate, atypical, negative for pneumonia); (3) six categories \cite{prokop2020co} (very low probability, low, indeterminate, high, very high, PCR positive); and (4) various scoring systems of lesion severity in the lungs \cite{larson2021regulatory} (e.g. a 24-scale based on a 0 to 4 severity rating for each of six lung zones \cite{wang2020temporal} or a 35-scale based on a 0 to 7 severity rating for each of five lung lobes \cite{huang2020timely}). The possible use of multiple definitions for the same clinical task can greatly limit the clinical interoperability, benchmarking and acceptance of the resulting imaging AI algorithms. To ensure widespread acceptance, future imaging AI algorithms should be designed based on consensus definitions of the clinical tasks, as established and maintained by recognised, non-for-profit entities such as medical societies. These definitions should detail the criteria for making the clinical assessment, as well as the descriptions of the imaging measurements, image labelling instructions and classification categories.
    \paragraph{Software standardisation}
    With the rapid advancement in the fields of AI and medical imaging an increasing number of new software tools, solutions and libraries are becoming available. However, these tools and libraries differ in their scope, approach, programming language, documentation, integrability with other solutions, and in their adherence to standards such as the DICOM standard \cite{diaz2021data}. The libraries and frameworks upon which an imaging AI solution is developed need to be inter-compatible while also allowing to fulfil the functional requirements of the clinical task at hand. Moreover, to allow for deployment in different clinical settings, AI solutions need to be able to run on different hardware with different operating systems, within different software systems and IT landscapes, and under different system performance, security, privacy and data processing constraints. A principled approach towards choice of well-established libraries and proven frameworks can help to prevent potential incompatibility issues at later project stages while also facilitating analysis, extension, maintenance, upgrades, monitoring, migration, integration, and audit of the imaging AI solution. Even if reference implementations and designs are followed, it is recommended to document the rationales behind each of the design decisions in AI model development such as framework choices, modules, and whether the AI solution will be a standalone solution or rather serve as integral part of a larger image processing platform. 
    \paragraph{Image annotation standardisation}
    Despite efforts and guidelines towards structured, standardised diagnostic reporting and data labelling, reports in radiology are conventionally written in free text formats \cite{european2018esr}. This increases uncertainty and subsequent relabelling and annotation efforts when preparing such data for automated processing and as input for AI solutions \cite{willemink2020preparing}. Standardisation in reports, labelling and annotation ensures the completeness and interoperability of medical imaging datasets. For instance, annotations should follow one common format such as contours, as opposed to bounding boxes or circles around anatomies of interest. Once such a common format is found to satisfy clinical use-case requirements, and once this format is agreed upon with the annotating clinicians and end-users, guidelines for annotation are to be establish that introduce a consistent annotation collection process, which further aids standardisation across annotations, file formats, and annotation storage options. In this regards, and as many different tools and platforms exists for the purpose of annotation \cite{diaz2021data}, it is recommendable that the annotators uses the same annotation software, that should be mature, well established and documented, as well as chosen based on its compliance with the requirements of the clinical use-case at hand.
    \paragraph{Standards for quantification of imaging biomarkers}    
    Existing research has shown that the values of common imaging features, such as radiomics features, can vary when they are calculated from different software packages, which implement the same equation using varying configurations and image processing codes \cite{mcnitt2020standardization, zwanenburg2020image}. This has motivated the establishment of the Imaging Biomarker Standardisation Initiative (IBSI), an international consortium of 25 teams which defined conventions that provide unique schemes for calculating radiomics-based imaging biomarkers \cite{zwanenburg2020image}.
    \paragraph{Criteria and metrics for imaging AI evaluation}
    To enable objective, widely accepted, community benchmarking of future imaging AI algorithms, standard criteria and metrics should be used for their evaluations based on the consensus literature. For instance, for evaluating the accuracy of image segmentation tools, the Dice Similarity Coefficient (DSC) and the Hausdorff Distance (HD) have been universally adopted in the image computing community. To assess the robustness of imaging features, the coefficient of variation (CV) and the intraclass correlation coefficient (ICC) have been widely used. 
    \paragraph{Reference imaging datasets for AI benchmarking}
    To promote objective and comparative assessment of algorithm performance, reference imaging datasets have been proposed. They consist of curated sets of images acquired from representative real-world cases in which the resulting imaging AI algorithms will be typically used. There are already several reference datasets in many subspecialties of medical imaging, including in brain MRI (e.g. ADNI: Alzheimer's Disease Neuroimaging Initiative \cite{weiner2017alzheimer, wyman2013standardization}, BraTS: Brain Tumor Segmentation Dataset \cite{menze2014multimodal}), cardiac MRI (M\&Ms: Multi-Centre, Multi-Vendor \& Multi-Disease Cardiac Image Segmentation \cite{campello2021multi}), echocardiography (CAMUS: Cardiac Acquisitions for Multi-structure Ultrasound Segmentation \cite{leclerc2019deep}), chest X-ray (CheXpert: Chest Radiography Dataset with Expert Comparison \cite{irvin2019chexpert}), bone imaging (Spineweb: Resource for Spinal Imaging\footnote{\href{http://spineweb.digitalimaginggroup.ca/Index.php?n=Main.Datasets}{http://spineweb.digitalimaginggroup.ca/Index.php?n=Main.Datasets}}), COVID-19 imaging (RICORD: RSNA International COVID-19 Open Radiology Database\footnote{\href{https://www.rsna.org/covid-19/covid-19-ricord}{https://www.rsna.org/covid-19/covid-19-ricord}}), breast imaging (QIN-Breast: Longitudinal Breast Imaging from the Quantitative Imaging Network \cite{li2015multiparametric}), and many more (see for example the curated cancer imaging collections hosted at the Cancer Imaging Archive\footnote{\href{https://www.cancerimagingarchive.net/collections}{https://www.cancerimagingarchive.net/collections}} \cite{clark2013cancer}).
    \paragraph{Reporting of imaging AI studies}
    To enable wide acceptance of imaging AI, it is important that key details of the algorithms are clearly reported. This will enable developers, researchers and other stakeholders to critically appraise the relevant information on the design, development and validation, as well as to replicate the AI algorithms and results. Even before the advent of AI, guidelines were proposed for standardised and comprehensive reporting of predictive models in medicine. The most widely used of such guidelines is the TRIPOD statement\footnote{\href{https://www.tripod-statement.org/}{https://www.tripod-statement.org/}} (Transparent Reporting of a multivariable prediction model for Individual Prognosis Or Diagnosis) \cite{collins2015transparent}, which lists key items to report when describing and evaluating clinical prediction models, including: (i) Title, abstract, background, and objectives; (ii) Methods: Source of data, participants, predictors, sample size, missing data, type of prediction model and other model-building procedures; (iii) Results: Participants (number and characteristics), performance measures, confidence intervals, model updating; (iv) Discussion: Limitations (e.g. non-representative sample, missing data), interpretation (incl. comparison to similar studies), implications (e.g. potential clinical use); (v) Other information: Supplementary information, funding. Recently, the TRIPOD steering committee announced that they are working on extended reporting guidelines named TRIPOD-AI \cite{collins2019reporting} and focused on AI-driven predictive models in healthcare, which will be published in the foreseeable future. Furthermore, in medical imaging, the TRIPOD has been adjusted into the Radiomics Quality Score (RQS) for reporting radiomics-based predictive models\footnote{\href{https://www.radiomics.world/rqs}{https://www.radiomics.world/rqs}} \cite{lambin2017radiomics}.
In summary, we recommend the following checklist for maximising the universality of imaging AI algorithms:
\begin{itemize}
    \item \textit{Definition of clinical task}: The AI manufacturers should ensure that the clinical imaging tasks they aim to address using AI are based on universal clinical definitions, as defined and promoted by recognised non-for-profit medical societies in the area of interest.
    \item \textit{Software standardisation}: A standardised approach to AI software design enables developers, maintainers, and auditors to understand, analyse, maintain, migrate, integrate, and extend the imaging AI solution. AI software solution design conventions, code standards, and proven libraries and frameworks should be used to readily allow for extension and integration with other clinical software systems. 
    \item \textit{Image annotation standardisation}:The collection and storage of clinical annotations should follow a standardised approach, where annotations are comparable, reproducible and in one common format (e.g., delineated organ contours). A common, standardised and reproducible format for annotating and labelling imaging datasets enables training AI models transparently and with clear focus on the respective clinical tasks.
    \item \textit{Variation of quantified biomarkers}: Universal definitions and calculation methods for estimating imaging biomarkers should be employed when building feature-based AI models in medical imaging, such as by using IBSI-compliant software packages or ComBat harmonisation of the radiomics features.
    \item \textit{Evaluation metric selection and reporting}: When evaluating imaging AI algorithms, universal criteria and metrics should be used to enable comparative, community-driven assessment of the model's performance and properties.
    \item \textit{Reference dataset evaluation}: Furthermore, when possible, evaluation results should be generated and reported based on reference, open-access imaging datasets that are representative of real-world clinical cases.
    \item \textit{Reporting standards compliance}: Finally, standardised guidelines such as TRIPOD-AI and the RQS should be employed for reporting imaging AI studies.
\end{itemize}

\section{Traceability – For Transparent and Dynamic AI in Medical Imaging} \label{sec:traceability}
The High-Level Expert Group on AI (AI HLEG), a group of experts appointed by the European Commission to provide advice on its artificial intelligence strategy, recently released a document with guidelines to attain “trustworthy AI” \cite{eucommission2019} mentioning seven key requirements, with transparency being one of these requirements. Transparency is among the top principles promoted by other international initiatives on ethics in AI \cite{jobin2019global, AsilomarAI2017},  as mapped by a recent review on existing ethics guidelines for AI across many fields including medicine \cite{jobin2019global}. Transparency is a complex construct that evades simple definitions. It can refer to explainability, interpretability, openness, accessibility, and visibility, among others \cite{felzmann2019robots}. However, in the AI HLEG’s document, transparency was explained in terms of three components: \textit{traceability, explainability, and open communication about the limitations of the AI system}. In the present section, we will focus on the aspect of traceability, a key requirement for trustworthy artificial intelligence (AI), and related to \textit{"the need to maintain a complete account of the provenance of data, processes, and artefacts involved in the production of an AI model"}\cite{mora2021traceability}. 

In essence, traceability refers to the mandate to document the whole development process and to track the functioning of an AI model or an AI-based system used to support medical imaging analysis and interpretation. As the variability of AI in the medical imaging space is high, the documentation should be complete and detailed, in compliance with the best practices and the standards for software development regulated by certification organisations, as in the case of software as a medical device \cite{eu01, fda2021}. In other words, the data sets, the processes, the reference clinical gold standards, and the contributors that yield the AI system should be documented to the best possible standard to allow for traceability and an increase in transparency \cite{bucker2021transparency}. This entails to provide details about data gathering, with information about the clinical sites, the devices used, the acquisition protocols, dataset composition (see the \textit{Fairness} principle \ref{sec:fairness}), data labelling, also with respect to annotation contributors, used annotation tooling, the underlying reference standards (e.g., the version of PI-RADS or BI-RADS used by radiologists), as well as the development framework, and the algorithms used. The endeavour of documentation also includes the decisions made by the AI system \cite{goodman2016does}, to enable identifying the reasons why an AI-decision was erroneous, which, in turn, could help prevent future mistakes. 

Documenting the development process of an AI model and making the model \textit{transparent and traceable by design} \cite{felzmann2020towards} is key to avoid any "grey" area about what happens if something goes wrong when the model is used in clinical practice. 

In this respect, traceability in AI shares part of its scope with general-purpose recommendations for provenance and it is also supported to different extents by specific tools used by practitioners as part of their efforts in making data analytic processes reproducible or repeatable \cite{mora2021traceability}.

Provenance, as defined by the PROV W3C recommendations \cite{W3C}, is \textit{"information about entities, activities, and people involved in producing a piece of data or thing, which can be used to form assessments about its quality, reliability or trustworthiness"}. In the field of medical imaging oriented AI-based systems, provenance data can be used to manage, track, and share machine learning models, but also to many other applications, such as to detect poisonous data and mitigate attacks \cite{baracaldo2017mitigating}. Transparency in AI development and deployment requires clear communication of a variety of tasks, such as data management, model development, deployment, and updating/refinement, as well as tasks related to the functional details of the system. In particular, in recent years, importance has been placed on data provenance and on the tracking of the entire machine learning lifecycle. Two concepts are key in relation to this: Data transparency and model transparency.

\begin{enumerate}
    \item \textit{Data transparency}: 
    As the outcomes of AI/ML systems depend directly on the data training process, transparency in data collection, utilisation and storage, is an area of significant concern in trustworthy AI. Data provenance (or data lineage) methods are required to improve replication, tracing, quality assessment in data use and data transformation processes \cite{herschel2017survey}. In recent years, a series of standards have appeared for recording data provenance such as Open Provenance Model \cite{moreau2011open}, Provenir \cite{sahoo2009provenir}, and the W3C standard PROV-O \cite{W3C_2}, whereas many of them have been devised specifically for tracking of data and data transformations during the machine learning lifecycle such as  PROV-ML, ProvLake \cite{souza2019provenance}, and Hippo \cite{zhang2017diagnosing}. While these solutions assist with internal data provenance, several researchers have also advocated for private, secure, and standardised methods for data sharing. Datasheets for datasets, is a standardisation method proposed by Gebru et al \cite{gebru2018datasheets} for documenting, among others, every dataset's motivation, composition, collection process, and recommended uses.  Such dataset documentation aims to improve the communication between dataset creators and its users, while also encouraging the prioritisation of transparency and accountability in the ML community.
    \item \textit{Model Transparency}: 
    Due to the rising complexity in modelling, model transparency and provenance methods have also gained interest. Research has focused both on end-to-end tracking of provenance information in the machine learning lifecycle, and on evaluating models for performance and trust. In this context, several modelling provenance solutions have been proposed. Schelter et al \cite{schelter2017automatically} propose a system for the extraction and storage of meta-data and provenance information commonly observed in the ML lifecycle. Hummer et al \cite{hummer2019modelops} propose ModelOps, a cloud-based framework for end-to-end AI pipeline management. One of the key components of ModelOps is a domain abstraction language with first-class support for the common artefacts in AI solutions. This includes datasets, model definitions, trained models, applications, and monitoring events, as well as the algorithms and platforms used to process data, train models, or deploy applications. Further, several tools for complete asset tracking of AI pipelines have also been developed, focusing on tracking model inputs, results, and production processes \cite{gharibi2021automated, zaharia2018accelerating}. In regards to AI documentation, a recent trend is the use of FactSheets \cite{arnold2019factsheets} for communicating \textit{"purpose, performance, safety, security, and provenance information"} from the creator to the user of an AI service.
\end{enumerate}

In light of these considerations, transparency and traceability are instrumental to address other key concerns about AI models and systems, namely reproducibility, auditability and accountability. Accountability is one of the key principles for Trustworthy AI \cite{eucommission2019}, as stated by the AI High-Level Expert Group, and has been translated into the Assessment List on Trustworthy Artificial Intelligence (ALTAI) and included into the Proposal for a Regulation Laying down Harmonised Rules on Artificial Intelligence by the European Commission \cite{eu02}. Currently, the question of accountability -when an AI-based system is deployed in real clinical settings and either fails or its outcomes produce unexpected side effects- is still open and burning \cite{geis2019ethics}. The problem affects any algorithmic application that supports decision-making and it is well known and debated in the ethic, social and legal communities \cite{mittelstadt2016ethics}. In the medical and radiology domains, this question is considered in a collaborative work of the American and European radiology and medical physics societies as well as in the guidelines published by the Royal Australian and New Zealand College of Radiologists \cite{principles2019}. Under current laws, physicians that are compliant with the standard of care are not held liable for an unwanted outcome and this still holds when the decision is based on the results of an AI model \cite{sullivan2019current}. Some works have debated the implications on liability when AI is in place in radiology and healthcare in general \cite{neri2020artificial, price2019potential}; however, a dedicated regulation is still an open issue. Documenting and tracking the development process and the functioning of an AI model is key to reconstruct all the pieces of information the physician or the radiologist used to make their decision when using AI models. Overall, the recommendations and code of conduct proposed in this paper are a step forward to regulate and support the definition of an AI-aided standard of care.

Considering the data-inductive and dynamic nature of AI models and systems, traceability does not end with documenting the development process and the testing activities, but it also entails the process of maintaining the AI model or system, by tracking its behaviour over time and detecting any drift from its training settings or previous states. Indeed, AI models are adaptive, non-deterministic systems, whose testing does not and cannot involve all the variables and changing contexts of real-world settings. Thus, testing the systems before its deployment, although extensive to guard against potential ethical concerns, cannot cover the whole host of scenarios and cases that could be encountered in practice. Moreover, clinical practices and technologies continuously evolve as new imaging technologies, evidences and clinical findings come forth, thus yielding new guidelines, new protocols, or new diagnostic devices and procedures. Finally, training is often limited to a set of cases that can exhibit several kinds of biases or limitations, as debated in section \ref{sec:fairness} on \textit{Fairness}.

These phenomena call for ongoing surveillance of AI models, and a maintenance system that tracks over time the performance, vitality and conduct of the AI models after their deployment in clinical settings. Such a maintenance system might implement a continuous monitoring of the AI models to guarantee their sustained quality, but also to close the \textit{data feedback loop}, by taking advantage of the new data, new knowledge and feedback coming from the clinical production settings. It is well-known that a model's performance degrades over time when evaluated in real world, as several phenomena drive a decay (i.e., mainly model/concept and data drifts).

\textbf{Concept drift} \cite{Sammut2010} refers to a phenomenon in the practical application of AI in which some underlying statistics or characteristics of one or more variables changes after the deployment of a model and as a result the AI model's predictive accuracy changes. In a recent study \cite{jameel2018fully} the authors have presented few potential types of concept drift, e.g. novel class arrival and class evolution. Many static AI models for medical imaging have been developed with curated, hand-picked datasets. As a result, static AI models are not designed to remain connected to real-time changes in their production environments and are prone to \textit{"concept drifting"} in time. This is demonstrated in \cite{pianykh2020continuous}, where three versions of the same model were trained on gradually ageing radiology data. Despite better quality achieved with large training sets (12 months of data), all three models became significantly inaccurate when their training data aged beyond 8 months. Concept drift is a problem that can be managed by periodically or continuously testing and updating the models or in some cases deploying models that take the possibility of concept drift into account \cite{casado2021concept}.

\textbf{Data drift}, on the other hand, which underlies model drift and is also often referred to as dataset shift \cite{quionero2009dataset}, is defined as a change in the distribution of data. Production data can diverge or drift from the baseline data over time due to changes in the real world. In the domain of medical imaging this can, for example, be a result of new imaging modalities introduced in a local site, a new version of their software, or adjustments to data acquisition procedures.

Therefore, once implemented, ongoing surveillance is needed to monitor and recalibrate AI algorithms \cite{minne2012effect}. This surveillance is also needed for dynamic algorithms that continuously update themselves based on practice data and published clinical evidence. Thus, evaluation is likely to shift from a one-off activity to a continuous process to ensure that the use of AI, including those incorporating dynamic algorithms, is meeting expectations and adherence to clinical standards of care. The model maintenance can, hence, be seen as a way to nurture the model, as it can take advantage of the new knowledge coming from the real-setting scenario, thus producing an improvement of the original version released.
Not by chance, when it comes to traceability, the ALTAI tool\footnote{\href{https://futurium.ec.europa.eu/en/european-ai-alliance/pages/altai-assessment-list-trustworthy-artificial-intelligence}{https://futurium.ec.europa.eu/en/european-ai-alliance/pages/altai-assessment-list-trustworthy-artificial-intelligence}} explicitly includes some recommendations pertaining to monitoring the quality and to logging the outputs of an AI system (see the ALTAI document for the complete list of questions) 
In accordance to these considerations, we envisage that to ensure traceability \textit{a governance framework/tool for the whole AI model lifecycle} should be put in place to ensure the following key aspects:
\begin{enumerate}[label=\roman*]
    \item the maintenance of up-to-date documentation on policies, motivations, responsibilities and logging information.
    \item the validation procedures and sandbox test analyses for safety purposes.
    \item the continuous on-line or periodic monitoring of model conduct and performance, to orchestrate any remediation needed to keep AI models well-performing, unbiased and ethical for as long as models are in use in the clinical settings.
\end{enumerate}
%
One sign of the importance of this issue is the rise of MLOps (DevOps for ML) as a signal of an industry shift from technology R\&D (how to build models) to operations (how to run models). To ensure the transparency and traceability by design of the AI models, we propose keeping a structured documentation of each step of the production process, as suggested in the following recommendations.
\\
\\
\textit{Recommendations for Traceability:}

\begin{itemize}
    \item \textit{Model scope}: when starting the development of a model, a precise definition of the model's scope should be agreed upon with the radiologists and/or the clinicians and described in terms of model's intended use, use case scenarios, the intended output, supported model inputs, the underlying biological phenomenon and any known limitations of the diagnostic/prognostic problem faced. In this regard, it is also recommended to discuss and document related use cases and scenarios that are outside the model's scope, which serves to transparently highlight the limitations and realistic expectations of the model.
    \item \textit{Data provenance}: In addition to the recommendations coming from the Fairness principle, a complete documentation of the imaging and the related clinical/genomic/pathology data required for the development of the AI model should be maintained in accordance with an appropriate data provenance standard (e.g., DataSheet or PROV-O). This should be done by including information about data provenance and ownership, acquisition protocols, devices, and timing. For the imaging data, the DICOM tags, duly anonymised, should be retained to keep details about the acquisition parameters. 
    \item \textit{Data localisation}: we also suggest to keep track of the data location over the network, which could be relevant to federated learning approaches, and to analyse dataset statistics with respect to the capability to represent the phenomenon at hand (e.g., distribution analyses). This analysis is relevant to detect concept and model drifts. Quantifying missing values and any gaps or known biases is also advisable as well as documenting breaks in the data supply and noting down when the input data are erroneous, incorrect, inaccurate or mismatched in format.
    \item \textit{Documenting data preparation}: A multitude of diverse data preparation tools and approaches exist , which illustrates the importance of documenting the pre-processing pipelines when preparing and curating the data. It should be reported whether and which data quality and standardisation procedures are put in place and, then, specifying their details, by describing whether the algorithms are ad-hoc or borrowed from existing libraries and tools, by specifying the requirements for their applications as well as any parameters set for them.
    \item \textit{Specification of clinical references}: the radiological or clinical standards or biomarkers used as reference should be carefully detailed (e.g., PIRADS, BIRADS, Gleason score). If data labelling or segmentation are used, the authorship and authors’ expertise and experience, alongside the tools and approaches used, should be detailed along with the results of any stability or consensus analysis and known limitations.
    \item \textit{Training recording}: the model training process should be carefully documented by including the standardised descriptions of imaging and non-imaging features, by detailing the training approach, the assumptions made, the methods used for parameters’ and hyper-parameters’ optimisation, the weight initialisation and the framework used (also including the framework version). A ModelOps framework can be of service to keep record of all these pieces of information.
    \item \textit{Validation documentation}: The validation process should be duly described in terms of evaluation metrics, cross-validation approach, decision thresholds as also agreed with clinicians, confidence intervals, degree of uncertainty of the output, benchmarking information, and auditors. Additional information about the results provided by the AI model should come from the explainability principle, which will be further discussed in section \ref{sec:explainability}.
    \item \textit{Final model details}: The final model released should be described with a standardised description of the model's architecture, interfaces, I/O data structures, and the limitations and known points of failure.
    \item \textit{Traceability tool}: Each AI model should be developed together with a traceability tool, that will enable to monitor the live functioning of the AI tool to, for instance, flag and record errors, deviations, and degradation in performance. The main statistics of the model should be recorded in a model registry and include the model's functions and predictions while running in a clinical production settings, but also the model's evolution over time. Feedback from clinicians/radiologists/decision makers should be recorded whenever possible. The traceability tool might be included in the ModelOps framework. In the desirable scenario where the model is capable of and configured for online learning from production data or from the feedback provided by health professionals, this learning and the tools and processes used therein should also be recorded in the traceability tool.
    \item \textit{Model passport}: All the above pieces of information should be included in a standardised format into a passport of the AI model, which should accompany the model during its whole lifecycle with a rich set of metadata that guides the model adoption in clinical practice, supports its usability and makes the model auditable. The passport should also include general information, such as the team responsible for the whole development or part of it; the date and information of each of the model's releases; the model's current version; the model type, reference and license; the contact details. This information will make the passport a viable solution to sustain accountability of the model. The passport should be maintained up-to-date by (automatically) recording any new version and information on the live functioning of the model, while also including a clear plan for periodic checks and updates of the model. 
    \item \textit{Accountability and risk specification}: the model passport should be kept updated and should contain information about the code of conduct followed apart from an evaluation of the risks that may be raised by usage of the AI model or system. The risk evaluation may be in accordance with the Proposal for a Regulation Laying down Harmonised Rules on Artificial Intelligence by the European Commission.
\end{itemize}

\section{Usability – For Effective and Beneficial AI in Medical Imaging} \label{sec:usability}
According to ISO 9241-11 definition \cite{ISO_63500}, "Usability is the extent to which a product can be used by specified users to achieve specified goals with effectiveness, efficiency, and satisfaction in a specified context of use". Usability is a key characteristic of every product offering tangible advantages such as faster acceptability, saving of costs and user satisfaction. Although the basic principles remain the same, each application domain needs to be analysed in detail in order to promote enhanced usability and design dedicated usability testing \cite{hertzum2020usability}. A recent study \cite{filice2020case} stressed the unmet need to consider human factors and employ user-centred design to achieve maximal usability, accelerate AI adoption and achieve the desired paradigm shift based on Radiology AI solutions. In more detail, the authors suggest user-centred design principles for each of the AI development model life cycle phases e.g. observation of the clinical environment and user needs assessment in the design phase, iterative user testing in the development phase, clinical workflow integration in the implementation phase and performance monitoring in the long-term use phase. Critically, they stress the need to avoid previous mistakes in health technology applications related to   computer-aided diagnosis (CAD) or electronic health records (EHR) where user factors were neglected leading to poor adoption and errors. The FDA has issued a guidance document for medical device manufacturers on human factors and usability engineering \cite{fda2016}.

Despite the numerous AI tools, human factors are still not being adequately addressed since there are very few relevant scientific publications mainly addressing the need for future usability testing in AI. However, previous relevant experience mainly in Picture Archiving and Communications System (PACS) systems has highlighted the need for assessing usability and demonstrated that user satisfaction can vary significant when different products are evaluated. In a relevant study \cite{jorritsma2014merits}, two sequential versions of commercial PACS software were evaluated 6 months apart by five radiologists with varying PACS experience. They reported 22\% improvement in performance time and 30\% decrease in the number of errors in the second version compared to the first. In a more recent study \cite{abbasi2020investigating}, user satisfaction was assessed by three resident radiologists and the results revealed that PACS has not fully met all the demands of physicians and has not achieved its predetermined objectives, such as all-access from different locations. In a more extensive study \cite{farzandipour2021usability}, 200 individuals using the PACS in several hospitals performed usability evaluation based on the standard Computer System Usability Questionnaire (CSUQ) consisting of 5 sections and 19 items. The results highlighted significant differences in terms of information quality, user interface quality, overall user satisfaction and usability of PACS. At the same time, it was demonstrated that there is a need to speed up the image processing tasks and avoid system failures while it was suggested that the information quality and user interface of systems be improved by using appropriate analysis and needs assessment of the end users. Last, a usability study on integrating CAD to PACS reported numerous weaknesses that users considered important in the context of the integrative workflow such as efficient handling and fast computation \cite{geldermann2013black}. 

AI and Radiomics have shown great potential in many areas of healthcare, including clinical oncology; however, the clinical use of this technology is still in its infancy \cite{sollini2019towards, kelly2019key, faes2019automated}. For the last five years, we have witnessed a tsunami of radiomics publications \cite{leiner2021bringing, song2020review}, demonstrating potential novel clinical applications of radiomic signatures or nomograms predominantly in the field of oncologic imaging. The problems that radiomics attempt to address are either tasks already accomplished by humans currently offering unsatisfactory performance or tasks that are currently impossible to achieve through human visual inspection and interpretation of medical images by radiologists. More specifically, the topics that radiomics are primarily focusing on are related to the prediction of treatment response, before, during or early after the completion of therapy, the accurate patient stratification related to disease prognosis taking as end-points survival-related outcomes (OS, PFS), and the prediction of risk for local or distal recurrence \cite{sollini2020artificial}.

In addition, another popular topic is the association of radiomics features often called radiomics signatures with surrogate biomarkers including molecular\cite{khodabakhshi2021non}, genomic \cite{hoshino2021radiogenomics}, or pathomics \cite{alvarez2020identifying} since radiomics are non-distractive and non-invasive; therefore, they can be easily obtained throughout the entire disease continuum. The biggest problems regarding the latter efforts were related to the study design that in the vast majority of the cases was based on a small retrospective cohort \cite{nie2016rectal}, coming from a single institution, using hundreds or even thousands \cite{forghani2019radiomics} of radiomics features extracted from each patient to construct the radiomics signatures possibly using the same dataset after splitting to address multiple target variables, ignoring essential concepts like multiple comparisons corrections and type I errors \cite{chalkidou2015false}. Given the fact that medical imaging technology is fast evolving, such study designs are resulting in clinically non-usable models. They, therefore, might contribute to further confusion and lack of trust in AI models. To make things worse, clinical user requirements in the design process of such new technologies are often neglected \cite{recht2020integrating} which means that the user experience is still not considered as a critical development variable not least due to the fact that there is still a lack of a generalised consensus of how to design effective end-user interaction with machine learning systems \cite{amershi2011effective}. This problem introduces a serious risk of losing early adopters and increasing the cost and the complexity of the product when trying to resolve usability a posteriori. 

There are very few available sources to elaborate on usability issues with respect to AI. For this reason, we start by analysing the basic pillars of usability and enriching them with AI-specific considerations. Based on a domain agnostic usability textbook \cite{wilson2009user}, usability is traditionally associated with five important attributes:

\begin{enumerate}[label=\roman*]
    \item \textit{Learnability}: 
    How quickly a new user learns how to use an AI system is critical for the fast adoption of AI since the users need to rapidly learn and incorporate the models in their clinical workflow. An important element regarding this usability dimension is the ability of the user interface to allow exploratory learning by including e.g. undo functions or comprehensive wizards for novice users. 
    \item \textit{Efficiency}: 
    Clinicians usually face heavy workloads and one important aspect of AI in medical imaging is to empower the user (e.g., radiologist) by alleviating his/her heavy workload (e.g. reading of mammograms) while providing efficiency coupled with high level of productivity. This is an important factor when validating such systems in order to ensure that the inclusion of the AI system preserves or improves diagnostic time while increasing performance and efficiency. 
    \item \textit{Memorability}: 
    This is a very particular property of AI systems characterising how easy it is to remember how to use a particular tool.  Several users in the clinical environment deal with different tools and technologies and being able to easily remember how the system works after some period of not having used it without having to learn everything all over again, is very important for promoting acceptability. This attribute is also an important design element for taking into consideration the casual user needs of AI systems.
    \item \textit{Limited and non-catastrophic errors}: 
    An AI medical imaging system should produce very limited (if any) errors and in any case warn the user about the possibility of a critical error (e.g. in prediction) based on uncertainty estimates. As it has been mentioned before, training and testing of current AI models in relatively small datasets might give rise to errors during the actual deployment of the systems within the clinical setting and this remains a poorly addressed issue calling for more extensive research and definition of widely acceptable guidelines. 
    \item \textit{Satisfaction}:
    It is very often the case that AI developers believe that a satisfactory performance is equal to user satisfaction, assuming that all the clinical users want is high accuracy and robustness. In practice, the system should be pleasant to use and subjective satisfaction needs to be monitored in order to assist the users to incorporate it in their daily routine and ensure wide adoption. The latter also depends on the variability of clinical user attributes (e.g. computer enthusiasts vs more conservatives) and it is therefore very important to engage diverse clinical users in the early design phases as well as in measuring subjective satisfaction after deployment. It is also argued that limited explainability of current AI models might negatively influence user satisfaction.
\end{enumerate}

As mentioned in the beginning usability is a milestone for achieving the goals of an AI system with efficiency and user satisfaction at the same time. So, it is becoming evident that for an AI model to be usable, superior effectiveness and efficiency compared to the current modus operandi in the clinical setting needs to be demonstrated and, most importantly, end-users must be satisfied. The latter might refer to different stakeholders, including the individual patient, the caregiver, and the hospital administrator. They might be interested in the sustainability of the health care services provided by their institution or even to policymakers to craft data-informed decisions and clinical guidelines.

Poor model usability might be responsible for limited translation of research in the field of AI and Radiomics to the clinics. In a recent article discussing the plethora of AI tools in COVID-19 medical imaging, it is argued that the poor clinical adoption of such systems maybe partially attributed to the lack of awareness/ and understanding of the user needs and propose an improved workflow in the AI development process including iterative usability studies \cite{born2021role}. In order to properly design AI projects with a purely clinical indent, active engagement of experts is necessary. AI projects must be supported by a multi-disciplinary team where each and every member should provide his or her own expertise, however the leader should be an individual that has a deep understanding both of the clinical and technical domain.

One of the reasons that can explain the limited translation rate to the clinics is the level of involvement of stakeholders with domain knowledge that was minimal or very superficial, failing to safeguard aspects related to whether such efforts are clinically relevant and usable. The latter contributes to the lack of trust and, therefore, lack of translation to the clinical environment since usability and trust are interconnected. Without engaging domain experts in the model's design, there is always the risk to make assumptions that are clinically irrelevant. Therefore, the developed model will be clinically irrelevant, as well.

Usability lies at the tip of an iceberg, and many qualities that a model should have might influence its usability. Availability of high-quality curated data is the cornerstone of any ML model, but their presence is not enough. ML modelling strategies should respect best practices and avoid data leakage, which might be responsible for overfitted models with limited generalisability. So far, the focus when developing an ML model was to maximise its performance, often implementing models with limited interpretability, the so-called "black-box" models. However, such a strategy may harm the trust and usability of the model since its decisions cannot be explained, and even in the case that the end-user is willing to accept the predictions, regulatory and legal reasons may prohibit the clinical applications of such models, especially in Europe where GDPR grants the right to patients to get explanations and justifications on which grounds a specific decision affecting their life was made by a model.

When designing the validation of an AI model, it is often the case that tailored metrics reflecting the relative cost of a false positive or false negative prediction should be utilised instead of the "of the selves" metrics that are used in most cases. For example, the metric on which the evaluation of the performance of an ML model that is expected to detect cancers in a screening setting is very different to the metric that should be used when designing a model to help with treatment decisions. Even the optimisation of hyperparameters often must be performed on such custom metrics rather than on typical accuracy, AUC or F1 scores.

In Radiomics, most of the efforts concern the development and validation, overlooking how and where the signature will be deployed and used. Consequently, essential concepts of integration with current, mostly rigid, clinical workflows should be considered even from the first phases of model development to avoid unpleasant surprises regarding data availability prospectively. There are many challenges related to low interoperability, fragmentation of clinical systems, restrictions to the easy flow of data between different systems, presence of unstructured information that needs to be transformed and curated to become eligible for further utilisation. As far as the place where the ML models will be deployed and consumed, those can be either in the cloud or in local servers as endpoints within the hospital's firewall. 

According to Park et al \cite{park2020evaluating}, usability evaluation should assess the extent to which the end-users are able to discover, understand, and use system features. An important field to quantify and more objectively assess the latter requirements is called usability testing and it comprise simulation studies or scenario-based testing that run realistic clinical scenarios. During these tests data pipeline integrity and data flowing are checked, as well as potential disagreement between the human decision and the AI model output is investigated through root-cause analysis to identify potential reasons including software/hardware malfunction, poor model fit, bias or finally human error.
\\
\\
\textit{Recommendations for Usability:}

\begin{itemize}
    \item \textit{Engage all the stakeholders in the development phase}: Active engagement of a multi-disciplinary team including AI developers, Medical Imaging Scientists, Radiologists, Radiation Oncologists, Patients and Health Care Administrators to ensure clinical usability, effectiveness, and cost-benefit of the proposed AI models that can be the basis to further convince clinical users and regulatory bodies on the usability and value of the AI solution in healthcare practice. This engagement should be achieved as early as possible and in any case cover all the product life cycle phases starting even from the design of the AI product. Hands-on sessions during the model design and evaluation phases with a multidisciplinary team are essential to ensure that all user perspectives are taken into consideration. 
    \item \textit{Understanding user needs}: It is important to understand in-depth the clinical user needs regarding AI in medical imaging solutions. In particular, to ensure that the solution has a favourable learning curve, it is efficient in terms of time and performance (e.g. for a diagnostic task) and reduce as much as possible the possibility of serious errors. Understanding the clinical needs in a holistic fashion is a milestone for promoting user-centric design while addressing the actual clinical unmet needs. To this end, we propose at least one co-creation development workshop with end users and developers prior to the AI development phase to understand the needs of the users and define relevant constrains and reliability/evaluation metrics metrics. 
    \item \textit{User Interface Design}: The absence of adequate human-computer interfaces is a major obstacle for the adoption of ML models into clinical practice \cite{osuala2019bringing} and understanding user needs is a sine qua non condition to design such adequate human-computer interfaces. It is therefore highly recommended to actively engage radiologists in the design of the UI in order to ensure that the users will be able to use all the provided functionality in an efficient way fulfilling all their needs e.g. in terms of execution time.
    \item \textit{Explainability for usability}: During the design phase it is critical to follow methodologies that produce more explainable results in order to increase trustworthiness and promote clinical adoption.
    \item \textit{Usability testing}: In parallel, as it is has been recently suggested \cite{born2021role}, usability studies should be performed before the final AI solution is released in order to avoid poor user satisfaction and promote the faster and wider adoption. This means that there should be enough time and resources planned to re-evaluate and re-design certain aspects of the product’s functionality until the user needs are met. Usability testing should be done with multiple clinicians of different characteristics (experience) to identify differences and varying preferences. Even though the assessment of AI model usability should be performed in clinical environments with real world data, engaging with end-users in the early stages of designing solutions, and keeping them involved as the solution evolves, is the recommended strategy to gain an understanding of the nature of the problem to be addressed and the issues that emerge during implementation. End-users can also act as great advocates for solutions in their organisations and among their colleagues, which can greatly improve adoption rates \cite{filice2020case}.
    \item \textit{In-silico usability validation}: In order to accelerate usability evaluation, it is also recommended to re-use existing retrospective data in a prospective fashion which blinds the researcher to the outcomes, simulating “real” clinical conditions. It is recommended that at least three radiologists from different clinical sites and degree of experience participate in the study. The validation results should include both usability aspects (e.g. user satisfaction) and agreement metrics between the clinician’s decision and the AI’s recommendation.
    \item \textit{Usability metrics}: It is recommended to define the metrics that will be used throughout the product development. These should include usability questionnaires that measures several usability aspects, including time required to perform the task, learnability, efficiency, explainability, user satisfaction and intention-to-use. It is very important to incorporate in these metrics widely accepted recommendations such as the FDA guidance document for medical device manufacturers on human factors and usability engineering \cite{fda2016}.
    \item \textit{Deployment and integration in the clinical environment}: AI developers should have a clear strategy on how to seamlessly integrate the developed models into current clinical workflows, including Electronic Health Records and PACS systems. Usability should also be assessed in terms of the integrative functional aspects (e.g. when using an AI tools integrated with the hospital’s PACS system).
    \item \textit{Provision of training resources for end-users}: Training resources such as user guides, training material and user workshops can help to reduce the perceived complexity of the AI solution for end-users. This allows end-users to adopt the AI solution with less effort into their clinical practice. For instance, a user guide can resolve doubts on how to install, calibrate, update, interact with, and interpret the AI solution and its results.
    \item \textit{Continuous monitoring of user satisfaction}: AI Model development should be a continuous process influenced by real world conditions that can only be identified after model deployment to address real time changes that occurs in the input data (images and radiomics features), as well as in the output variables. This will promote error reduction, generalisability and trust. 
\end{itemize}

\section{Robustness – For Reliable AI in Medical Imaging} \label{sec:robustness}
In FUTURE-AI, the \textit{Robustness} principle refers to the ability of an imaging AI technology to maintain its model accuracy when it is applied under highly variable conditions in the real world, outside the controlled environment of the laboratory where the algorithm is built. Compared to other types of biomedical data, medical images are known to be associated with significant variations (both expected and unexpected) across radiological studies, which can impact the performance of the AI algorithms. There are several sources for this data heterogeneity, which, alongside their causal reasons and relationships \cite{castro2020causality}, need to be closely taken into consideration when developing, evaluating, and deploying new imaging AI algorithms for the real-world, including:

\begin{enumerate}[label=\roman*]
    \item \textit{Equipment-related heterogeneity}: 
    For a given imaging modality (e.g., MRI), there are multiple manufacturers of imaging scanners (e.g., for MRI, Philips, Siemens, General Electric, Toshiba, Canon, Fujifilm). While the main physical principles that govern the manufacturing of these imaging scanners are consistent, there are vendor-specific variations that can make imaging studies vary in image conditions between scanners. The recent M\&Ms study on deep learning based cardiac image segmentation in a multi-centre and multi-vendor context showed that models trained with cardiac MRI images from a given vendor generalise poorly to new images from a distinct vendor, losing up-to 40\% of their initial performance \cite{campello2021multi}. An additional study based on image data from multiple centres and multiple vendors for the classification of prostate cancer showed that radiomics models that have a decent performance when tested on data from the same centre and/or scanner (AUC of 0.75) may show a significant drop in performance when applied to external data (AUC of 0.54) \cite{castillo2021multi}.
    \item \textit{Centre-specific imaging parameters}: Despite the existence of reference imaging protocols in the clinical literature, the specific parameters of the image acquisitions, such as image resolution, slice thickness, orientation, contrast type, and post-injection scan delays, generally vary between clinical centres. This can lead to imaging studies with different intensity distributions that can impact the robustness of the AI algorithms when applied in new clinical centres. The evaluation of the influence of MRI scanning parameters on quantitative imaging features showed significant differences in many texture features when varying different MRI acquisition parameters such as magnet strength, flip-angle, number of excitations and scanner platform, emphasising the need for a standardised MRI technique \cite{buch2018quantitative}.
    \item \textit{Operator-related heterogeneity}: Imaging studies can greatly vary between shift acquisitions, in image quality, scan positioning, level of noise and artefacts, as well as organ coverage and tissue/lesion appearance, depending on the operator’s experience, dexterity and workload. This variability is particularly pronounced for certain imaging modalities, such as ultrasound, where the operator is required to carefully and precisely manipulate a probe on the patient’s body to identify optimal image planes for subsequent image quantification. In MR images, local intensity shift artefacts can be minimised but not eliminated with optimal patient location, coil design and tuning. Improper coil or patient positioning can produce subtle or, in some cases, severe signal intensity artefacts, which can also occur in a perfectly functioning coil if protocols are not optimised. Improper coil tuning manifests as a shading artefacts that can mimic other findings. Operators are recommended to be familiar with the various causes of signal intensity artefacts to maintain optimal image quality as part of an MR imaging quality assurance program \cite{jones2000signal}.
    \item \textit{Patient-related heterogeneity}: The anatomical properties of the patients, such as the size of the organs of interest, the amount of body fat, anatomical variations, or the tissue density (e.g., brain volumetry related to age), can result in highly variable quality between imaging studies. Large differences between subject variability were observed in quantitative cerebral blood flow measurements in normal subjects using various PET and MRI techniques \cite{henriksen2012estimation}. Another source of image heterogeneity is the level of cooperation of the patients during scanning, such as their propensity to remain still or to move during the image acquisitions. This can be observed when using MRI in paediatric patients, whose major challenge is the need for sedation or general anaesthesia \cite{thukral2015problems}.
    \item \textit{Context-related heterogeneity}: AI algorithms are particularly sensitive to unexpected changes and artefacts in medical image data depending on the context in which the scanning took place. For example, an AI system trained to detect pulmonary lesions on chest X-ray images may be impacted if the X-ray technician forgets to remove adhesive ECG lead connectors on the patient’s chest from a recent inpatient ECG, or if the patients inadvertently place their hands on the chest during the X-ray scan \cite{yu2019framing}.
    \item \textit{Variability in image annotation and segmentation}: Some imaging AI algorithms require the prior definition of regions of interest in the image such as bounding boxes around lesions or contours around tissue/organ boundaries. However, it is well-known that radiologists and clinicians annotate and delineate the images with significant intra- and inter-observer variability, especially when they have different levels of time, expertise, and experience. With annotations used for AI training, this affects the robustness of subsequent AI-driven predictions, especially in feature-based models. A study in breast MRI found that the variability in lesion segmentation based on four different observers resulted in only 20-30\% of robust radiomics features for complex tumours \cite{granzier2020mri}. Automated or semi-automated techniques for medical image segmentation are expected to generate more consistent results, but most existing segmentation software still requires manual inputs and corrections.
\end{enumerate}

Because these variations are an integral part of real-world radiology, and given the differences in clinical practices between radiology departments within as well as across centres and countries, it is important to implement preventive and corrective measures to enhance the robustness of the AI algorithms against changing imaging conditions. In the following, we recommend respective guidelines.
\\
\\
\textit{Recommendations for Robustness:}

\begin{itemize}
    \item \textit{Image harmonisation}: If differences in imaging and acquisition protocols cannot be prevented between centres, robustness should be enhanced by implementing image harmonisation tools and techniques such as histogram normalisation and discretisation \cite{mollura2020artificial}, ComBat harmonisation \cite{fortin2017harmonization, radua2020increased}, and data augmentation solutions with neural style transfer methods, Generative Adversarial Networks and unsupervised image-to-image translation units \cite{gao2019universal, tor2020unsupervised, osuala2021review}. It is recommended to assess and report the variation across features alongside the reduction in variation after applying harmonisation methods to these features in the dataset.
    \item \textit{Feature harmonisation}: Following the craze for radiomics features, their lack of reliability raised the question of the generalisability of classification models. The design of feature harmonisation pipelines is essential to investigate the repeatability and reproducibility of these features in order to evaluate their temporal stability with respect to a controlled scenario (test–retest), as well as their dependence on acquisition parameters such as slice thickness, or tube current \cite{jha2021repeatability}. Feature selection strategies have been proposed to incorporate only robust and stable imaging features into prognostication/prediction models to improve generalisability across multiple institutions \cite{park2019reproducibility}.
    \item \textit{Intra and inter-observer variability}: Dedicated experiments should be performed to separately assess the effect of intra- and inter-observer manual biases on the imaging AI algorithms. One approach is to gather and analyse multiple annotations per image annotated by a diverse set of clinicians and by the same clinician at different points in time.  In supervised machine learning and annotation-related tasks, a common practice to generate ground truth label data is to merge observer annotations. A detailed study was performed to analyse how the high intra- and inter-observer variability resulting from factors such as image quality, different levels of user expertise and domain knowledge may affect the performance of automated image segmentation solutions and their uncertainty. The results highlighted the large impact of intra- and inter-observer variability and the negative effect of annotation merging methods applied in deep learning to obtain reliable estimates of segmentation uncertainty \cite{jungo2018effect}. Samples with high annotation variation above a certain threshold can be detected by  variation proxy measures such as the coefficient of variance (CV), the Dice Similarity Coefficient (DSC) and the Hausdorff Distance (HD). Reporting the analysed observer variation increases study reproducibility, while clinician reassessment of these high variation data samples can allow to find the most suitable and robust consensus annotation in each such case.
    \item \textit{Quality control}: Quality control capabilities should be implemented to identify abnormal deviations or artefacts in the imaging studies. Inter- and intra-observer variability of manual quality control is high and may lead to inclusion of poor-quality scans and exclusion of scans of usable quality. A recent quantitative quality control tool (MRQy) based on unsupervised learning techniques (clustering) was developed to help interrogate MR imaging datasets for site- or scanner-specific variations in image resolution or image contrast, and imaging artefacts such as noise or inhomogeneity; which need correction prior to model development \cite{sadri2020mrqy}. Other relevant solutions include: Qoala-T, a brain-only tool based on an easy and free to use supervised-learning model to reduce observer bias and misclassification in manual quality control procedures using FreeSurfer-processed scans \cite{klapwijk2019qoala}; QC-Automator, based on different CNN architectures and applied on diffusion MR imaging data only, it can handle a variety of artefacts such as motion, multiband interleaving, ghosting, susceptibility, herringbone, and chemical shifts \cite{samani2020qc}; and PI-QUAL, a prostate-specific tool to assess the diagnostic quality of a scan against a set of objective criteria as per Prostate Imaging-Reporting and Data System recommendations, together with criteria obtained from the image \cite{giganti2021understanding}.
    \item \textit{Phantoms}: Phantoms should be scanned across multiple centres and used to calibrate and harmonise future patient images and measurements. The use of a standardised quantitative calibration phantom and a well-recognised and accepted procedure by the medical imaging and radiology community would decrease inter-scanner variability \cite{keenan2018quantitative, prohl2019reproducibility}. Some existing methods make use of an imaging phantom by which volume change can be applied in a highly controlled way for standardising measurements of brain atrophy rates between different scanners \cite{amiri2019novel}.
    \item \textit{Data augmentation for model training}: Robustness in training can be improved through synthetic data by simulating a wide range of challenging imaging conditions (e.g., noise, artefacts, extreme cases) to augment the available data in magnitude and dispersity. A recent study defined two families of data augmentations: spatial transformations to increase sample size through rotation, flipping, scaling or deformation of the original images; and intensity-driven techniques, which maintain the spatial configuration of the anatomical structures but modify their image appearance (e.g., with standard image transformations such as histogram matching, blurring, change in brightness, gamma and contrast, or addition of Gaussian noise; and advanced image synthesis by using generative adversarial networks (GANs) or variational auto-encoders (VAE))\cite{campello2021multi}. An additional study proposed the use of adversarial attacks to generate small synthetic image perturbations for image reconstruction tasks. By introducing robust training into a reconstruction network, the rate of false negative features in image reconstruction was shown to be reduced \cite{caliva2020adversarial}. In the case of the necessity of deployment to clinical domains where the data distribution is unknown a priori, inference/test time adaptive AI models \cite{karani2021test} can adapt to the new data distribution and, hence, further foster AI robustness for the unseen clinical domain.
    \item \textit{Training on heterogeneous data}: The imaging AI algorithms should be trained and evaluated with heterogeneous datasets from multiple clinical centres, vendors, and protocols. One of the most recommended strategies to promote further research and scientific benchmarking in the field of generalisable deep learning is the organisation of task-specific challenges which include multi-centre, multi-vendor and multi-disease data \cite{litjens2014evaluation}. If access to shared anonymised multi-centre imaging samples is not feasible, privacy-preserving federated learning should be considered by the AI developers in collaboration with the participating clinical centres \cite{kaissis2020secure}.
    \item \textit{Uncertainty estimation}: Uncertainty estimation must be considered part of any AI system in imaging, to estimate confidence scores or maps given the imaging characteristics and inform the radiologist in potential lack of robustness \cite{cerda2020confidence}. An entire framework based on Bayesian CNNs was proposed to diagnose ischemic stroke patients incorporating Bayesian uncertainty into the analysis procedure, which resulted in not only an improvement at image-level prediction and uncertainty estimation but also for the detection of uncertain aggregations at the patient-level \cite{herzog2020integrating}. Other innovative proposals include the calculation of a new score beyond the classifier's discriminant or confidence score, called the trust score, which constitutes a measure of uncertainty for any trained (possibly black-box) classifier which is more effective than the classifier's own implied confidence (e.g., softmax probability for a neural network) \cite{jiang2018trust}. 
    \item \textit{Equity in accessibility}: If the AI tool is intended for global radiology, it should be optimised (such as by using transfer learning and/or domain adaptation) and tested with new imaging samples from resource-limited settings in low-to-middle income countries \cite{mollura2020artificial}.
\end{itemize}

\section{Explainability – For Enhanced Understanding of AI in Medical Imaging} \label{sec:explainability}

In recent years, AI models have begun to outperform radiologists at certain diagnostic tasks using medical images \cite{mckinney2020international, tschandl2019comparison}. However, AI solutions in general, and deep neural networks in particular, lack transparency, leading to the term “black box AI”, referring to the fact that these models learn complex functions that are inaccessible and often incomprehensible to humans \cite{yang2021unbox}. One promising, but not yet widely applied, exception to this are causal models that enable comparisons between observed and counterfactual medical imaging data, which help to explain outcomes causally \cite{pawlowski2020deep}. Nonetheless, the common lack of AI model transparency hinders the incorporation of AI solutions in standard-of-care clinical workflows, as clinicians likely cannot accept AI solutions in their workflows without some understanding of the underlying principles, even if the algorithms routinely outperform experts \cite{lipton2017doctor}. For example, it is important to ensure that AI solution performs the diagnosis based on the patient’s phenotype rather than on image features that are clinically irrelevant for the task, such as the presence of a ruler in the image. A recent study showed that a highly accurate deep learning solution for COVID-19 detection from chest radiographs performs prediction based on confounding variables such as laterality or text markers on radiographs \cite{degrave2021ai}. Similarly, a deep learning solution for skin cancer diagnosis can assign importance to irrelevant image regions such as dark corners on the images and still achieve high performance \cite{young2019deep}. Such solutions perform poorly when tested on unseen (real-world) data from new hospitals despite high accuracy during initial testing. Hence, it is important to partially or entirely understand the decision-making process of the AI solution for troubleshooting these problems.

The European Union’s General Data Protection Regulation (GDPR) specifies a \textit{"right to explanation"} for the patient in Article 22 that makes it legally binding to offer explanation regarding the automated decision-making process \cite{selbst2018meaningful}. It is a legal, ethical, and clinical requirement to focus on the explainability of the AI algorithm in order to integrate its predictions into clinical practice. Explainability affirms and embraces the need for providing insight into the mechanisms behind AI decision making processes thereby allowing for clinical validation and scrutinisation of these decisions.  Towards a definition of explainability, the need for a general agreement on the term explainability of AI algorithms is to be pointed out, as it is often used interchangeably with interpretability \cite{marcinkevivcs2020interpretability,linardatos2021explainable}. Holzinger et al  defines explainability as \textit{"Given a certain audience, explainability refers to the details and reasons a model gives to make its functioning clear or easy to understand"}. Explainable artificial intelligence (XAI) for medical imaging refers to AI solutions that give end-users insight into its functioning.

Explainability of AI solutions should start at the design and requirement gathering process by incorporating the desires, objectives, and challenges of clinicians to understand what type of explanations best suit their needs. These elucidations come in a variety of formats, each addressing different questions. 
Local explanations, for instance, provide reasons behind a particular prediction by the AI model for an individual image while global explanations identify the common characteristics that the model considers important for a particular class. 
Post-hoc explainability methods aim to provide an understanding of how the model works after building the model. Attribution maps or heat-maps are one of the most common types of post-hoc explainability methods that highlight the relevant regions on the input image that the AI model considers important. Examples of attribution methods include Grad-CAM \cite{selvaraju2017grad}, Integrated Gradients \cite{sundararajan2017axiomatic}, and Guided BackProp \cite{springenberg2014striving}.

In a recent study for COVID-19 detection from chest X-rays and CT-Scans, Grad-CAM based attribution maps highlighted the infected area in the lungs showing that the deep learning classifier considers it important for prediction \cite{panwar2020deep}. The attribution maps do not offer any information on how these salient regions influence the decision-making process. Local Interpretable Model-agnostic Explanations (LIME) method evaluates the contribution of a feature to the prediction made by the AI model \cite{ribeiro2016should}. LIME perturbs the features extracted from a medical image to measure its impact on the classifier’s prediction. LIME assumes that every complex model behaves like a linear model locally. A new linear model trained using the generated perturbations and classifier’s output determines the contribution of each feature by approximating the behaviour of the model locally. For example, a study for predicting Isocitrate Dehydrogenase Mutations (IDH) in gliomas using dynamic susceptibility contrast magnetic resonance imaging (DSC–MRI)-based radiomics used LIME for explaining predictions made by a random forest model \cite{manikis2021multicenter}. In this case, LIME analysis revealed that dependence count variance, complexity and normalised grey level non-uniformity as strongest radiomics features for IDH status mutation prediction. 

Shapley additive explanations (SHAP) is an interpretability method derived from game theory, which helps in determining the effect of individual features on the predictions made by the classifier 
\cite{lundberg2017unified}. For instance, SHAP analysis for interpreting the predictions of Support Vector Machine model trained with radiomics features extracted from the MRI of patients with non-metastatic nasopharyngeal carcinoma after intensity modulation radiation therapy revealed that tumour shape sphericity, first-order mean absolute deviation, T stage, and overall stage are important features for predicting disease prognosis \cite{du2019radiomics}.

Concept attribution associates high-level clinical concepts quantitatively to model predictions. Testing with Concept Activations Vectors (TCAV) method provides global explanations by determining the influence of high-level image concepts on the neural network’s internal states \cite{kim2018interpretability}. A study on interpretability of deep learning models for predicting Diabetic Retinopathy (DR) level using a five point grading scale showed that the TCAV identified correct diagnostic concepts for some DR levels \cite{kim2018interpretability}. Micro-aneurysm diagnostic concept was assigned a high TCAV score for the diagnosis of DR level 1 and aneurysm diagnostic concept was assigned a high TCAV score for the diagnosis of DR level 2. The TCAV method was extended to determine the influence of continuous variables such as radiomics features on the neural network layer activations. In a further study, the TCAV method used radiomics features as concepts to determine that nuclei texture is relevant for the detection of tumour tissue in breast lymph node histopathology samples by a deep learning model \cite{graziani2018regression}.

Interpretable models inherently provide explanations along with their predictions or, alternatively, their reasoning process is explainable by design. For instance, Concept Bottleneck Models make the decision-making process interpretable i.e. as in \citet{koh2020concept} by first predicting clinical concepts and then predicting the severity grade based entirely on these clinical concepts. In this case, the Concept Bottleneck Models used 10 clinical concepts describing bone spurs, calcification, etc. to predict the severity of knee osteoarthritis. This also allows clinicians to intervene and change the clinical concepts to observe the effect on the model's prediction.

Prototypical Part Network (ProtoPNet) \citep{chen2018looks} is a deep neural network that is interpretable and performs classification by comparing the features extracted from the input image against class discriminative prototypes. ProtoPNet was utilised for Alzheimer’s disease classification with DenseNet-121 as a feature extractor and the analysis showed that the ProtoPNet provided reasoning for its prediction that can facilitate its adoption in clinical practice \citep{mohammadjafariusing}. Anatomical priors and other domain specific information related to the medical image analysis task can be incorporated in the model to make its predictions interpretable. 

It is a general perception that the performance of the algorithm is inversely proportional to its interpretability. A reason for this perception is that deep neural networks have achieved state-of-the-art performance on many medical imaging tasks while remaining mostly uninterpretable. Both Concept Bottleneck Models and ProtoPNet are interpretable and achieve performance at par with black-box deep learning models, showing that there is, in principle, no need to compromise between performance and interpretability \citep{yang2021unbox}.

Explainability methods are difficult to evaluate because they are subjective, application-specific, and often lacking an available ground truth. However, it is important to ensure that the explainability methods produce explanations that are robust, sensitive, and faithful to the model, the data, and the prediction. A study utilised model parameter randomisation and data randomisation tests to evaluate the sanity of different attribution methods. Model parameter randomisation test evaluates the effect of using randomly initialised model weights and trained model weights on attribution methods, while data randomisation test evaluates the effect of using random data labels and correct labels on attribution methods. These tests revealed that the attribution maps generated by Gradient and Grad-CAM method pass these tests, while some other methods produce inconsistent attribution maps \citep{adebayo2018sanity}.

Explainability methods can reveal diagnostic information that may add additional clinical value for diagnosis. For instance, an explainability study on the diagnosis of skin cancer revealed that the attribution maps generated by Grad-CAM for prediction of pigmented actinic keratosis consider the area outside the lesion also important for the prediction \citep{tschandl2020human}. This observation is consistent with the findings that chronic sun damage is responsible for pigmented actinic keratosis. An increase in diagnostic accuracy was observed when clinicians were told to pay extra-attention to chronic sun damage. 

A study for fetal head circumference estimation from ultrasound images used perturbation analysis and Area Over the Perturbation Curve (AOPC) for the quantitative evaluation of the attribution maps \citep{zhang2020explainability}. The AOPC metric is based on perturbation analysis in an ordered manner by modifying the important regions first to observe the performance decay and a large APOC value corresponds to an informative attribution map. It is also important to determine if clinical end-users and stakeholder groups are satisfied with the content and quality of the explanations.

System Causability Scale (SCS) determines the extent of the utility of explanation for the end-user. For example, clinicians can use SCS to evaluate the quality of different explainability methods \citep{holzinger2020measuring}. Generally, there is a need for a thorough quantitative and qualitative evaluation of explanations. This is particularly the case taking into account that deep learning networks are susceptible to adversarial attacks \citep{goodfellow2014explaining}, which exposes them and their explainability methods to security concerns. A study investigating the effect of small input perturbations on attribution maps generated by DeepLIFT and Integrated Gradients revealed that input images that look similar and predict the same label produced very different attribution maps \citep{ghorbani2019interpretation}. Also, ProtoPNet is prone to corruption by noise and JPEG compression artefacts \citep{hoffmann2021looks}. Therefore, there is also a need to investigate the effect of adversarial perturbations and noise on the explainability outputs to inform choice and design of the respective explainability methods.

In summary, we recommend the following guidelines to enhance interpretability and explainability of AI solutions to foster trust in the predictions made by the AI models.
\\
\\
\textit{Recommendations for Explainability:}

\begin{itemize}
    \item \textit{Clinical requirements on explainability}: Clinicians should be involved early on in the design phase to discuss options, wishes and requirements regarding the explainability of AI models. The different explainability methods should be presented to the clinicians in an intuitive manner to allow for a clear understanding of their usage, advantages, and limitations. A small trial using example explanations of each of the explainability methods is recommended to be conducted in order to determine which methods are considered suitable for the task by the clinical end-users and stakeholders.
    \item \textit{Incorporation of clinical concepts}: If possible, clinical concepts important for the diagnosis of a particular disease should also be consistently annotated. These additional concepts can be utilised to make the model interpretable at an added annotation cost, as exemplified in the following. Explainable capsule networks (X-Cap) can predict lung nodule malignancy by encoding high-level visual object attributes such as sphericity, margin, subtlety, and texture in capsule vectors and perform diagnosis based entirely on these concepts \citep{lalonde2020encoding}. The latent space of a deep neural network can be disentangled to understand the model behaviour using annotated clinical concepts. A study for cardiac resynchronisation therapy response prediction from cine MRI showed that the latent space of a variational autoencoder can be disentangled for interpretability in terms of clinical concepts such as the presence of septal flash by using secondary classifiers \citep{puyol2020interpretable}. Anatomical priors and other domain-specific clinical information for the medical image analysis task should also be utilised to design interpretable deep learning models. A deep learning model for detecting midline shift in MRI images can be designed by exploiting structural knowledge. A two-step approach that first estimates the midline and then predicts midline shift from the generated curve is more interpretable than predicting midline shift directly from the entire MRI scan \citep{pisov2019incorporating}.
    \item \textit{Multiple explanation methods}: Multiple explainability methods that provide complementary explanations should be explored for understanding the decision-making process of the AI model. For example, Gamble et al \citep{gamble2021determining} used attribution maps to provide local explanations for a specific image in an effort to understand the reasoning process of the AI model that predicts breast cancer biomarker status from hematoxylin and eosin-stained images . They also utilised TCAV for providing global explanations to identify the characteristics that influence the decisions of the AI model for a particular class.
    \item \textit{Identifying explainable imaging biomarkers}: In order to increase clinical value, explainability methods should be used to identify imaging features or structures that can serve as an imaging biomarker for diagnosis and prognosis of a disease. The detected imaging biomarkers, if previously known, can increase trust in the AI algorithms. For instance, post-hoc explainability methods identified the outside region of the skin indicating chronic sun damage as a potential biomarker for the diagnosis of pigmented actinic keratosis \citep{tschandl2020human, li2019deep}. The explainability methods can also help in hypothesising new imaging biomarkers. A study utilising Layerwise Relevance Propagation (LRP) \citep{bach2015pixel} attribution method for explaining the deep learning model for the prediction of Estrogen Receptor status from H\&E Images revealed a number of image features such as nuclear and stromal morphology as potential biomarkers \citep{seegerer2020interpretable}.
    \item \textit{Quantitative evaluation of explainability}: Quantitative evaluation of explainability methods should be performed to ensure that the explanations are trustworthy, consistent and robust. Quantitative evaluation metrics such as Area Over the Perturbation Curve (AOPC) score may be used for evaluation \citep{samek2016evaluating}. A study investigated four attribution methods for robustness by training multiple deep learning models for Alzheimer’s disease classification \citep{eitel2019testing}. Attribution sum, attribution density and gain of attribution evaluation metrics were used for quantitative comparison.
    \item \textit{Qualitative evaluation of explainability}: Qualitative evaluation of the generated explanations should be performed with the help of clinicians in order to determine the usefulness of the explainability methods. System Causability Scale (SCS) measures the utility of explanations for the end-users, and clinicians can use it to evaluate the quality of the explanations \citep{Suzuki2019}.
    \item \textit{Robustness of explainability against ad- versarial attacks}: The robustness of explainability methods against adversarial attacks should be assessed and, whenever possible, enhanced by training with adversarial examples. The input images are subjected to small perturbations and noise to determine if the explanations remain consistent. Security is a critical factor in automated decision-making systems for healthcare. A study showed that the input images that are subjected to small perturbations and noise to produce visually indistinguishable images can highlight entirely different regions in the attribution maps generated by DeepLIFT and Integrated Gradients methods \citep{ghorbani2019interpretation}. 
    \item \textit{Explainability in clinical practice}: Application-grounded evaluations \citep{doshi2017towards} involving clinical experts should be performed for evaluating the effect of using the explainability methods in clinical practice. A collaborative human-AI study should be performed in which the doctor performs the clinical task using the AI tool with and without explanations. It is important to identify any resulting bias from the use of explainability methods. For example, one study investigated the impact of deep learning model predictions and corresponding integrated gradients attribution maps on 10 ophthalmologists for the diagnosis of Diabetic Retinopathy, revealing that the use of attribution maps results in over-estimation of normal cases \citep{sayres2019using}. 
\end{itemize}

\section{FUTURE-AI Quality Check} \label{sec:checklist}
To facilitate testing whether and to what extent a given imaging AI solution is compliant with the FUTURE-AI guiding principles, a pragmatic quality check is presented in the following. This quality check consists of a set of 55 practical questions, which, as a whole, summarise the five guiding principles and encapsulate each of their aforementioned recommendations. AI research and development teams may consult the quality check to identify potential improvements in their endeavour of building a trustworthy, deployment-ready medical imaging AI solution. The quality check questionnaire can be found in table \ref{Table:quality-check}. Furthermore, an alternative version of the quality check can also be accessed online\footnote{FUTURE-AI's online AI model quality check: \href{https://future-ai.eu/checklist/}{https://future-ai.eu/checklist/}}.

\begin{table*}[!htbp] 
\centering
\scriptsize
\caption{Overview of \textcolor{mycorrect}{our practical FUTURE-AI medical imaging checklist} organised per principle and ranked according to the AI stages in which the respective checklist item is applicable. The stages consist of (1) clinical conceptualisation, (2) end-user requirement gathering, (3) technical design, (4) data collection and preparation, (5) AI implementation and optimisation, (6) AI evaluation, and (7) AI deployment and monitoring.}\label{Table:quality-check}
\scalebox{1.0}{
\begin{tabular}{{
p{0.03\textwidth}p{0.28\textwidth}
p{0.60\textwidth}}}
    \hline
    \textbf{Stage} &
    \textbf{Title} &
    \textbf{Question(s)} 
    \\
    \hline\hline
    \textbf{Fairness}
    \\
    \hline
    1,2	&
    Multi-disciplinarity	&
    Did you design your AI algorithm with a diverse team of stakeholders? Did you collect requirements from a diverse set of end-users? 
    \\
    \hline
    1,2,4	&
    Definition of fairness	&
    Did you define fairness for your specific imaging application? Did you ask clinicians about hidden sources of data imbalance? 
    \\
    \hline
    4	&
    Metadata labelling	&
    At data collection, did you record key metadata variables on individuals and groups?
    \\
    \hline
    5,6	&
    Estimation of data (im)balance	&
    Did you inspect and ensure the diversity of the training and evaluation data?
    \\
    \hline
    4-6	&
    Multi-centre datasets	&
    Did you train and assess your algorithm for multi-centre imaging samples? Does your algorithm maintain accuracy across radiology units and geographical locations? In particular, is it applicable in centres with reduced imaging quality (e.g. in resource-limited countries)?
    \\
    \hline
    5-7	&
    Transparency of fairness	&
    Did you document the data characteristics, including existing (im)balances?
    \\
    \hline
    6	&
    Fairness evaluation and metrics	&
    Did you thoroughly evaluate the fairness of your AI algorithm? Did you use a suitable dataset and dedicated metrics? Do you have a mechanism for continuous evaluation of your algorithm's fairness?
    \\
    \hline
    5,6	&
    Fairness corrective measures	&
    If you identified sources of bias in the data, did you implement mitigation measures?
    \\
    \hline
    6	&
    Continuous monitoring of fairness	&
    Do you have a mechanism for continuous testing of your algorithm’s fairness over its lifetime?
    \\
    \hline
    6	&
    Information and training on fairness	&
    Did you prepare information and training material for the radiologists and clinicians, to inform on potential biases and maximise fairness during the algorithm’s use?
    \\
    \hline\hline
    \textbf{Universality}
    \\
    \hline
    1,2 &
    Definition of clinical task &
    Did you use a universal definition of the clinical task?
    \\
    \hline
    2,3,5 &
    Software standardisation &
    Did you design and implement the imaging AI solution using proven libraries and framework standards that readily allow for extension and  maintenance? 
    \\
    \hline
    3,4 &
    Image annotation standardisation &
    Did you annotate your dataset in an objective, reproducible and standardised way?
    \\
    \hline
    3,4 &
    Variation of quantified biomarkers &
    Are the methods you used for feature quantification compliant with consensus provided by standards initiatives? 
    \\
    \hline
    6 &
    Evaluation metric selection and reporting &
    Did you use universal, transparent, comparable, and reproducible criteria and metrics for your model's performance assessment?
    \\
    \hline
    6 &
    Reference dataset evaluation &
    Did you evaluate your model on at least one open access benchmark dataset that is representative of your model's task and expected real-world data exposure after deployment?
    \\
    \hline
    6 &
    Reporting standards compliance &
    Did you adhere to a standardised reporting guideline when assessing and communicating the design and findings of your study?
    \\
    \hline\hline
    \textbf{Traceability}
    \\
    \hline
    1-3 &
    Model scope &
    Did you agree with the clinicians/radiologists a precise definition of the model's scope? Did you precisely define the model's intended use, the input imaging modalities, the necessary steps to provide the input to the AI model, the reference ground truth if any, the intended output and the use case scenarios? Did you check for any known limitations of the diagnostic/prognostic problem faced?
    \\
    \hline
    4 &
    Data provenance &
    Did you prepare a complete documentation of the imaging dataset you used? Did you include the relevant DICOM tags? Did you structurally list the related clinical/genomic/pathology data?
    \\
    \hline
    4,5 &
    Data localization and distribution &
    Did you annotate the location of data over the network? Did you analyse dataset statistics with respect to the capability to represent the phenomenon at hand over the various clinical sites? Did you quantify missing values and any gaps or known biases?
    \\
    \hline
    4,5 &
    Data-preparation documentation &
    Did you keep track in a structured manner of the whole pre-processing pipeline of imaging and related data? Did you specify input/output, nature, prerequisites and requirements of your pre-processing and data preparation methods?
    \\
    \hline
    3-5 &
    Specification of clinical references &
    Did you include a clear description of the radiological/clinical standards or biomarkers used as reference? Did you include a complete record of the  segmentation process, if any?
    \\
    \hline
    5 &
    Training recording &
    Did you record the details of the training process? Did you included a careful description of imaging and non-imaging features?
    \\
    \hline
    6 &
    Validation documentation &
    Did you document your validation process and the model selection approach agreed with clinicians?
    \\
    \hline
    5-7 &
    Final model details &
    Did you detail the characteristics of the final model released?
    \\
    \hline
    7 &
    Traceability tool &
    Did you equipped your model with a traceability tool? Did you manage the dynamics of your model? 
    \\
    \hline
    1-7 &
    AI Model passport &
    Did you prepare a full metadata record of all the pieces of information of your model? 
    \\
    \hline
    5-7 &
    Accountability and risk specification &
    Did you make a risk analysis for your model? Did you prepare a tool to keep track of the usage of your model?
    \\
    \hline
    \end{tabular}
}
\end{table*}

\begin{table*}[!htbp] 
\centering
\scriptsize
\ContinuedFloat
\caption{Overview of \textcolor{mycorrect}{our practical FUTURE-AI medical imaging checklist} organised per principle and ranked according to the AI stages in which the respective checklist item is applicable.The stages consist of (1) clinical conceptualisation, (2) end-user requirement gathering, (3) technical design, (4) data collection and preparation, (5) AI implementation and optimisation, (6) AI evaluation, and (7) AI deployment and monitoring. (cont'd)}\label{Table:quality-check}
\scalebox{1.0}{
\begin{tabular}{{
p{0.03\textwidth}p{0.28\textwidth}
p{0.60\textwidth}}}
    \hline
    \textbf{Stage} &
    \textbf{Title} &
    \textbf{Question(s)} 
    \\
    \hline
    \hline
    \textbf{Usability}
    \\
    \hline
    1 &
    User engagement &
    Did you engage users in the design and development of the AI tool?
    \\
    \hline
    2 &
    Requirements definition &
    Did you compile end-user requirements?
    \\
    \hline
    3 &
    User interfaces &
    Did you design appropriate user interfaces?
    \\
    \hline
    5 &
    Usable explainability &
    Did you implement any type of explainability that will be usable and actionable by the radiologist?
    \\
    \hline
    6 &
    Usability testing &
    Did you design an appropriate usability study?
    \\
    \hline
    6 &
    In-silico validation  &
    Did you consider an in-silico validation of usability?
    \\
    \hline
    6 &
    Usability metrics &
    Did you define the appropriate usability metrics for evaluation?
    \\
    \hline
    6,7 &
    Clinical Integration &
    Did you evaluate the usability of your tool after integration in the clinical workflows of the clinical sites? 
    \\
    \hline
    6,7 &
    Training material  &
    Did you provide end-users with resources to learn to adopt and appropriately work with your tool?
    \\
    \hline
    5-7 &
    Usability monitoring &
    Did you implement monitoring mechanisms to assess changes in user needs and re-evaluate the appropriateness of the AI solution though time?
    \\
    \hline\hline
    \textbf{Robustness}
    \\
    \hline
    4 &
    Image harmonisation &
    Did you implement any image harmonisation solutions to account for image heterogeneity?
    \\
    \hline
    4 &
    Feature harmonisation &
    Did you perform any feature harmonisation study before developing your predictive models? Did you assess, minimise, and report the variation across features?
    \\
    \hline
    4 &
    Intra- and inter-observer variability &
    Did you perform any intra- and inter-observer annotation studies?
    \\
    \hline
    4 &
    Quality control &
    Did you use any quality control tools to identify abnormal deviations or artefacts in images?
    \\
    \hline
    4 &
    Phantoms &
    Did you use phantoms to harmonise patient images and/or measurements?
    \\
    \hline
    4,5 &
    Data augmentation for model training &
    Did you use data augmentation techniques to improve training of AI models?
    \\
    \hline
    5,6 &
    Training on heterogeneous data &
    Did you train and evaluate your tools with heterogeneous datasets from multiple clinical centres, vendors, and protocols?
    \\
    \hline
    5,6 &
    Uncertainty estimation &
    Did you report any kind of model uncertainty beyond the classifier's discriminant or confidence score?
    \\
    \hline
    6,7 &
    Equity in accessibility &
    Did you optimise your tool with images from resource-limited settings in low-to-middle countries?
    \\
    \hline\hline
    \textbf{Explainability}
    \\
    \hline
    1,2 &
    Clinical requirements on explainability &
    Did you consult with the clinicians to determine which explainability methods suit them? Did you intuitively present the different explanation methods to the clinicians and did they develop a clear understanding of them?
    \\
    \hline
    1-5 &
    Incorporation of clinical concepts &
    Did you consider using clinical annotations and clinical concepts as parameters of the AI algorithms or neural networks to explicitly introduce a level of clinical interpretability?
    \\
    \hline
    1,3,5 &
    Multiple explanation methods &
    Did you explore multiple and complementary explainability methods? 
    \\
    \hline
    6 &
    Identifying explainable imaging biomarkers &
    To increase clinical value, did you evaluate if the explainability methods enable to identify imaging features or structures that can serve as imaging biomarkers? Did you determine if the identified imaging biomarkers are previously known? 
    \\
    \hline
    6 &
    Quantitative evaluation of explainability &
    Did you use some quantitative evaluation tests to determine if the explanations are robust and trustworthy? 
    \\
    \hline
    6 &
    Qualitative evaluation of explainability &
    Did you perform some qualitative evaluation tests with clinicians? 
    \\
    \hline
    6 &
    Robustness of explainability against adversarial attacks &
    Did you evaluate robustness to adversarial attacks, by assessing if the explanations remain consistent when the input images are subjected to small input perturbations and noise?
    \\
    \hline
    6,7 &
    Explainability in clinical practice &
    Did you evaluate clinical effects of explainability methods by performing a human-AI study, where clinician performs clinical tasks using the AI tool with \& without explanations? Did you identify any resulting bias introduced by explainability methods? 
    \\
    \hline
    \end{tabular}
}
\end{table*}

As we recognise that each project stage in the development of the imaging AI solution comprises its own specific challenges and dynamics, the quality check provides guidance as to which project stages are affected by each of its questions. The successive project stages start with (1) clinical conceptualisation, followed by (2) end-user requirement gathering for  co-creation, (3) technical design and specification, (4) data selection, collection and/or preparation, (5) AI implementation and optimisation, (6) AI evaluation (retrospective, prospective, in-silico), and (7) AI deployment and monitoring. 

\section{Discussion and Conclusion} \label{sec:conclusion}

First and foremost, it is to be noted that the FUTURE-AI principles are a living framework, whereby, after diligent examination, new consensus and emerging best practices can be adopted into the framework. As such, this document represents work in progress and is a living document calling for updates and refinement.

In this work, \textcolor{mycorrect}{based on the FUTURE-AI framework}, we have provided \textcolor{mycorrect}{concrete implementation guidelines including suggestions for specific tools and methodologies for the medical imaging domain based on experiences from five large-scale European AI research projects}. 
\textcolor{mycorrect}{Structuring our work around the FUTURE-AI principles, namely, section (\ref{sec:fairness}) \textit{Fairness}, (\ref{sec:universality}) \textit{Universality}, (\ref{sec:traceability}) \textit{Traceability}, (\ref{sec:usability}) \textit{Usability}, (\ref{sec:robustness}) \textit{Robustness} and (\ref{sec:explainability}) \textit{Explainability}, we have analysed for each principle respective recent medical imaging publications and discussed the necessary features for building FUTURE-ready imaging AI solutions.}

We note that the first principle, \textit{Fairness} (section \ref{sec:fairness}), is highly subjective and requires a clear definition in the context of the imaging AI solution. In this regard, a diverse perspective of a multi-disciplinary team including clinicians, social scientists and ethicist can uncover hidden biases. To develop equitable AI solutions, it is important to label metadata in imaging datasets such as sex, gender, ethnicity, skin colour, socioeconomics, or geography, while, at the same time, ensuring patient privacy preservation. Multi-centre data collection can increase the diversity of datasets. Data imbalances and biases within these datasets can be difficult to identify and estimate, but are crucial for subpopulation fairness analysis, evaluation, reporting, countermeasure planning, and continuous fairness monitoring after deployment.

In \textit{Universality} (section \ref{sec:universality}), the need for universally applicable, interoperable, and standardised imaging AI solutions was elaborated. After (a) a standardised definition of the clinical problem, a principled approach towards solving that clinical problem with AI include (b) standardisation of the software solution for maintainability (e.g., using established libraries and proven frameworks), (c) dataset annotations for objectivity, (c) adherence to standardisation initiatives for reproducible imaging biomarker quantification, (e) standardised evaluation criteria with (f) benchmark evaluation for comparability, and (g) adherence to reporting standards (e.g., TRIPOD-AI) for unambiguous communication of the AI solution. 

In \textit{Traceability} (section \ref{sec:traceability}), the importance of model and data transparency was discussed in order to determine and countermeasure concept drifts and data drifts after model deployment. Also, the model passport was introduced that travels together during the entire AI model lifecycle i.e. from the lab to the clinical centres. The model passport transparently provides clinical and technical stakeholders with updated model metadata, statistics, model scope, model data provenance, and monitoring information. A tracebility tool together with the model passport enable AI accountability and risk awareness, for instance, tracing errors back to the recorded training, validation, and data preparation processes.

Breaking down the concept of \textit{Usability} (section \ref{sec:usability}), we discussed its associated key concepts, which are learnability, efficiency, memorability, limited and non-catastrophic errors, and user satisfaction. It is important to listen to and empower the voice of each group in some form affected  by the AI solution to gather a diverse set of stakeholder needs, expectations and requirements. AI solution explainability, end-user training material, and an adequate human-computer interface allow for user-friendly clinical adoption of the AI solution, further guided by validation via usability tests, in-silico trials, usability metrics, and continuous user satisfaction monitoring before and also after clinical deployment.

The \textit{Robustness} principle (section \ref{sec:robustness}) is particularly important in medical imaging due to the multitude of different sources of variations in medical images. For instance, we discussed variation due to equipment-related imaging heterogeneity, varying centre-specific imaging parameters, operator-related heterogeneity, patient-related heterogeneity, context-related heterogeneity, and intra- and inter-observer image annotation variability. To account and countermeasure against the data and domain shifts resulting from these variations, dedicated experiments and quality controls are necessary to estimate, report, and trace-back imaging variations to their origin. Also, optimisation of both the data (e.g., image, feature, and annotation harmonisation) and the AI tool (e.g., domain-invariance, domain-adaptation, domain-generalisation, uncertainty estimation) enable quantifying and reducing the error of the AI model ascribable to data heterogeneity.

Lastly, in \textit{Explainability} (section \ref{sec:explainability}), we strive to motivate the transition from the development of “black box AI” models towards explainable and interpretable AI. Not only is this transition technically, ethically, and scientifically desirable, but it increasingly is adopted into binding legal frameworks e.g., requiring to offer patients an explanation of automated decision-making processes. After gathering the needs and preferences of automated AI model explanation from clinicians, the annotation of respective clinical concepts and the complementary usage of multiple explainability methods (e.g., attribution maps, Local Interpretable Model-agnostic Explanations Shapley additive explanations Testing with Concept Activations Vectors, Layerwise Relevance Propagation, Explainable Capsule Networks) can improve local (image-level) and global (model-level) explainability, as is to be shown via quantitative (e.g., Area Over the Perturbation Curve) and qualitative (e.g., System Causability Scale) explainability evaluation measures. Furthermore, also the robustness of explainability measures (e.g., against adversarial examples), as well as the effect of using the explainability methods in clinical practice is to be evaluated.

\section*{Acknowledgments}
This project has received funding from the European Union’s Horizon 2020 research and innovation programme under grant agreements No 952103 (EuCanImage), 826494 (PRIMAGE), 952172 (CHAIMELEON), 952179 (INCISIVE), 952159 (ProCancer-I).



\bibliography{elsarticle-template}

\begin{thebibliography}{190}
\providecommand{\natexlab}[1]{#1}
\providecommand{\url}[1]{\texttt{#1}}
\expandafter\ifx\csname urlstyle\endcsname\relax
  \providecommand{\doi}[1]{doi: #1}\else
  \providecommand{\doi}{doi: \begingroup \urlstyle{rm}\Url}\fi

\bibitem[eu0(2020-04-24)]{eu01}
Regulation (eu) 2017/745 of the european parliament and of the council of 5
  april 2017 on medical devices, amending directive 2001/83/ec, regulation (ec)
  no 178/2002 and regulation (ec) no 1223/2009 and repealing council directives
  90/385/eec and 93/42/eec.
\newblock \emph{OJ}, L 117:\penalty0 1--–175, 2020-04-24.
\newblock URL \url{http://data.europa.eu/eli/reg/2017/745/oj}.

\bibitem[eu0(2021-04-21)]{eu02}
{Proposal for a REGULATION OF THE EUROPEAN PARLIAMENT AND OF THE COUNCIL LAYING
  DOWN HARMONISED RULES ON ARTIFICIAL INTELLIGENCE (ARTIFICIAL INTELLIGENCE
  ACT) AND AMENDING CERTAIN UNION LEGISLATIVE ACTS COM/2021/206 final}.
\newblock \emph{OJ}, 2021-04-21.
\newblock URL
  \url{https://eur-lex.europa.eu/legal-content/EN/TXT/?uri=CELEX:52021PC0206}.

\bibitem[Abbasi et~al.(2020)Abbasi, Jabali, Khajouei, and
  Tadayon]{abbasi2020investigating}
Reza Abbasi, Monireh~Sadeqi Jabali, Reza Khajouei, and Hamidreza Tadayon.
\newblock Investigating the satisfaction level of physicians in regards to
  implementing medical picture archiving and communication system (pacs).
\newblock \emph{BMC medical informatics and decision making}, 20\penalty0
  (1):\penalty0 1--8, 2020.

\bibitem[Adebayo et~al.(2018)Adebayo, Gilmer, Muelly, Goodfellow, Hardt, and
  Kim]{adebayo2018sanity}
Julius Adebayo, Justin Gilmer, Michael Muelly, Ian Goodfellow, Moritz Hardt,
  and Been Kim.
\newblock Sanity checks for saliency maps.
\newblock \emph{arXiv preprint arXiv:1810.03292}, 2018.

\bibitem[Alexander et~al.(2020)Alexander, Jiang, Ferreira, and
  Zurkiya]{alexander2020intelligent}
Alan Alexander, Adam Jiang, Cara Ferreira, and Delphine Zurkiya.
\newblock An intelligent future for medical imaging: a market outlook on
  artificial intelligence for medical imaging.
\newblock \emph{Journal of the American College of Radiology}, 17\penalty0
  (1):\penalty0 165--170, 2020.

\bibitem[Alvarez-Jimenez et~al.(2020)Alvarez-Jimenez, Sandino, Prasanna, Gupta,
  Viswanath, and Romero]{alvarez2020identifying}
Charlems Alvarez-Jimenez, Alvaro~A Sandino, Prateek Prasanna, Amit Gupta,
  Satish~E Viswanath, and Eduardo Romero.
\newblock Identifying cross-scale associations between radiomic and pathomic
  signatures of non-small cell lung cancer subtypes: preliminary results.
\newblock \emph{Cancers}, 12\penalty0 (12):\penalty0 3663, 2020.

\bibitem[{American College of Radiology}()]{fda_cleared}
{American College of Radiology}.
\newblock {FDA Cleared AI Algorithms}.
\newblock URL \url{https://models.acrdsi.org/}.

\bibitem[Amershi et~al.(2011)Amershi, Fogarty, Kapoor, and
  Tan]{amershi2011effective}
Saleema Amershi, James Fogarty, Ashish Kapoor, and Desney Tan.
\newblock Effective end-user interaction with machine learning.
\newblock In \emph{Proceedings of the AAAI Conference on Artificial
  Intelligence}, volume~25, 2011.

\bibitem[Amiri et~al.(2019)Amiri, Brouwer, Kuijer, de~Munck, Barkhof, and
  Vrenken]{amiri2019novel}
Houshang Amiri, Iman Brouwer, Joost~PA Kuijer, Jan~C de~Munck, Frederik
  Barkhof, and Hugo Vrenken.
\newblock Novel imaging phantom for accurate and robust measurement of brain
  atrophy rates using clinical mri.
\newblock \emph{NeuroImage: Clinical}, 21:\penalty0 101667, 2019.

\bibitem[{ANSI}(2017)]{ISO_6794475}
{ANSI}.
\newblock Artificial intelligence.
\newblock Standard ISO/IEC TR 29110-1:2016, American National Standards
  Institute, Washington DC, US, 2017.
\newblock URL \url{https://www.iso.org/committee/6794475.html}.

\bibitem[Arnold et~al.(2019)Arnold, Bellamy, Hind, Houde, Mehta,
  Mojsilovi{\'c}, Nair, Ramamurthy, Olteanu, Piorkowski,
  et~al.]{arnold2019factsheets}
Matthew Arnold, Rachel~KE Bellamy, Michael Hind, Stephanie Houde, Sameep Mehta,
  Aleksandra Mojsilovi{\'c}, Ravi Nair, K~Natesan Ramamurthy, Alexandra
  Olteanu, David Piorkowski, et~al.
\newblock Factsheets: Increasing trust in ai services through supplier's
  declarations of conformity.
\newblock \emph{IBM Journal of Research and Development}, 63\penalty0
  (4/5):\penalty0 6--1, 2019.

\bibitem[Bach et~al.(2015)Bach, Binder, Montavon, Klauschen, M{\"u}ller, and
  Samek]{bach2015pixel}
Sebastian Bach, Alexander Binder, Gr{\'e}goire Montavon, Frederick Klauschen,
  Klaus-Robert M{\"u}ller, and Wojciech Samek.
\newblock On pixel-wise explanations for non-linear classifier decisions by
  layer-wise relevance propagation.
\newblock \emph{PloS one}, 10\penalty0 (7):\penalty0 e0130140, 2015.

\bibitem[Banerjee et~al.(2021)Banerjee, Bhimireddy, Burns, Celi, Chen, Correa,
  Dullerud, Ghassemi, Huang, Kuo, et~al.]{banerjee2021reading}
Imon Banerjee, Ananth~Reddy Bhimireddy, John~L Burns, Leo~Anthony Celi,
  Li-Ching Chen, Ramon Correa, Natalie Dullerud, Marzyeh Ghassemi, Shih-Cheng
  Huang, Po-Chih Kuo, et~al.
\newblock Reading race: Ai recognises patient's racial identity in medical
  images.
\newblock \emph{arXiv preprint arXiv:2107.10356}, 2021.

\bibitem[Baracaldo et~al.(2017)Baracaldo, Chen, Ludwig, and
  Safavi]{baracaldo2017mitigating}
Nathalie Baracaldo, Bryant Chen, Heiko Ludwig, and Jaehoon~Amir Safavi.
\newblock Mitigating poisoning attacks on machine learning models: A data
  provenance based approach.
\newblock In \emph{Proceedings of the 10th ACM Workshop on Artificial
  Intelligence and Security}, pages 103--110, 2017.

\bibitem[Barocas et~al.(2017)Barocas, Hardt, and
  Narayanan]{barocas2017fairness}
Solon Barocas, Moritz Hardt, and Arvind Narayanan.
\newblock Fairness in machine learning.
\newblock \emph{Nips tutorial}, 1:\penalty0 2017, 2017.

\bibitem[Bellamy et~al.(2019)Bellamy, Dey, Hind, Hoffman, Houde, Kannan, Lohia,
  Martino, Mehta, Mojsilovi{\'c}, et~al.]{bellamy2019ai}
Rachel~KE Bellamy, Kuntal Dey, Michael Hind, Samuel~C Hoffman, Stephanie Houde,
  Kalapriya Kannan, Pranay Lohia, Jacquelyn Martino, Sameep Mehta, Aleksandra
  Mojsilovi{\'c}, et~al.
\newblock Ai fairness 360: An extensible toolkit for detecting and mitigating
  algorithmic bias.
\newblock \emph{IBM Journal of Research and Development}, 63\penalty0
  (4/5):\penalty0 4--1, 2019.

\bibitem[Born et~al.(2021)Born, Beymer, Rajan, Coy, Mukherjee, Manica,
  Prasanna, Ballah, Guindy, Shaham, et~al.]{born2021role}
Jannis Born, David Beymer, Deepta Rajan, Adam Coy, Vandana~V Mukherjee, Matteo
  Manica, Prasanth Prasanna, Deddeh Ballah, Michal Guindy, Dorith Shaham,
  et~al.
\newblock On the role of artificial intelligence in medical imaging of
  covid-19.
\newblock \emph{Patterns}, 2021.

\bibitem[Buch et~al.(2018)Buch, Kuno, Qureshi, Li, and
  Sakai]{buch2018quantitative}
Karen Buch, Hirofumi Kuno, Muhammad~M Qureshi, Baojun Li, and Osamu Sakai.
\newblock Quantitative variations in texture analysis features dependent on mri
  scanning parameters: A phantom model.
\newblock \emph{Journal of applied clinical medical physics}, 19\penalty0
  (6):\penalty0 253--264, 2018.

\bibitem[B{\"u}cker et~al.(2021)B{\"u}cker, Szepannek, Gosiewska, and
  Biecek]{bucker2021transparency}
Michael B{\"u}cker, Gero Szepannek, Alicja Gosiewska, and Przemyslaw Biecek.
\newblock Transparency, auditability, and explainability of machine learning
  models in credit scoring.
\newblock \emph{Journal of the Operational Research Society}, pages 1--21,
  2021.

\bibitem[Caliv{\'a} et~al.(2020)Caliv{\'a}, Cheng, Shah, and
  Pedoia]{caliva2020adversarial}
Francesco Caliv{\'a}, Kaiyang Cheng, Rutwik Shah, and Valentina Pedoia.
\newblock Adversarial robust training of deep learning mri reconstruction
  models.
\newblock \emph{arXiv preprint arXiv:2011.00070}, 2020.

\bibitem[Campello et~al.(2021)Campello, Gkontra, Izquierdo, Mart{\'\i}n-Isla,
  Sojoudi, Full, Maier-Hein, Zhang, He, Ma, et~al.]{campello2021multi}
V{\'\i}ctor~M Campello, Polyxeni Gkontra, Cristian Izquierdo, Carlos
  Mart{\'\i}n-Isla, Alireza Sojoudi, Peter~M Full, Klaus Maier-Hein, Yao Zhang,
  Zhiqiang He, Jun Ma, et~al.
\newblock Multi-centre, multi-vendor and multi-disease cardiac segmentation:
  The m\&ms challenge.
\newblock \emph{IEEE Transactions on Medical Imaging}, 2021.

\bibitem[Casado et~al.(2021)Casado, Lema, Criado, Iglesias, Regueiro, and
  Barro]{casado2021concept}
Fernando~E Casado, Dylan Lema, Marcos~F Criado, Roberto Iglesias, Carlos~V
  Regueiro, and Sen{\'e}n Barro.
\newblock Concept drift detection and adaptation for federated and continual
  learning.
\newblock \emph{arXiv preprint arXiv:2105.13309}, 2021.

\bibitem[Castillo et~al.(2021)Castillo, Starmans, Arif, Niessen, Klein, Bangma,
  Schoots, and Veenland]{castillo2021multi}
T~JM Castillo, MPA Starmans, M~Arif, WJ~Niessen, Stefan Klein, Chris~H Bangma,
  Ivo~G Schoots, and JF~Veenland.
\newblock A multi-center, multi-vendor study to evaluate the generalizability
  of a radiomics model for classifying prostate cancer: High grade vs. low
  grade.
\newblock \emph{Diagnostics (Basel, Switzerland)}, 11\penalty0 (2), 2021.

\bibitem[Castro et~al.(2020)Castro, Walker, and Glocker]{castro2020causality}
Daniel~C Castro, Ian Walker, and Ben Glocker.
\newblock Causality matters in medical imaging.
\newblock \emph{Nature Communications}, 11\penalty0 (1):\penalty0 1--10, 2020.

\bibitem[Cerd{\'a}~Alberich et~al.(2020)Cerd{\'a}~Alberich, Sang{\"u}esa~Nebot,
  Alberich-Bayarri, Carot~Sierra, Mart{\'\i}nez de~las Heras, Veiga~Canuto,
  Ca{\~n}ete, and Mart{\'\i}-Bonmat{\'\i}]{cerda2020confidence}
Leonor Cerd{\'a}~Alberich, Cinta Sang{\"u}esa~Nebot, Angel Alberich-Bayarri,
  Jos{\'e}~Miguel Carot~Sierra, Blanca Mart{\'\i}nez de~las Heras, Diana
  Veiga~Canuto, Adela Ca{\~n}ete, and Luis Mart{\'\i}-Bonmat{\'\i}.
\newblock A confidence habitats methodology in mr quantitative diffusion for
  the classification of neuroblastic tumors.
\newblock \emph{Cancers}, 12\penalty0 (12):\penalty0 3858, 2020.

\bibitem[Chalkidou et~al.(2015)Chalkidou, O’Doherty, and
  Marsden]{chalkidou2015false}
Anastasia Chalkidou, Michael~J O’Doherty, and Paul~K Marsden.
\newblock False discovery rates in pet and ct studies with texture features: a
  systematic review.
\newblock \emph{PloS one}, 10\penalty0 (5):\penalty0 e0124165, 2015.

\bibitem[Chen et~al.(2018)Chen, Li, Tao, Barnett, Su, and Rudin]{chen2018looks}
Chaofan Chen, Oscar Li, Chaofan Tao, Alina~Jade Barnett, Jonathan Su, and
  Cynthia Rudin.
\newblock This looks like that: deep learning for interpretable image
  recognition.
\newblock \emph{arXiv preprint arXiv:1806.10574}, 2018.

\bibitem[Clark et~al.(2013)Clark, Vendt, Smith, Freymann, Kirby, Koppel, Moore,
  Phillips, Maffitt, Pringle, et~al.]{clark2013cancer}
Kenneth Clark, Bruce Vendt, Kirk Smith, John Freymann, Justin Kirby, Paul
  Koppel, Stephen Moore, Stanley Phillips, David Maffitt, Michael Pringle,
  et~al.
\newblock {The Cancer Imaging Archive (TCIA): maintaining and operating a
  public information repository}.
\newblock \emph{Journal of digital imaging}, 26\penalty0 (6):\penalty0
  1045--1057, 2013.

\bibitem[Collins and Moons(2019)]{collins2019reporting}
Gary~S Collins and Karel~GM Moons.
\newblock Reporting of artificial intelligence prediction models.
\newblock \emph{The Lancet}, 393\penalty0 (10181):\penalty0 1577--1579, 2019.

\bibitem[Collins et~al.(2015)Collins, Reitsma, Altman, and
  Moons]{collins2015transparent}
Gary~S Collins, Johannes~B Reitsma, Douglas~G Altman, and Karel~GM Moons.
\newblock Transparent reporting of a multivariable prediction model for
  individual prognosis or diagnosis (tripod): the tripod statement.
\newblock \emph{Journal of British Surgery}, 102\penalty0 (3):\penalty0
  148--158, 2015.

\bibitem[DeGrave et~al.(2021)DeGrave, Janizek, and Lee]{degrave2021ai}
Alex~J DeGrave, Joseph~D Janizek, and Su-In Lee.
\newblock Ai for radiographic covid-19 detection selects shortcuts over signal.
\newblock \emph{Nature Machine Intelligence}, pages 1--10, 2021.

\bibitem[Diaz et~al.(2021)Diaz, Kushibar, Osuala, Linardos, Garrucho, Igual,
  Radeva, Prior, Gkontra, and Lekadir]{diaz2021data}
Oliver Diaz, Kaisar Kushibar, Richard Osuala, Akis Linardos, Lidia Garrucho,
  Laura Igual, Petia Radeva, Fred Prior, Polyxeni Gkontra, and Karim Lekadir.
\newblock Data preparation for artificial intelligence in medical imaging: A
  comprehensive guide to open-access platforms and tools.
\newblock \emph{Physica Medica}, 83:\penalty0 25--37, 2021.
\newblock ISSN 1120-1797.
\newblock \doi{10.1016/j.ejmp.2021.02.007}.

\bibitem[Doshi-Velez and Kim(2017)]{doshi2017towards}
Finale Doshi-Velez and Been Kim.
\newblock Towards a rigorous science of interpretable machine learning.
\newblock \emph{arXiv preprint arXiv:1702.08608}, 2017.

\bibitem[Du et~al.(2019)Du, Lee, Yuan, Lam, Pang, Chen, Lam, Khong, Lee, Kwong,
  et~al.]{du2019radiomics}
Richard Du, Victor~H Lee, Hui Yuan, Ka-On Lam, Herbert~H Pang, Yu~Chen,
  Edmund~Y Lam, Pek-Lan Khong, Anne~W Lee, Dora~L Kwong, et~al.
\newblock Radiomics model to predict early progression of nonmetastatic
  nasopharyngeal carcinoma after intensity modulation radiation therapy: a
  multicenter study.
\newblock \emph{Radiology: Artificial Intelligence}, 1\penalty0 (4):\penalty0
  e180075, 2019.

\bibitem[Eitel et~al.(2019)Eitel, Ritter, (ADNI, et~al.]{eitel2019testing}
Fabian Eitel, Kerstin Ritter, Alzheimer’s Disease Neuroimaging~Initiative
  (ADNI, et~al.
\newblock Testing the robustness of attribution methods for convolutional
  neural networks in mri-based alzheimer’s disease classification.
\newblock In \emph{Interpretability of Machine Intelligence in Medical Image
  Computing and Multimodal Learning for Clinical Decision Support}, pages
  3--11. Springer, 2019.

\bibitem[{EU Commission}(2019)]{eucommission2019}
{EU Commission}.
\newblock Ethics guidelines for trustworthy ai. 2019., 2019.
\newblock URL
  \url{https://digital-strategy.ec.europa.eu/en/library/ethics-guidelines-trustworthy-ai}.

\bibitem[Faes et~al.(2019)Faes, Wagner, Fu, Liu, Korot, Ledsam, Back, Chopra,
  Pontikos, Kern, et~al.]{faes2019automated}
Livia Faes, Siegfried~K Wagner, Dun~Jack Fu, Xiaoxuan Liu, Edward Korot,
  Joseph~R Ledsam, Trevor Back, Reena Chopra, Nikolas Pontikos, Christoph Kern,
  et~al.
\newblock Automated deep learning design for medical image classification by
  health-care professionals with no coding experience: a feasibility study.
\newblock \emph{The Lancet Digital Health}, 1\penalty0 (5):\penalty0
  e232--e242, 2019.

\bibitem[Farzandipour et~al.(2021)Farzandipour, Jabali, Nickfarjam, and
  Tadayon]{farzandipour2021usability}
Mehrdad Farzandipour, Monireh~Sadeqi Jabali, Ali~Mohammad Nickfarjam, and
  Hamidreza Tadayon.
\newblock Usability evaluation of selected picture archiving and communication
  systems at the national level: Analysis of users’ viewpoints.
\newblock \emph{International Journal of Medical Informatics}, 147:\penalty0
  104372, 2021.

\bibitem[{FDA}(2016)]{fda2016}
{FDA}.
\newblock {Applying Human Factors and Usability Engineering to Medical
  Devices}, 2016.
\newblock URL \url{https://www.fda.gov/media/80481/download}.

\bibitem[{FDA}(2021)]{fda2021}
{FDA}.
\newblock {Proposed Rregulatory Framework for Modifications to Artificial
  Intelligence/Machine Learning (AI/ML)-Based-Software as a Medical Device
  (SaMD)}, 2021.
\newblock URL \url{https://www.fda.gov/media/122535/download}.

\bibitem[Felzmann et~al.(2019)Felzmann, Fosch-Villaronga, Lutz, and
  Tamo-Larrieux]{felzmann2019robots}
Heike Felzmann, Eduard Fosch-Villaronga, Christoph Lutz, and Aurelia
  Tamo-Larrieux.
\newblock Robots and transparency: The multiple dimensions of transparency in
  the context of robot technologies.
\newblock \emph{IEEE Robotics \& Automation Magazine}, 26\penalty0
  (2):\penalty0 71--78, 2019.

\bibitem[Felzmann et~al.(2020)Felzmann, Fosch-Villaronga, Lutz, and
  Tam{\`o}-Larrieux]{felzmann2020towards}
Heike Felzmann, Eduard Fosch-Villaronga, Christoph Lutz, and Aurelia
  Tam{\`o}-Larrieux.
\newblock Towards transparency by design for artificial intelligence.
\newblock \emph{Science and Engineering Ethics}, 26\penalty0 (6):\penalty0
  3333--3361, 2020.

\bibitem[Filice and Ratwani(2020)]{filice2020case}
Ross~W Filice and Raj~M Ratwani.
\newblock The case for user-centered artificial intelligence in radiology,
  2020.

\bibitem[Forghani et~al.(2019)Forghani, Savadjiev, Chatterjee, Muthukrishnan,
  Reinhold, and Forghani]{forghani2019radiomics}
Reza Forghani, Peter Savadjiev, Avishek Chatterjee, Nikesh Muthukrishnan,
  Caroline Reinhold, and Behzad Forghani.
\newblock Radiomics and artificial intelligence for biomarker and prediction
  model development in oncology.
\newblock \emph{Computational and structural biotechnology journal},
  17:\penalty0 995, 2019.

\bibitem[Fortin et~al.(2017)Fortin, Parker, Tun{\c{c}}, Watanabe, Elliott,
  Ruparel, Roalf, Satterthwaite, Gur, Gur, et~al.]{fortin2017harmonization}
Jean-Philippe Fortin, Drew Parker, Birkan Tun{\c{c}}, Takanori Watanabe, Mark~A
  Elliott, Kosha Ruparel, David~R Roalf, Theodore~D Satterthwaite, Ruben~C Gur,
  Raquel~E Gur, et~al.
\newblock Harmonization of multi-site diffusion tensor imaging data.
\newblock \emph{Neuroimage}, 161:\penalty0 149--170, 2017.

\bibitem[Gamble et~al.(2021)Gamble, Jaroensri, Wang, Tan, Moran, Brown,
  Flament-Auvigne, Rakha, Toss, Dabbs, et~al.]{gamble2021determining}
Paul Gamble, Ronnachai Jaroensri, Hongwu Wang, Fraser Tan, Melissa Moran,
  Trissia Brown, Isabelle Flament-Auvigne, Emad~A Rakha, Michael Toss, David~J
  Dabbs, et~al.
\newblock Determining breast cancer biomarker status and associated
  morphological features using deep learning.
\newblock \emph{Communications Medicine}, 1\penalty0 (1):\penalty0 1--12, 2021.

\bibitem[Gao et~al.(2019)Gao, Liu, Wang, Shi, and Yu]{gao2019universal}
Yuan Gao, Yingchao Liu, Yuanyuan Wang, Zhifeng Shi, and Jinhua Yu.
\newblock A universal intensity standardization method based on a many-to-one
  weak-paired cycle generative adversarial network for magnetic resonance
  images.
\newblock \emph{IEEE transactions on medical imaging}, 38\penalty0
  (9):\penalty0 2059--2069, 2019.

\bibitem[Gebru et~al.(2018)Gebru, Morgenstern, Vecchione, Vaughan, Wallach,
  Daum{\'e}~III, and Crawford]{gebru2018datasheets}
Timnit Gebru, Jamie Morgenstern, Briana Vecchione, Jennifer~Wortman Vaughan,
  Hanna Wallach, Hal Daum{\'e}~III, and Kate Crawford.
\newblock Datasheets for datasets.
\newblock \emph{arXiv preprint arXiv:1803.09010}, 2018.

\bibitem[Geis et~al.(2019)Geis, Brady, Wu, Spencer, Ranschaert, Jaremko,
  Langer, Kitts, Birch, Shields, et~al.]{geis2019ethics}
J~Raymond Geis, Adrian~P Brady, Carol~C Wu, Jack Spencer, Erik Ranschaert,
  Jacob~L Jaremko, Steve~G Langer, Andrea~Borondy Kitts, Judy Birch, William~F
  Shields, et~al.
\newblock Ethics of artificial intelligence in radiology: summary of the joint
  european and north american multisociety statement.
\newblock \emph{Canadian Association of Radiologists Journal}, 70\penalty0
  (4):\penalty0 329--334, 2019.

\bibitem[Geldermann et~al.(2013)Geldermann, Grouls, Kuhl, Deserno, and
  Spreckelsen]{geldermann2013black}
Ina Geldermann, Christoph Grouls, Christiane Kuhl, Thomas~M Deserno, and Cord
  Spreckelsen.
\newblock Black box integration of computer-aided diagnosis into pacs deserves
  a second chance: results of a usability study concerning bone age assessment.
\newblock \emph{Journal of digital imaging}, 26\penalty0 (4):\penalty0
  698--708, 2013.

\bibitem[Gharibi et~al.(2021)Gharibi, Walunj, Nekadi, Marri, and
  Lee]{gharibi2021automated}
Gharib Gharibi, Vijay Walunj, Raju Nekadi, Raj Marri, and Yugyung Lee.
\newblock Automated end-to-end management of the modeling lifecycle in deep
  learning.
\newblock \emph{Empirical Software Engineering}, 26\penalty0 (2):\penalty0
  1--33, 2021.

\bibitem[Ghorbani et~al.(2019)Ghorbani, Abid, and
  Zou]{ghorbani2019interpretation}
Amirata Ghorbani, Abubakar Abid, and James Zou.
\newblock Interpretation of neural networks is fragile.
\newblock In \emph{Proceedings of the AAAI Conference on Artificial
  Intelligence}, volume~33, pages 3681--3688, 2019.

\bibitem[Giganti et~al.(2021)Giganti, Kirkham, Kasivisvanathan, Papoutsaki,
  Punwani, Emberton, Moore, and Allen]{giganti2021understanding}
Francesco Giganti, Alex Kirkham, Veeru Kasivisvanathan, Marianthi-Vasiliki
  Papoutsaki, Shonit Punwani, Mark Emberton, Caroline~M Moore, and Clare Allen.
\newblock Understanding pi-qual for prostate mri quality: a practical primer
  for radiologists.
\newblock \emph{Insights into Imaging}, 12\penalty0 (1):\penalty0 1--19, 2021.

\bibitem[Glover~IV et~al.(2017)Glover~IV, Daye, Khalilzadeh, Pianykh,
  Rosenthal, Brink, and Flores]{glover2017socioeconomic}
McKinley Glover~IV, Dania Daye, Omid Khalilzadeh, Oleg Pianykh, Daniel~I
  Rosenthal, James~A Brink, and Efr{\'e}n~J Flores.
\newblock Socioeconomic and demographic predictors of missed opportunities to
  provide advanced imaging services.
\newblock \emph{Journal of the American College of Radiology}, 14\penalty0
  (11):\penalty0 1403--1411, 2017.

\bibitem[Goodfellow et~al.(2014)Goodfellow, Shlens, and
  Szegedy]{goodfellow2014explaining}
Ian~J Goodfellow, Jonathon Shlens, and Christian Szegedy.
\newblock Explaining and harnessing adversarial examples.
\newblock \emph{arXiv preprint arXiv:1412.6572}, 2014.

\bibitem[Goodman et~al.(2016)Goodman, Fanelli, and Ioannidis]{goodman2016does}
Steven~N Goodman, Daniele Fanelli, and John~PA Ioannidis.
\newblock What does research reproducibility mean?
\newblock \emph{Science translational medicine}, 8\penalty0 (341):\penalty0
  341ps12--341ps12, 2016.

\bibitem[Granzier et~al.(2020)Granzier, Verbakel, Ibrahim, van Timmeren, van
  Nijnatten, Leijenaar, Lobbes, Smidt, and Woodruff]{granzier2020mri}
RWY Granzier, NMH Verbakel, A~Ibrahim, JE~van Timmeren, TJA van Nijnatten, RTH
  Leijenaar, MBI Lobbes, ML~Smidt, and HC~Woodruff.
\newblock Mri-based radiomics in breast cancer: feature robustness with respect
  to inter-observer segmentation variability.
\newblock \emph{Scientific reports}, 10\penalty0 (1):\penalty0 1--11, 2020.

\bibitem[Graziani et~al.(2018)Graziani, Andrearczyk, and
  M{\"u}ller]{graziani2018regression}
Mara Graziani, Vincent Andrearczyk, and Henning M{\"u}ller.
\newblock Regression concept vectors for bidirectional explanations in
  histopathology.
\newblock In \emph{Understanding and Interpreting Machine Learning in Medical
  Image Computing Applications}, pages 124--132. Springer, 2018.

\bibitem[Henriksen et~al.(2012)Henriksen, Larsson, Hansen, Gr{\"u}ner, Law, and
  Rostrup]{henriksen2012estimation}
Otto~M Henriksen, Henrik~BW Larsson, Adam~E Hansen, Julie~M Gr{\"u}ner, Ian
  Law, and Egill Rostrup.
\newblock Estimation of intersubject variability of cerebral blood flow
  measurements using mri and positron emission tomography.
\newblock \emph{Journal of Magnetic Resonance Imaging}, 35\penalty0
  (6):\penalty0 1290--1299, 2012.

\bibitem[Herschel et~al.(2017)Herschel, Diestelk{\"a}mper, and
  Lahmar]{herschel2017survey}
Melanie Herschel, Ralf Diestelk{\"a}mper, and Houssem~Ben Lahmar.
\newblock A survey on provenance: What for? what form? what from?
\newblock \emph{The VLDB Journal}, 26\penalty0 (6):\penalty0 881--906, 2017.

\bibitem[Hertzum(2020)]{hertzum2020usability}
Morten Hertzum.
\newblock Usability testing: A practitioner's guide to evaluating the user
  experience.
\newblock \emph{Synthesis Lectures on Human-Centered Informatics}, 13\penalty0
  (1):\penalty0 i--105, 2020.

\bibitem[Herzog et~al.(2020)Herzog, Murina, D{\"u}rr, Wegener, and
  Sick]{herzog2020integrating}
Lisa Herzog, Elvis Murina, Oliver D{\"u}rr, Susanne Wegener, and Beate Sick.
\newblock Integrating uncertainty in deep neural networks for mri based stroke
  analysis.
\newblock \emph{Medical Image Analysis}, 65:\penalty0 101790, 2020.

\bibitem[Hoffmann et~al.(2021)Hoffmann, Fanconi, Rade, and
  Kohler]{hoffmann2021looks}
Adrian Hoffmann, Claudio Fanconi, Rahul Rade, and Jonas Kohler.
\newblock This looks like that... does it? shortcomings of latent space
  prototype interpretability in deep networks.
\newblock \emph{arXiv preprint arXiv:2105.02968}, 2021.

\bibitem[Holzinger et~al.(2020)Holzinger, Carrington, and
  M{\"u}ller]{holzinger2020measuring}
Andreas Holzinger, Andr{\'e} Carrington, and Heimo M{\"u}ller.
\newblock Measuring the quality of explanations: the system causability scale
  (scs).
\newblock \emph{KI-K{\"u}nstliche Intelligenz}, pages 1--6, 2020.

\bibitem[Horvat et~al.(2018)Horvat, Veeraraghavan, Khan, Blazic, Zheng, Capanu,
  Sala, Garcia-Aguilar, Gollub, and Petkovska]{horvat2018mr}
Natally Horvat, Harini Veeraraghavan, Monika Khan, Ivana Blazic, Junting Zheng,
  Marinela Capanu, Evis Sala, Julio Garcia-Aguilar, Marc~J Gollub, and Iva
  Petkovska.
\newblock Mr imaging of rectal cancer: radiomics analysis to assess treatment
  response after neoadjuvant therapy.
\newblock \emph{Radiology}, 287\penalty0 (3):\penalty0 833--843, 2018.

\bibitem[Hoshino and Yokota(2021)]{hoshino2021radiogenomics}
Isamu Hoshino and Hajime Yokota.
\newblock Radiogenomics of gastroenterological cancer: The dawn of personalized
  medicine with artificial intelligence-based image analysis.
\newblock \emph{Annals of Gastroenterological Surgery}, 2021.

\bibitem[Huang et~al.(2020)Huang, Gong, Wang, Wang, Guo, Cai, Li, Li, Yu, and
  Lin]{huang2020timely}
Guoquan Huang, Tao Gong, Guangbin Wang, Jianwen Wang, Xinfu Guo, Erpeng Cai,
  Shirong Li, Xiaohu Li, Yongqiang Yu, and Liangjie Lin.
\newblock Timely diagnosis and treatment shortens the time to resolution of
  coronavirus disease (covid-19) pneumonia and lowers the highest and last ct
  scores from sequential chest ct.
\newblock \emph{American Journal of Roentgenology}, 215\penalty0 (2):\penalty0
  367--373, 2020.

\bibitem[Hummer et~al.(2019)Hummer, Muthusamy, Rausch, Dube, El~Maghraoui,
  Murthi, and Oum]{hummer2019modelops}
Waldemar Hummer, Vinod Muthusamy, Thomas Rausch, Parijat Dube, Kaoutar
  El~Maghraoui, Anupama Murthi, and Punleuk Oum.
\newblock Modelops: Cloud-based lifecycle management for reliable and trusted
  ai.
\newblock In \emph{2019 IEEE International Conference on Cloud Engineering
  (IC2E)}, pages 113--120. IEEE, 2019.

\bibitem[Irvin et~al.(2019)Irvin, Rajpurkar, Ko, Yu, Ciurea-Ilcus, Chute,
  Marklund, Haghgoo, Ball, Shpanskaya, et~al.]{irvin2019chexpert}
Jeremy Irvin, Pranav Rajpurkar, Michael Ko, Yifan Yu, Silviana Ciurea-Ilcus,
  Chris Chute, Henrik Marklund, Behzad Haghgoo, Robyn Ball, Katie Shpanskaya,
  et~al.
\newblock Chexpert: A large chest radiograph dataset with uncertainty labels
  and expert comparison.
\newblock In \emph{Proceedings of the AAAI conference on artificial
  intelligence}, volume~33, pages 590--597, 2019.

\bibitem[{ISO}(2018)]{ISO_63500}
{ISO}.
\newblock Iso 9241-11:2018. ergonomics of human-system interaction — part 11:
  Usability: Definitions and concepts.
\newblock Standard {ISO} 9241-11:2018, International Organization for
  Standardization, 2018.
\newblock URL \url{https://www.iso.org/standard/63500.html}.

\bibitem[Jameel et~al.(2018)Jameel, Hashmani, Alhussain, and
  Budiman]{jameel2018fully}
Syed~Muslim Jameel, Manzoor~Ahmed Hashmani, Hitham Alhussain, and Arif Budiman.
\newblock A fully adaptive image classification approach for industrial
  revolution 4.0.
\newblock In \emph{International Conference of Reliable Information and
  Communication Technology}, pages 311--321. Springer, 2018.

\bibitem[Jha et~al.(2021)Jha, Mithun, Jaiswar, Sherkhane, Purandare, Prabhash,
  Rangarajan, Dekker, Wee, and Traverso]{jha2021repeatability}
AK~Jha, S~Mithun, V~Jaiswar, UB~Sherkhane, NC~Purandare, K~Prabhash,
  V~Rangarajan, A~Dekker, L~Wee, and A~Traverso.
\newblock Repeatability and reproducibility study of radiomic features on a
  phantom and human cohort.
\newblock \emph{Scientific reports}, 11\penalty0 (1):\penalty0 1--12, 2021.

\bibitem[Jiang et~al.(2018)Jiang, Kim, Guan, and Gupta]{jiang2018trust}
Heinrich Jiang, Been Kim, Melody~Y Guan, and Maya Gupta.
\newblock To trust or not to trust a classifier.
\newblock \emph{arXiv preprint arXiv:1805.11783}, 2018.

\bibitem[Jobin et~al.(2019)Jobin, Ienca, and Vayena]{jobin2019global}
Anna Jobin, Marcello Ienca, and Effy Vayena.
\newblock The global landscape of ai ethics guidelines.
\newblock \emph{Nature Machine Intelligence}, 1\penalty0 (9):\penalty0
  389--399, 2019.

\bibitem[Jones and Witte(2000)]{jones2000signal}
Randall~W Jones and Robert~J Witte.
\newblock Signal intensity artifacts in clinical mr imaging.
\newblock \emph{Radiographics}, 20\penalty0 (3):\penalty0 893--901, 2000.

\bibitem[Jorritsma et~al.(2014)Jorritsma, Cnossen, and van
  Ooijen]{jorritsma2014merits}
Wiard Jorritsma, Fokie Cnossen, and Peter~MA van Ooijen.
\newblock Merits of usability testing for pacs selection.
\newblock \emph{international journal of medical informatics}, 83\penalty0
  (1):\penalty0 27--36, 2014.

\bibitem[Jungo et~al.(2018)Jungo, Meier, Ermis, Blatti-Moreno, Herrmann, Wiest,
  and Reyes]{jungo2018effect}
Alain Jungo, Raphael Meier, Ekin Ermis, Marcela Blatti-Moreno, Evelyn Herrmann,
  Roland Wiest, and Mauricio Reyes.
\newblock On the effect of inter-observer variability for a reliable estimation
  of uncertainty of medical image segmentation.
\newblock In \emph{International Conference on Medical Image Computing and
  Computer-Assisted Intervention}, pages 682--690. Springer, 2018.

\bibitem[Kaissis et~al.(2020)Kaissis, Makowski, R{\"u}ckert, and
  Braren]{kaissis2020secure}
Georgios~A Kaissis, Marcus~R Makowski, Daniel R{\"u}ckert, and Rickmer~F
  Braren.
\newblock Secure, privacy-preserving and federated machine learning in medical
  imaging.
\newblock \emph{Nature Machine Intelligence}, 2\penalty0 (6):\penalty0
  305--311, 2020.

\bibitem[Karani et~al.(2021)Karani, Erdil, Chaitanya, and
  Konukoglu]{karani2021test}
Neerav Karani, Ertunc Erdil, Krishna Chaitanya, and Ender Konukoglu.
\newblock Test-time adaptable neural networks for robust medical image
  segmentation.
\newblock \emph{Medical Image Analysis}, 68:\penalty0 101907, 2021.

\bibitem[Kaushal et~al.(2020)Kaushal, Altman, and
  Langlotz]{kaushal2020geographic}
Amit Kaushal, Russ Altman, and Curt Langlotz.
\newblock Geographic distribution of us cohorts used to train deep learning
  algorithms.
\newblock \emph{Jama}, 324\penalty0 (12):\penalty0 1212--1213, 2020.

\bibitem[Keenan et~al.(2018)Keenan, Ainslie, Barker, Boss, Cecil, Charles,
  Chenevert, Clarke, Evelhoch, Finn, et~al.]{keenan2018quantitative}
Kathryn~E Keenan, Maureen Ainslie, Alex~J Barker, Michael~A Boss, Kim~M Cecil,
  Cecil Charles, Thomas~L Chenevert, Larry Clarke, Jeffrey~L Evelhoch, Paul
  Finn, et~al.
\newblock Quantitative magnetic resonance imaging phantoms: a review and the
  need for a system phantom.
\newblock \emph{Magnetic resonance in medicine}, 79\penalty0 (1):\penalty0
  48--61, 2018.

\bibitem[Kelly et~al.(2019)Kelly, Karthikesalingam, Suleyman, Corrado, and
  King]{kelly2019key}
Christopher~J Kelly, Alan Karthikesalingam, Mustafa Suleyman, Greg Corrado, and
  Dominic King.
\newblock Key challenges for delivering clinical impact with artificial
  intelligence.
\newblock \emph{BMC medicine}, 17\penalty0 (1):\penalty0 1--9, 2019.

\bibitem[Khodabakhshi et~al.(2021)Khodabakhshi, Mostafaei, Arabi, Oveisi,
  Shiri, and Zaidi]{khodabakhshi2021non}
Zahra Khodabakhshi, Shayan Mostafaei, Hossein Arabi, Mehrdad Oveisi, Isaac
  Shiri, and Habib Zaidi.
\newblock Non-small cell lung carcinoma histopathological subtype phenotyping
  using high-dimensional multinomial multiclass ct radiomics signature.
\newblock \emph{Computers in biology and medicine}, 136:\penalty0 104752, 2021.

\bibitem[Kim et~al.(2018)Kim, Wattenberg, Gilmer, Cai, Wexler, Viegas,
  et~al.]{kim2018interpretability}
Been Kim, Martin Wattenberg, Justin Gilmer, Carrie Cai, James Wexler, Fernanda
  Viegas, et~al.
\newblock Interpretability beyond feature attribution: Quantitative testing
  with concept activation vectors (tcav).
\newblock In \emph{International conference on machine learning}, pages
  2668--2677. PMLR, 2018.

\bibitem[Kinyanjui et~al.(2020)Kinyanjui, Odonga, Cintas, Codella, Panda,
  Sattigeri, and Varshney]{kinyanjui2020fairness}
Newton~M Kinyanjui, Timothy Odonga, Celia Cintas, Noel~CF Codella, Rameswar
  Panda, Prasanna Sattigeri, and Kush~R Varshney.
\newblock Fairness of classifiers across skin tones in dermatology.
\newblock In \emph{International Conference on Medical Image Computing and
  Computer-Assisted Intervention}, pages 320--329. Springer, 2020.

\bibitem[Klapwijk et~al.(2019)Klapwijk, Van De~Kamp, Van Der~Meulen, Peters,
  and Wierenga]{klapwijk2019qoala}
Eduard~T Klapwijk, Ferdi Van De~Kamp, Mara Van Der~Meulen, Sabine Peters, and
  Lara~M Wierenga.
\newblock Qoala-t: A supervised-learning tool for quality control of freesurfer
  segmented mri data.
\newblock \emph{Neuroimage}, 189:\penalty0 116--129, 2019.

\bibitem[Koh et~al.(2020)Koh, Nguyen, Tang, Mussmann, Pierson, Kim, and
  Liang]{koh2020concept}
Pang~Wei Koh, Thao Nguyen, Yew~Siang Tang, Stephen Mussmann, Emma Pierson, Been
  Kim, and Percy Liang.
\newblock Concept bottleneck models.
\newblock In \emph{International Conference on Machine Learning}, pages
  5338--5348. PMLR, 2020.

\bibitem[LaLonde et~al.(2020)LaLonde, Torigian, and Bagci]{lalonde2020encoding}
Rodney LaLonde, Drew Torigian, and Ulas Bagci.
\newblock Encoding visual attributes in capsules for explainable medical
  diagnoses.
\newblock In \emph{International Conference on Medical Image Computing and
  Computer-Assisted Intervention}, pages 294--304. Springer, 2020.

\bibitem[Lambin et~al.(2017)Lambin, Leijenaar, Deist, Peerlings, De~Jong,
  Van~Timmeren, Sanduleanu, Larue, Even, Jochems, et~al.]{lambin2017radiomics}
Philippe Lambin, Ralph~TH Leijenaar, Timo~M Deist, Jurgen Peerlings, Evelyn~EC
  De~Jong, Janita Van~Timmeren, Sebastian Sanduleanu, Ruben~THM Larue, Aniek~JG
  Even, Arthur Jochems, et~al.
\newblock Radiomics: the bridge between medical imaging and personalized
  medicine.
\newblock \emph{Nature reviews Clinical oncology}, 14\penalty0 (12):\penalty0
  749--762, 2017.

\bibitem[Larrazabal et~al.(2020)Larrazabal, Nieto, Peterson, Milone, and
  Ferrante]{Larrazabal12592}
Agostina~J. Larrazabal, Nicol{\'a}s Nieto, Victoria Peterson, Diego~H. Milone,
  and Enzo Ferrante.
\newblock Gender imbalance in medical imaging datasets produces biased
  classifiers for computer-aided diagnosis.
\newblock \emph{Proceedings of the National Academy of Sciences}, 117\penalty0
  (23):\penalty0 12592--12594, 2020.
\newblock ISSN 0027-8424.

\bibitem[Larson et~al.(2021)Larson, Harvey, Rubin, Irani, Justin, and
  Langlotz]{larson2021regulatory}
David~B Larson, Hugh Harvey, Daniel~L Rubin, Neville Irani, R~Tse Justin, and
  Curtis~P Langlotz.
\newblock Regulatory frameworks for development and evaluation of artificial
  intelligence--based diagnostic imaging algorithms: summary and
  recommendations.
\newblock \emph{Journal of the American College of Radiology}, 18\penalty0
  (3):\penalty0 413--424, 2021.

\bibitem[Leclerc et~al.(2019)Leclerc, Smistad, Pedrosa, {\O}stvik, Cervenansky,
  Espinosa, Espeland, Berg, Jodoin, Grenier, et~al.]{leclerc2019deep}
Sarah Leclerc, Erik Smistad, Joao Pedrosa, Andreas {\O}stvik, Frederic
  Cervenansky, Florian Espinosa, Torvald Espeland, Erik Andreas~Rye Berg,
  Pierre-Marc Jodoin, Thomas Grenier, et~al.
\newblock Deep learning for segmentation using an open large-scale dataset in
  2d echocardiography.
\newblock \emph{IEEE transactions on medical imaging}, 38\penalty0
  (9):\penalty0 2198--2210, 2019.

\bibitem[Leiner et~al.(2021)Leiner, Bennink, Mol, Kuijf, and
  Veldhuis]{leiner2021bringing}
Tim Leiner, Edwin Bennink, Christian~P Mol, Hugo~J Kuijf, and Wouter~B
  Veldhuis.
\newblock Bringing ai to the clinic: blueprint for a vendor-neutral ai
  deployment infrastructure.
\newblock \emph{Insights into Imaging}, 12\penalty0 (1):\penalty0 1--11, 2021.

\bibitem[Lekadir et~al.(2023)Lekadir, Feragen, Fofanah, Frangi, Buyx, Emelie,
  Lara, Porras, Chan, Navarro, et~al.]{lekadir2023future}
Karim Lekadir, Aasa Feragen, Abdul~Joseph Fofanah, Alejandro~F Frangi, Alena
  Buyx, Anais Emelie, Andrea Lara, Antonio~R Porras, An-Wen Chan, Arcadi
  Navarro, et~al.
\newblock Future-ai: International consensus guideline for trustworthy and
  deployable artificial intelligence in healthcare.
\newblock \emph{arXiv preprint arXiv:2309.12325}, 2023.

\bibitem[Li et~al.(2015)Li, Abramson, Arlinghaus, Kang, Chakravarthy, Abramson,
  Farley, Mayer, Kelley, Meszoely, et~al.]{li2015multiparametric}
Xia Li, Richard~G Abramson, Lori~R Arlinghaus, Hakmook Kang, Anuradha~Bapsi
  Chakravarthy, Vandana~G Abramson, Jaime Farley, Ingrid~A Mayer, Mark~C
  Kelley, Ingrid~M Meszoely, et~al.
\newblock Multiparametric magnetic resonance imaging for predicting
  pathological response after the first cycle of neoadjuvant chemotherapy in
  breast cancer.
\newblock \emph{Investigative radiology}, 50\penalty0 (4):\penalty0 195--204,
  2015.

\bibitem[Li et~al.(2019)Li, Wu, Chen, and Jiang]{li2019deep}
Xiaoxiao Li, Junyan Wu, Eric~Z Chen, and Hongda Jiang.
\newblock From deep learning towards finding skin lesion biomarkers.
\newblock In \emph{2019 41st Annual International Conference of the IEEE
  Engineering in Medicine and Biology Society (EMBC)}, pages 2797--2800. IEEE,
  2019.

\bibitem[Li et~al.(2021)Li, Cui, Wu, Gu, and Harada]{li2021estimating}
Xiaoxiao Li, Ziteng Cui, Yifan Wu, Li~Gu, and Tatsuya Harada.
\newblock Estimating and improving fairness with adversarial learning.
\newblock \emph{arXiv preprint arXiv:2103.04243}, 2021.

\bibitem[Linardatos et~al.(2021)Linardatos, Papastefanopoulos, and
  Kotsiantis]{linardatos2021explainable}
Pantelis Linardatos, Vasilis Papastefanopoulos, and Sotiris Kotsiantis.
\newblock Explainable ai: A review of machine learning interpretability
  methods.
\newblock \emph{Entropy}, 23\penalty0 (1):\penalty0 18, 2021.

\bibitem[Lipton(2017)]{lipton2017doctor}
Zachary~C Lipton.
\newblock The doctor just won't accept that!
\newblock \emph{arXiv preprint arXiv:1711.08037}, 2017.

\bibitem[Litjens et~al.(2014)Litjens, Toth, van~de Ven, Hoeks, Kerkstra, van
  Ginneken, Vincent, Guillard, Birbeck, Zhang, et~al.]{litjens2014evaluation}
Geert Litjens, Robert Toth, Wendy van~de Ven, Caroline Hoeks, Sjoerd Kerkstra,
  Bram van Ginneken, Graham Vincent, Gwenael Guillard, Neil Birbeck, Jindang
  Zhang, et~al.
\newblock Evaluation of prostate segmentation algorithms for mri: the promise12
  challenge.
\newblock \emph{Medical image analysis}, 18\penalty0 (2):\penalty0 359--373,
  2014.

\bibitem[Liu et~al.(2019)Liu, Faes, Kale, Wagner, Fu, Bruynseels, Mahendiran,
  Moraes, Shamdas, Kern, et~al.]{liu2019comparison}
Xiaoxuan Liu, Livia Faes, Aditya~U Kale, Siegfried~K Wagner, Dun~Jack Fu, Alice
  Bruynseels, Thushika Mahendiran, Gabriella Moraes, Mohith Shamdas, Christoph
  Kern, et~al.
\newblock A comparison of deep learning performance against health-care
  professionals in detecting diseases from medical imaging: a systematic review
  and meta-analysis.
\newblock \emph{The lancet digital health}, 1\penalty0 (6):\penalty0
  e271--e297, 2019.

\bibitem[Lundberg and Lee(2017)]{lundberg2017unified}
Scott~M Lundberg and Su-In Lee.
\newblock A unified approach to interpreting model predictions.
\newblock In \emph{Proceedings of the 31st international conference on neural
  information processing systems}, pages 4768--4777, 2017.

\bibitem[Manikis et~al.(2021)Manikis, Ioannidis, Siakallis, Nikiforaki, Iv,
  Vozlic, Surlan-Popovic, Wintermark, Bisdas, and
  Marias]{manikis2021multicenter}
Georgios~C Manikis, Georgios~S Ioannidis, Loizos Siakallis, Katerina
  Nikiforaki, Michael Iv, Diana Vozlic, Katarina Surlan-Popovic, Max
  Wintermark, Sotirios Bisdas, and Kostas Marias.
\newblock Multicenter dsc--mri-based radiomics predict idh mutation in gliomas.
\newblock \emph{Cancers}, 13\penalty0 (16):\penalty0 3965, 2021.

\bibitem[Marcinkevi{\v{c}}s and Vogt(2020)]{marcinkevivcs2020interpretability}
Ri{\v{c}}ards Marcinkevi{\v{c}}s and Julia~E Vogt.
\newblock Interpretability and explainability: A machine learning zoo
  mini-tour.
\newblock \emph{arXiv preprint arXiv:2012.01805}, 2020.

\bibitem[Martin-Isla et~al.(2020)Martin-Isla, Campello, Izquierdo,
  Raisi-Estabragh, Bae{\ss}ler, Petersen, and Lekadir]{martin2020image}
Carlos Martin-Isla, Victor~M Campello, Cristian Izquierdo, Zahra
  Raisi-Estabragh, Bettina Bae{\ss}ler, Steffen~E Petersen, and Karim Lekadir.
\newblock Image-based cardiac diagnosis with machine learning: a review.
\newblock \emph{Frontiers in cardiovascular medicine}, 7:\penalty0 1, 2020.

\bibitem[McCartney et~al.(2019)McCartney, Popham, McMaster, and
  Cumbers]{mccartney2019defining}
Gerry McCartney, Franck Popham, Robert McMaster, and Andrew Cumbers.
\newblock Defining health and health inequalities.
\newblock \emph{Public health}, 172:\penalty0 22--30, 2019.

\bibitem[McKinney et~al.(2020)McKinney, Sieniek, Godbole, Godwin, Antropova,
  Ashrafian, Back, Chesus, Corrado, Darzi, et~al.]{mckinney2020international}
Scott~Mayer McKinney, Marcin Sieniek, Varun Godbole, Jonathan Godwin, Natasha
  Antropova, Hutan Ashrafian, Trevor Back, Mary Chesus, Greg~S Corrado, Ara
  Darzi, et~al.
\newblock International evaluation of an ai system for breast cancer screening.
\newblock \emph{Nature}, 577\penalty0 (7788):\penalty0 89--94, 2020.

\bibitem[McNitt-Gray et~al.(2020)McNitt-Gray, Napel, Jaggi, Mattonen,
  Hadjiiski, Muzi, Goldgof, Balagurunathan, Pierce, Kinahan,
  et~al.]{mcnitt2020standardization}
M~McNitt-Gray, S~Napel, A~Jaggi, SA~Mattonen, L~Hadjiiski, M~Muzi, D~Goldgof,
  Y~Balagurunathan, LA~Pierce, PE~Kinahan, et~al.
\newblock Standardization in quantitative imaging: a multicenter comparison of
  radiomic features from different software packages on digital reference
  objects and patient data sets.
\newblock \emph{Tomography}, 6\penalty0 (2):\penalty0 118--128, 2020.

\bibitem[Menze et~al.(2014)Menze, Jakab, Bauer, Kalpathy-Cramer, Farahani,
  Kirby, Burren, Porz, Slotboom, Wiest, et~al.]{menze2014multimodal}
Bjoern~H Menze, Andras Jakab, Stefan Bauer, Jayashree Kalpathy-Cramer, Keyvan
  Farahani, Justin Kirby, Yuliya Burren, Nicole Porz, Johannes Slotboom, Roland
  Wiest, et~al.
\newblock The multimodal brain tumor image segmentation benchmark (brats).
\newblock \emph{IEEE transactions on medical imaging}, 34\penalty0
  (10):\penalty0 1993--2024, 2014.

\bibitem[Minne et~al.(2012)Minne, Eslami, de~Keizer, de~Jonge, de~Rooij, and
  Abu-Hanna]{minne2012effect}
Lilian Minne, Saeid Eslami, Nicolette de~Keizer, Evert de~Jonge, Sophia~E
  de~Rooij, and Ameen Abu-Hanna.
\newblock Effect of changes over time in the performance of a customized
  saps-ii model on the quality of care assessment.
\newblock \emph{Intensive care medicine}, 38\penalty0 (1):\penalty0 40--46,
  2012.

\bibitem[Mittelstadt et~al.(2016)Mittelstadt, Allo, Taddeo, Wachter, and
  Floridi]{mittelstadt2016ethics}
Brent~Daniel Mittelstadt, Patrick Allo, Mariarosaria Taddeo, Sandra Wachter,
  and Luciano Floridi.
\newblock The ethics of algorithms: Mapping the debate.
\newblock \emph{Big Data \& Society}, 3\penalty0 (2):\penalty0
  2053951716679679, 2016.

\bibitem[Mohammadjafari et~al.()Mohammadjafari, Cevik, Thanabalasingam, and
  Basar]{mohammadjafariusing}
Sanaz Mohammadjafari, Mucahit Cevik, Mathusan Thanabalasingam, and Ayse Basar.
\newblock Using protopnet for interpretable alzheimer’s disease
  classification.

\bibitem[Mollura et~al.(2020)Mollura, Culp, Pollack, Battino, Scheel, Mango,
  Elahi, Schweitzer, and Dako]{mollura2020artificial}
Daniel~J Mollura, Melissa~P Culp, Erica Pollack, Gillian Battino, John~R
  Scheel, Victoria~L Mango, Ameena Elahi, Alan Schweitzer, and Farouk Dako.
\newblock Artificial intelligence in low-and middle-income countries:
  innovating global health radiology.
\newblock \emph{Radiology}, 297\penalty0 (3):\penalty0 513--520, 2020.

\bibitem[Mora-Cantallops et~al.(2021)Mora-Cantallops, S{\'a}nchez-Alonso,
  Garc{\'\i}a-Barriocanal, and Sicilia]{mora2021traceability}
Mar{\c{c}}al Mora-Cantallops, Salvador S{\'a}nchez-Alonso, Elena
  Garc{\'\i}a-Barriocanal, and Miguel-Angel Sicilia.
\newblock Traceability for trustworthy ai: A review of models and tools.
\newblock \emph{Big Data and Cognitive Computing}, 5\penalty0 (2):\penalty0 20,
  2021.

\bibitem[Moreau et~al.(2011)Moreau, Clifford, Freire, Futrelle, Gil, Groth,
  Kwasnikowska, Miles, Missier, Myers, et~al.]{moreau2011open}
Luc Moreau, Ben Clifford, Juliana Freire, Joe Futrelle, Yolanda Gil, Paul
  Groth, Natalia Kwasnikowska, Simon Miles, Paolo Missier, Jim Myers, et~al.
\newblock The open provenance model core specification (v1. 1).
\newblock \emph{Future generation computer systems}, 27\penalty0 (6):\penalty0
  743--756, 2011.

\bibitem[Morin et~al.(2021)Morin, Valli{\`e}res, Braunstein, Ginart, Upadhaya,
  Woodruff, Zwanenburg, Chatterjee, Villanueva-Meyer, Valdes,
  et~al.]{morin2021artificial}
Olivier Morin, Martin Valli{\`e}res, Steve Braunstein, Jorge~Barrios Ginart,
  Taman Upadhaya, Henry~C Woodruff, Alex Zwanenburg, Avishek Chatterjee,
  Javier~E Villanueva-Meyer, Gilmer Valdes, et~al.
\newblock An artificial intelligence framework integrating longitudinal
  electronic health records with real-world data enables continuous pan-cancer
  prognostication.
\newblock \emph{Nature Cancer}, 2\penalty0 (7):\penalty0 709--722, 2021.

\bibitem[{National Institute of Standards and Technology
  (NIST)}(2019)]{NIST2019}
{National Institute of Standards and Technology (NIST)}.
\newblock {U.S. LEADERSHIP IN AI: A Plan for Federal Engagement in Developing
  Technical Standards and Related Tools}, 2019.
\newblock URL
  \url{https://www.nist.gov/system/files/documents/2019/08/10/ai_standards_fedengagement_plan_9aug2019.pdf}.

\bibitem[Neri et~al.(2020)Neri, Coppola, Miele, Bibbolino, and
  Grassi]{neri2020artificial}
Emanuele Neri, Francesca Coppola, Vittorio Miele, Corrado Bibbolino, and
  Roberto Grassi.
\newblock Artificial intelligence: Who is responsible for the diagnosis?, 2020.

\bibitem[Nie et~al.(2016)Nie, Shi, Chen, Hu, Jabbour, Yue, Niu, and
  Sun]{nie2016rectal}
Ke~Nie, Liming Shi, Qin Chen, Xi~Hu, Salma~K Jabbour, Ning Yue, Tianye Niu, and
  Xiaonan Sun.
\newblock Rectal cancer: assessment of neoadjuvant chemoradiation outcome based
  on radiomics of multiparametric mri.
\newblock \emph{Clinical cancer research}, 22\penalty0 (21):\penalty0
  5256--5264, 2016.

\bibitem[Obermeyer et~al.(2019)Obermeyer, Powers, Vogeli, and
  Mullainathan]{obermeyer2019dissecting}
Ziad Obermeyer, Brian Powers, Christine Vogeli, and Sendhil Mullainathan.
\newblock Dissecting racial bias in an algorithm used to manage the health of
  populations.
\newblock \emph{Science}, 366\penalty0 (6464):\penalty0 447--453, 2019.

\bibitem[of~Radiology (ESR) communications@~myesr. org(2018)]{european2018esr}
European~Society of~Radiology (ESR) communications@~myesr. org.
\newblock Esr paper on structured reporting in radiology.
\newblock \emph{Insights into imaging}, 9:\penalty0 1--7, 2018.

\bibitem[Osuala et~al.(2019)Osuala, Li, and Arandjelovic]{osuala2019bringing}
Richard Osuala, Jieyi Li, and Ognjen Arandjelovic.
\newblock Bringing modern machine learning into clinical practice through the
  use of intuitive visualization and human--computer interaction.
\newblock \emph{Augmented Human Research}, 4\penalty0 (1):\penalty0 1--11,
  2019.

\bibitem[Osuala et~al.(2021)Osuala, Kushibar, Garrucho, Linardos, Szafranowska,
  Klein, Glocker, Diaz, and Lekadir]{osuala2021review}
Richard Osuala, Kaisar Kushibar, Lidia Garrucho, Akis Linardos, Zuzanna
  Szafranowska, Stefan Klein, Ben Glocker, Oliver Diaz, and Karim Lekadir.
\newblock A review of generative adversarial networks in cancer imaging: New
  applications, new solutions.
\newblock \emph{arXiv preprint arXiv:2107.09543}, 2021.

\bibitem[Panwar et~al.(2020)Panwar, Gupta, Siddiqui, Morales-Menendez,
  Bhardwaj, and Singh]{panwar2020deep}
Harsh Panwar, PK~Gupta, Mohammad~Khubeb Siddiqui, Ruben Morales-Menendez,
  Prakhar Bhardwaj, and Vaishnavi Singh.
\newblock A deep learning and grad-cam based color visualization approach for
  fast detection of covid-19 cases using chest x-ray and ct-scan images.
\newblock \emph{Chaos, Solitons \& Fractals}, 140:\penalty0 110190, 2020.

\bibitem[Park et~al.(2019)Park, Park, Kim, and Kim]{park2019reproducibility}
Ji~Eun Park, Seo~Young Park, Hwa~Jung Kim, and Ho~Sung Kim.
\newblock Reproducibility and generalizability in radiomics modeling: possible
  strategies in radiologic and statistical perspectives.
\newblock \emph{Korean journal of radiology}, 20\penalty0 (7):\penalty0
  1124--1137, 2019.

\bibitem[Park et~al.(2020)Park, Jackson, Foreman, Gruen, Hu, and
  Das]{park2020evaluating}
Yoonyoung Park, Gretchen~Purcell Jackson, Morgan~A Foreman, Daniel Gruen,
  Jianying Hu, and Amar~K Das.
\newblock Evaluating artificial intelligence in medicine: phases of clinical
  research.
\newblock \emph{JAMIA open}, 3\penalty0 (3):\penalty0 326--331, 2020.

\bibitem[Paulus and Kent(2020)]{paulus2020predictably}
Jessica~K Paulus and David~M Kent.
\newblock Predictably unequal: understanding and addressing concerns that
  algorithmic clinical prediction may increase health disparities.
\newblock \emph{NPJ digital medicine}, 3\penalty0 (1):\penalty0 1--8, 2020.

\bibitem[Pawlowski et~al.(2020)Pawlowski, Castro, and
  Glocker]{pawlowski2020deep}
Nick Pawlowski, Daniel~C Castro, and Ben Glocker.
\newblock Deep structural causal models for tractable counterfactual inference.
\newblock \emph{arXiv preprint arXiv:2006.06485}, 2020.

\bibitem[Pianykh et~al.(2020)Pianykh, Langs, Dewey, Enzmann, Herold,
  Schoenberg, and Brink]{pianykh2020continuous}
Oleg~S Pianykh, Georg Langs, Marc Dewey, Dieter~R Enzmann, Christian~J Herold,
  Stefan~O Schoenberg, and James~A Brink.
\newblock Continuous learning ai in radiology: implementation principles and
  early applications.
\newblock \emph{Radiology}, 297\penalty0 (1):\penalty0 6--14, 2020.

\bibitem[Pisov et~al.(2019)Pisov, Goncharov, Kurochkina, Morozov, Gombolevskiy,
  Chernina, Vladzymyrskyy, Zamyatina, Chesnokova, Pronin,
  et~al.]{pisov2019incorporating}
Maxim Pisov, Mikhail Goncharov, Nadezhda Kurochkina, Sergey Morozov, Victor
  Gombolevskiy, Valeria Chernina, Anton Vladzymyrskyy, Ksenia Zamyatina, Anna
  Chesnokova, Igor Pronin, et~al.
\newblock Incorporating task-specific structural knowledge into cnns for brain
  midline shift detection.
\newblock In \emph{Interpretability of Machine Intelligence in Medical Image
  Computing and Multimodal Learning for Clinical Decision Support}, pages
  30--38. Springer, 2019.

\bibitem[Pleiss et~al.(2017)Pleiss, Raghavan, Wu, Kleinberg, and
  Weinberger]{pleiss2017fairness}
Geoff Pleiss, Manish Raghavan, Felix Wu, Jon Kleinberg, and Kilian~Q
  Weinberger.
\newblock On fairness and calibration.
\newblock \emph{arXiv preprint arXiv:1709.02012}, 2017.

\bibitem[Pot et~al.(2021)Pot, Kieusseyan, and Prainsack]{pot2021not}
Mirjam Pot, Nathalie Kieusseyan, and Barbara Prainsack.
\newblock Not all biases are bad: equitable and inequitable biases in machine
  learning and radiology.
\newblock \emph{Insights into imaging}, 12\penalty0 (1):\penalty0 1--10, 2021.

\bibitem[Povyakalo et~al.(2013)Povyakalo, Alberdi, Strigini, and
  Ayton]{povyakalo2013discriminate}
Andrey~A Povyakalo, Eugenio Alberdi, Lorenzo Strigini, and Peter Ayton.
\newblock How to discriminate between computer-aided and computer-hindered
  decisions: a case study in mammography.
\newblock \emph{Medical Decision Making}, 33\penalty0 (1):\penalty0 98--107,
  2013.

\bibitem[Price et~al.(2019)Price, Gerke, and Cohen]{price2019potential}
W~Nicholson Price, Sara Gerke, and I~Glenn Cohen.
\newblock Potential liability for physicians using artificial intelligence.
\newblock \emph{Jama}, 322\penalty0 (18):\penalty0 1765--1766, 2019.

\bibitem[Prohl et~al.(2019)Prohl, Scherrer, Tomas-Fernandez, Filip-Dhima,
  Kapur, Velasco-Annis, Clancy, Carmody, Dean, Valle,
  et~al.]{prohl2019reproducibility}
Anna~K Prohl, Benoit Scherrer, Xavier Tomas-Fernandez, Rajna Filip-Dhima, Kush
  Kapur, Clemente Velasco-Annis, Sean Clancy, Erin Carmody, Meghan Dean, Molly
  Valle, et~al.
\newblock Reproducibility of structural and diffusion tensor imaging in the
  tacern multi-center study.
\newblock \emph{Frontiers in integrative neuroscience}, 13:\penalty0 24, 2019.

\bibitem[Prokop et~al.(2020)Prokop, Van~Everdingen, van Rees~Vellinga,
  Quarles~van Ufford, St{\"o}ger, Beenen, Geurts, Gietema, Krdzalic,
  Schaefer-Prokop, et~al.]{prokop2020co}
Mathias Prokop, Wouter Van~Everdingen, Tjalco van Rees~Vellinga, Henri{\"e}tte
  Quarles~van Ufford, Lauran St{\"o}ger, Ludo Beenen, Bram Geurts, Hester
  Gietema, Jasenko Krdzalic, Cornelia Schaefer-Prokop, et~al.
\newblock Co-rads: a categorical ct assessment scheme for patients suspected of
  having covid-19—definition and evaluation.
\newblock \emph{Radiology}, 296\penalty0 (2):\penalty0 E97--E104, 2020.

\bibitem[Puyol-Ant{\'o}n et~al.(2020)Puyol-Ant{\'o}n, Chen, Clough, Ruijsink,
  Sidhu, Gould, Porter, Elliott, Mehta, Rueckert,
  et~al.]{puyol2020interpretable}
Esther Puyol-Ant{\'o}n, Chen Chen, James~R Clough, Bram Ruijsink, Baldeep~S
  Sidhu, Justin Gould, Bradley Porter, Marc Elliott, Vishal Mehta, Daniel
  Rueckert, et~al.
\newblock Interpretable deep models for cardiac resynchronisation therapy
  response prediction.
\newblock In \emph{International Conference on Medical Image Computing and
  Computer-Assisted Intervention}, pages 284--293. Springer, 2020.

\bibitem[Puyol-Anton et~al.(2021)Puyol-Anton, Ruijsink, Piechnik, Neubauer,
  Petersen, Razavi, and King]{puyol2021fairness}
Esther Puyol-Anton, Bram Ruijsink, Stefan~K Piechnik, Stefan Neubauer,
  Steffen~E Petersen, Reza Razavi, and Andrew~P King.
\newblock Fairness in cardiac mr image analysis: An investigation of bias due
  to data imbalance in deep learning based segmentation.
\newblock \emph{arXiv preprint arXiv:2106.12387}, 2021.

\bibitem[Quionero-Candela et~al.(2009)Quionero-Candela, Sugiyama, Schwaighofer,
  and Lawrence]{quionero2009dataset}
Joaquin Quionero-Candela, Masashi Sugiyama, Anton Schwaighofer, and Neil~D
  Lawrence.
\newblock \emph{Dataset shift in machine learning}.
\newblock The MIT Press, 2009.

\bibitem[Radua et~al.(2020)Radua, Vieta, Shinohara, Kochunov, Quid{\'e}, Green,
  Weickert, Weickert, Bruggemann, Kircher, et~al.]{radua2020increased}
Joaquim Radua, Eduard Vieta, Russell Shinohara, Peter Kochunov, Yann Quid{\'e},
  Melissa~J Green, Cynthia~S Weickert, Thomas Weickert, Jason Bruggemann, Tilo
  Kircher, et~al.
\newblock Increased power by harmonizing structural mri site differences with
  the combat batch adjustment method in enigma.
\newblock \emph{NeuroImage}, 218:\penalty0 116956, 2020.

\bibitem[Rauscher et~al.(2013)Rauscher, Khan, Berbaum, and
  Conant]{rauscher2013potentially}
Garth~H Rauscher, Jenna~A Khan, Michael~L Berbaum, and Emily~F Conant.
\newblock Potentially missed detection with screening mammography: does the
  quality of radiologist's interpretation vary by patient socioeconomic
  advantage/disadvantage?
\newblock \emph{Annals of epidemiology}, 23\penalty0 (4):\penalty0 210--214,
  2013.

\bibitem[Recht et~al.(2020)Recht, Dewey, Dreyer, Curtis, Wiro, Prainsack, and
  Smith]{recht2020integrating}
Michael~P Recht, Marc Dewey, Keith Dreyer, Langlotz Curtis, Niessen Wiro,
  Barbara Prainsack, and John~J Smith.
\newblock Integrating artificial intelligence into the clinical practice of
  radiology: challenges and recommendations.
\newblock \emph{European radiology}, 30\penalty0 (6):\penalty0 3576--3584,
  2020.

\bibitem[Reyes et~al.(2020)Reyes, Meier, Pereira, Silva, Dahlweid,
  Tengg-Kobligk, Summers, and Wiest]{reyes2020interpretability}
Mauricio Reyes, Raphael Meier, S{\'e}rgio Pereira, Carlos~A Silva,
  Fried-Michael Dahlweid, Hendrik~von Tengg-Kobligk, Ronald~M Summers, and
  Roland Wiest.
\newblock On the interpretability of artificial intelligence in radiology:
  challenges and opportunities.
\newblock \emph{Radiology: Artificial Intelligence}, 2\penalty0 (3):\penalty0
  e190043, 2020.

\bibitem[Ribeiro et~al.(2016)Ribeiro, Singh, and Guestrin]{ribeiro2016should}
Marco~Tulio Ribeiro, Sameer Singh, and Carlos Guestrin.
\newblock " why should i trust you?" explaining the predictions of any
  classifier.
\newblock In \emph{Proceedings of the 22nd ACM SIGKDD international conference
  on knowledge discovery and data mining}, pages 1135--1144, 2016.

\bibitem[Rivenson et~al.(2018)Rivenson, Zhang, G{\"u}nayd{\i}n, Teng, and
  Ozcan]{rivenson2018phase}
Yair Rivenson, Yibo Zhang, Harun G{\"u}nayd{\i}n, Da~Teng, and Aydogan Ozcan.
\newblock Phase recovery and holographic image reconstruction using deep
  learning in neural networks.
\newblock \emph{Light: Science \& Applications}, 7\penalty0 (2):\penalty0
  17141--17141, 2018.

\bibitem[Sadri et~al.(2020)Sadri, Janowczyk, Zhou, Verma, Beig, Antunes,
  Madabhushi, Tiwari, and Viswanath]{sadri2020mrqy}
Amir~Reza Sadri, Andrew Janowczyk, Ren Zhou, Ruchika Verma, Niha Beig, Jacob
  Antunes, Anant Madabhushi, Pallavi Tiwari, and Satish~E Viswanath.
\newblock Mrqy—an open-source tool for quality control of mr imaging data.
\newblock \emph{Medical Physics}, 47\penalty0 (12):\penalty0 6029--6038, 2020.

\bibitem[Sahoo and Sheth(2009)]{sahoo2009provenir}
Satya~S Sahoo and Amit~P Sheth.
\newblock Provenir ontology: Towards a framework for escience provenance
  management.
\newblock 2009.

\bibitem[Samani et~al.(2020)Samani, Alappatt, Parker, Ismail, and
  Verma]{samani2020qc}
Zahra~Riahi Samani, Jacob~Antony Alappatt, Drew Parker, Abdol Aziz~Ould Ismail,
  and Ragini Verma.
\newblock Qc-automator: Deep learning-based automated quality control for
  diffusion mr images.
\newblock \emph{Frontiers in neuroscience}, 13:\penalty0 1456, 2020.

\bibitem[Samek et~al.(2016)Samek, Binder, Montavon, Lapuschkin, and
  M{\"u}ller]{samek2016evaluating}
Wojciech Samek, Alexander Binder, Gr{\'e}goire Montavon, Sebastian Lapuschkin,
  and Klaus-Robert M{\"u}ller.
\newblock Evaluating the visualization of what a deep neural network has
  learned.
\newblock \emph{IEEE transactions on neural networks and learning systems},
  28\penalty0 (11):\penalty0 2660--2673, 2016.

\bibitem[Sammut and Harries(2010)]{Sammut2010}
Claude Sammut and Michael Harries.
\newblock \emph{Concept Drift}, pages 202--205.
\newblock Springer US, Boston, MA, 2010.
\newblock ISBN 978-0-387-30164-8.
\newblock \doi{10.1007/978-0-387-30164-8_153}.
\newblock URL \url{https://doi.org/10.1007/978-0-387-30164-8_153}.

\bibitem[Sayres et~al.(2019)Sayres, Taly, Rahimy, Blumer, Coz, Hammel, Krause,
  Narayanaswamy, Rastegar, Wu, et~al.]{sayres2019using}
Rory Sayres, Ankur Taly, Ehsan Rahimy, Katy Blumer, David Coz, Naama Hammel,
  Jonathan Krause, Arunachalam Narayanaswamy, Zahra Rastegar, Derek Wu, et~al.
\newblock Using a deep learning algorithm and integrated gradients explanation
  to assist grading for diabetic retinopathy.
\newblock \emph{Ophthalmology}, 126\penalty0 (4):\penalty0 552--564, 2019.

\bibitem[Scheetz et~al.(2021)Scheetz, Rothschild, McGuinness, Hadoux, Soyer,
  Janda, Condon, Oakden-Rayner, Palmer, Keel, et~al.]{scheetz2021survey}
Jane Scheetz, Philip Rothschild, Myra McGuinness, Xavier Hadoux, H~Peter Soyer,
  Monika Janda, James~JJ Condon, Luke Oakden-Rayner, Lyle~J Palmer, Stuart
  Keel, et~al.
\newblock A survey of clinicians on the use of artificial intelligence in
  ophthalmology, dermatology, radiology and radiation oncology.
\newblock \emph{Scientific reports}, 11\penalty0 (1):\penalty0 1--10, 2021.

\bibitem[Schelter et~al.(2017)Schelter, Boese, Kirschnick, Klein, and
  Seufert]{schelter2017automatically}
Sebastian Schelter, Joos-Hendrik Boese, Johannes Kirschnick, Thoralf Klein, and
  Stephan Seufert.
\newblock Automatically tracking metadata and provenance of machine learning
  experiments.
\newblock In \emph{Machine Learning Systems Workshop at NIPS}, pages 27--29,
  2017.

\bibitem[Schlemper et~al.(2017)Schlemper, Caballero, Hajnal, Price, and
  Rueckert]{schlemper2017deep}
Jo~Schlemper, Jose Caballero, Joseph~V Hajnal, Anthony~N Price, and Daniel
  Rueckert.
\newblock A deep cascade of convolutional neural networks for dynamic mr image
  reconstruction.
\newblock \emph{IEEE transactions on Medical Imaging}, 37\penalty0
  (2):\penalty0 491--503, 2017.

\bibitem[Seegerer et~al.(2020)Seegerer, Binder, Saitenmacher, Bockmayr, Alber,
  Jurmeister, Klauschen, and M{\"u}ller]{seegerer2020interpretable}
Philipp Seegerer, Alexander Binder, Ren{\'e} Saitenmacher, Michael Bockmayr,
  Maximilian Alber, Philipp Jurmeister, Frederick Klauschen, and Klaus-Robert
  M{\"u}ller.
\newblock Interpretable deep neural network to predict estrogen receptor status
  from haematoxylin-eosin images.
\newblock In \emph{Artificial Intelligence and Machine Learning for Digital
  Pathology}, pages 16--37. Springer, 2020.

\bibitem[Selbst and Powles(2018)]{selbst2018meaningful}
Andrew Selbst and Julia Powles.
\newblock “meaningful information” and the right to explanation.
\newblock In \emph{Conference on Fairness, Accountability and Transparency},
  pages 48--48. PMLR, 2018.

\bibitem[Selvaraju et~al.(2017)Selvaraju, Cogswell, Das, Vedantam, Parikh, and
  Batra]{selvaraju2017grad}
Ramprasaath~R Selvaraju, Michael Cogswell, Abhishek Das, Ramakrishna Vedantam,
  Devi Parikh, and Dhruv Batra.
\newblock Grad-cam: Visual explanations from deep networks via gradient-based
  localization.
\newblock In \emph{Proceedings of the IEEE international conference on computer
  vision}, pages 618--626, 2017.

\bibitem[Seyyed-Kalantari et~al.(2020)Seyyed-Kalantari, Liu, McDermott, Chen,
  and Ghassemi]{seyyed2020chexclusion}
Laleh Seyyed-Kalantari, Guanxiong Liu, Matthew McDermott, Irene~Y Chen, and
  Marzyeh Ghassemi.
\newblock Chexclusion: Fairness gaps in deep chest x-ray classifiers.
\newblock In \emph{BIOCOMPUTING 2021: Proceedings of the Pacific Symposium},
  pages 232--243. World Scientific, 2020.

\bibitem[Simpson et~al.(2020)Simpson, Kay, Abbara, Bhalla, Chung, Chung, Henry,
  Kanne, Kligerman, Ko, et~al.]{simpson2020radiological}
Scott Simpson, Fernando~U Kay, Suhny Abbara, Sanjeev Bhalla, Jonathan~H Chung,
  Michael Chung, Travis~S Henry, Jeffrey~P Kanne, Seth Kligerman, Jane~P Ko,
  et~al.
\newblock Radiological society of north america expert consensus document on
  reporting chest ct findings related to covid-19: endorsed by the society of
  thoracic radiology, the american college of radiology, and rsna.
\newblock \emph{Radiology: Cardiothoracic Imaging}, 2\penalty0 (2):\penalty0
  e200152, 2020.

\bibitem[Sollini et~al.(2019)Sollini, Antunovic, Chiti, and
  Kirienko]{sollini2019towards}
Martina Sollini, Lidija Antunovic, Arturo Chiti, and Margarita Kirienko.
\newblock Towards clinical application of image mining: a systematic review on
  artificial intelligence and radiomics.
\newblock \emph{European journal of nuclear medicine and molecular imaging},
  46\penalty0 (13):\penalty0 2656--2672, 2019.

\bibitem[Sollini et~al.(2020)Sollini, Bartoli, Marciano, Zanca, Slart, and
  Erba]{sollini2020artificial}
Martina Sollini, Francesco Bartoli, Andrea Marciano, Roberta Zanca, Riemer~HJA
  Slart, and Paola~A Erba.
\newblock Artificial intelligence and hybrid imaging: the best match for
  personalized medicine in oncology.
\newblock \emph{European Journal of Hybrid Imaging}, 4\penalty0 (1):\penalty0
  1--22, 2020.

\bibitem[Song et~al.(2020)Song, Yin, Wang, Chang, Liu, and Cui]{song2020review}
Jiangdian Song, Yanjie Yin, Hairui Wang, Zhihui Chang, Zhaoyu Liu, and Lei Cui.
\newblock A review of original articles published in the emerging field of
  radiomics.
\newblock \emph{European journal of radiology}, 127:\penalty0 108991, 2020.

\bibitem[Souza et~al.(2019)Souza, Azevedo, Louren{\c{c}}o, Soares, Thiago,
  Brand{\~a}o, Civitarese, Brazil, Moreno, Valduriez,
  et~al.]{souza2019provenance}
Renan Souza, Leonardo Azevedo, V{\'\i}tor Louren{\c{c}}o, Elton Soares, Raphael
  Thiago, Rafael Brand{\~a}o, Daniel Civitarese, Emilio Brazil, Marcio Moreno,
  Patrick Valduriez, et~al.
\newblock Provenance data in the machine learning lifecycle in computational
  science and engineering.
\newblock In \emph{2019 IEEE/ACM Workflows in Support of Large-Scale Science
  (WORKS)}, pages 1--10. IEEE, 2019.

\bibitem[Springenberg et~al.(2014)Springenberg, Dosovitskiy, Brox, and
  Riedmiller]{springenberg2014striving}
Jost~Tobias Springenberg, Alexey Dosovitskiy, Thomas Brox, and Martin
  Riedmiller.
\newblock Striving for simplicity: The all convolutional net.
\newblock \emph{arXiv preprint arXiv:1412.6806}, 2014.

\bibitem[Sullivan and Schweikart(2019)]{sullivan2019current}
Hannah~R Sullivan and Scott~J Schweikart.
\newblock Are current tort liability doctrines adequate for addressing injury
  caused by ai?
\newblock \emph{AMA journal of ethics}, 21\penalty0 (2):\penalty0 160--166,
  2019.

\bibitem[Sundararajan et~al.(2017)Sundararajan, Taly, and
  Yan]{sundararajan2017axiomatic}
Mukund Sundararajan, Ankur Taly, and Qiqi Yan.
\newblock Axiomatic attribution for deep networks.
\newblock In \emph{International Conference on Machine Learning}, pages
  3319--3328. PMLR, 2017.

\bibitem[Suzuki et~al.(2019)Suzuki, Reyes, Syeda-Mahmood, Konukoglu, Glocker,
  Wiest, Gur, Greenspan, and Madabhushi]{Suzuki2019}
K.~Suzuki, M.~Reyes, T.~Syeda-Mahmood, E.~Konukoglu, B.~Glocker, R.~Wiest,
  Y.~Gur, H.~Greenspan, and A.~(Eds.) Madabhushi.
\newblock \emph{Interpretability of Machine Intelligence in Medical Image
  Computing and Multimodal Learning for Clinical Decision Support}.
\newblock Springer International Publishing, 2019.
\newblock URL \url{https://www.springer.com/gp/book/9783030338497}.

\bibitem[Tajbakhsh et~al.(2020)Tajbakhsh, Jeyaseelan, Li, Chiang, Wu, and
  Ding]{tajbakhsh2020embracing}
Nima Tajbakhsh, Laura Jeyaseelan, Qian Li, Jeffrey~N Chiang, Zhihao Wu, and
  Xiaowei Ding.
\newblock Embracing imperfect datasets: A review of deep learning solutions for
  medical image segmentation.
\newblock \emph{Medical Image Analysis}, 63:\penalty0 101693, 2020.

\bibitem[{THE FUTURE OF LIFE INSTITUTE (FLI)}(2017)]{AsilomarAI2017}
{THE FUTURE OF LIFE INSTITUTE (FLI)}.
\newblock {Asilomar AI Principles; Principles developed in conjunction with the
  2017 Asilomar conference 2017}, 2017.
\newblock URL \url{https://futureoflife.org/ai-principles/}.

\bibitem[{The Royal Australian and New Zealand College of
  Radiologists}(2019)]{principles2019}
{The Royal Australian and New Zealand College of Radiologists}.
\newblock {Ethical Principles for Artificial Intelligence in Medicine}, 2019.
\newblock URL
  \url{https://www.ranzcr.com/documents/4952-ethical-principles-for-ai-in-medicine/file}.

\bibitem[Thukral(2015)]{thukral2015problems}
Brij~Bhushan Thukral.
\newblock Problems and preferences in pediatric imaging.
\newblock \emph{The Indian journal of radiology \& imaging}, 25\penalty0
  (4):\penalty0 359, 2015.

\bibitem[Tor-Diez et~al.(2020)Tor-Diez, Porras, Packer, Avery, and
  Linguraru]{tor2020unsupervised}
Carlos Tor-Diez, Antonio~Reyes Porras, Roger~J Packer, Robert~A Avery, and
  Marius~George Linguraru.
\newblock Unsupervised mri homogenization: Application to pediatric anterior
  visual pathway segmentation.
\newblock In \emph{International Workshop on Machine Learning in Medical
  Imaging}, pages 180--188. Springer, 2020.

\bibitem[Tschandl et~al.(2019)Tschandl, Codella, Akay, Argenziano, Braun, Cabo,
  Gutman, Halpern, Helba, Hofmann-Wellenhof, et~al.]{tschandl2019comparison}
Philipp Tschandl, Noel Codella, Beng{\"u}~Nisa Akay, Giuseppe Argenziano,
  Ralph~P Braun, Horacio Cabo, David Gutman, Allan Halpern, Brian Helba, Rainer
  Hofmann-Wellenhof, et~al.
\newblock Comparison of the accuracy of human readers versus machine-learning
  algorithms for pigmented skin lesion classification: an open, web-based,
  international, diagnostic study.
\newblock \emph{The Lancet Oncology}, 20\penalty0 (7):\penalty0 938--947, 2019.

\bibitem[Tschandl et~al.(2020)Tschandl, Rinner, Apalla, Argenziano, Codella,
  Halpern, Janda, Lallas, Longo, Malvehy, et~al.]{tschandl2020human}
Philipp Tschandl, Christoph Rinner, Zoe Apalla, Giuseppe Argenziano, Noel
  Codella, Allan Halpern, Monika Janda, Aimilios Lallas, Caterina Longo, Josep
  Malvehy, et~al.
\newblock Human--computer collaboration for skin cancer recognition.
\newblock \emph{Nature Medicine}, 26\penalty0 (8):\penalty0 1229--1234, 2020.

\bibitem[{W3C}(2013)]{W3C_2}
{W3C}.
\newblock {PROV-O: The PROV Ontology}, 2013.
\newblock URL \url{https://www.w3.org/TR/prov-o/}.

\bibitem[{W3C Working Group}(2013)]{W3C}
{W3C Working Group}.
\newblock {PROV-Overview: An Overview of the PROV Family of Documents}, 2013.
\newblock URL \url{https://www.w3.org/TR/prov-overview/}.

\bibitem[Wang et~al.(2020)Wang, Dong, Hu, Li, Ren, Zhang, Shi, and
  Zhou]{wang2020temporal}
Yuhui Wang, Chengjun Dong, Yue Hu, Chungao Li, Qianqian Ren, Xin Zhang, Heshui
  Shi, and Min Zhou.
\newblock Temporal changes of ct findings in 90 patients with covid-19
  pneumonia: a longitudinal study.
\newblock \emph{Radiology}, 296\penalty0 (2):\penalty0 E55--E64, 2020.

\bibitem[Weiner et~al.(2017)Weiner, Veitch, Aisen, Beckett, Cairns, Green,
  Harvey, Jack~Jr, Jagust, Morris, et~al.]{weiner2017alzheimer}
Michael~W Weiner, Dallas~P Veitch, Paul~S Aisen, Laurel~A Beckett, Nigel~J
  Cairns, Robert~C Green, Danielle Harvey, Clifford~R Jack~Jr, William Jagust,
  John~C Morris, et~al.
\newblock The alzheimer's disease neuroimaging initiative 3: Continued
  innovation for clinical trial improvement.
\newblock \emph{Alzheimer's \& Dementia}, 13\penalty0 (5):\penalty0 561--571,
  2017.

\bibitem[Willemink et~al.(2020)Willemink, Koszek, Hardell, Wu, Fleischmann,
  Harvey, Folio, Summers, Rubin, and Lungren]{willemink2020preparing}
Martin~J Willemink, Wojciech~A Koszek, Cailin Hardell, Jie Wu, Dominik
  Fleischmann, Hugh Harvey, Les~R Folio, Ronald~M Summers, Daniel~L Rubin, and
  Matthew~P Lungren.
\newblock Preparing medical imaging data for machine learning.
\newblock \emph{Radiology}, 295\penalty0 (1):\penalty0 4--15, 2020.

\bibitem[Wilson(2009)]{wilson2009user}
Chauncey Wilson.
\newblock \emph{User experience re-mastered: your guide to getting the right
  design}.
\newblock Morgan Kaufmann, 2009.

\bibitem[{World Health Organization}(2020)]{10665-339554}
{World Health Organization}.
\newblock \emph{Basic documents}.
\newblock World Health Organization, 49th ed edition, 2020.

\bibitem[Wyman et~al.(2013)Wyman, Harvey, Crawford, Bernstein, Carmichael,
  Cole, Crane, DeCarli, Fox, Gunter, et~al.]{wyman2013standardization}
Bradley~T Wyman, Danielle~J Harvey, Karen Crawford, Matt~A Bernstein, Owen
  Carmichael, Patricia~E Cole, Paul~K Crane, Charles DeCarli, Nick~C Fox,
  Jeffrey~L Gunter, et~al.
\newblock Standardization of analysis sets for reporting results from adni mri
  data.
\newblock \emph{Alzheimer's \& Dementia}, 9\penalty0 (3):\penalty0 332--337,
  2013.

\bibitem[Xu et~al.(2019)Xu, Hosny, Zeleznik, Parmar, Coroller, Franco, Mak, and
  Aerts]{xu2019deep}
Yiwen Xu, Ahmed Hosny, Roman Zeleznik, Chintan Parmar, Thibaud Coroller, Idalid
  Franco, Raymond~H Mak, and Hugo~JWL Aerts.
\newblock Deep learning predicts lung cancer treatment response from serial
  medical imaging.
\newblock \emph{Clinical Cancer Research}, 25\penalty0 (11):\penalty0
  3266--3275, 2019.

\bibitem[Yang et~al.(2021)Yang, Ye, and Xia]{yang2021unbox}
Guang Yang, Qinghao Ye, and Jun Xia.
\newblock Unbox the black-box for the medical explainable ai via multi-modal
  and multi-centre data fusion: A mini-review, two showcases and beyond.
\newblock \emph{arXiv preprint arXiv:2102.01998}, 2021.

\bibitem[Young et~al.(2019)Young, Booth, Simpson, Dutton, and
  Shrapnel]{young2019deep}
Kyle Young, Gareth Booth, Becks Simpson, Reuben Dutton, and Sally Shrapnel.
\newblock Deep neural network or dermatologist?
\newblock In \emph{Interpretability of Machine Intelligence in Medical Image
  Computing and Multimodal Learning for Clinical Decision Support}, pages
  48--55. Springer, 2019.

\bibitem[Yu and Kohane(2019)]{yu2019framing}
Kun-Hsing Yu and Isaac~S Kohane.
\newblock Framing the challenges of artificial intelligence in medicine.
\newblock \emph{BMJ quality \& safety}, 28\penalty0 (3):\penalty0 238--241,
  2019.

\bibitem[Zaharia et~al.(2018)Zaharia, Chen, Davidson, Ghodsi, Hong, Konwinski,
  Murching, Nykodym, Ogilvie, Parkhe, et~al.]{zaharia2018accelerating}
Matei Zaharia, Andrew Chen, Aaron Davidson, Ali Ghodsi, Sue~Ann Hong, Andy
  Konwinski, Siddharth Murching, Tomas Nykodym, Paul Ogilvie, Mani Parkhe,
  et~al.
\newblock Accelerating the machine learning lifecycle with mlflow.
\newblock \emph{IEEE Data Eng. Bull.}, 41\penalty0 (4):\penalty0 39--45, 2018.

\bibitem[Zhang et~al.(2020)Zhang, Petitjean, Yger, and
  Ainouz]{zhang2020explainability}
Jing Zhang, Caroline Petitjean, Florian Yger, and Samia Ainouz.
\newblock Explainability for regression cnn in fetal head circumference
  estimation from ultrasound images.
\newblock In \emph{Interpretable and Annotation-Efficient Learning for Medical
  Image Computing}, pages 73--82. Springer, 2020.

\bibitem[Zhang et~al.(2017)Zhang, Sparks, and Franklin]{zhang2017diagnosing}
Zhao Zhang, Evan~R Sparks, and Michael~J Franklin.
\newblock Diagnosing machine learning pipelines with fine-grained lineage.
\newblock In \emph{Proceedings of the 26th International Symposium on
  High-Performance Parallel and Distributed Computing}, pages 143--153, 2017.

\bibitem[Zwanenburg et~al.(2020)Zwanenburg, Valli{\`e}res, Abdalah, Aerts,
  Andrearczyk, Apte, Ashrafinia, Bakas, Beukinga, Boellaard,
  et~al.]{zwanenburg2020image}
Alex Zwanenburg, Martin Valli{\`e}res, Mahmoud~A Abdalah, Hugo~JWL Aerts,
  Vincent Andrearczyk, Aditya Apte, Saeed Ashrafinia, Spyridon Bakas, Roelof~J
  Beukinga, Ronald Boellaard, et~al.
\newblock The image biomarker standardization initiative: standardized
  quantitative radiomics for high-throughput image-based phenotyping.
\newblock \emph{Radiology}, 295\penalty0 (2):\penalty0 328--338, 2020.

\end{thebibliography}

\end{document}